\documentclass{article}
\usepackage[T1]{fontenc}
\usepackage[utf8]{inputenc}

\usepackage[margin=1.5in]{geometry}
\usepackage{graphicx}

\usepackage{amsmath,amssymb,amsthm}
\numberwithin{equation}{section}

\usepackage[dvipsnames]{xcolor}

\usepackage{hyperref}

\usepackage{todonotes}
\usepackage{url}\usepackage{caption}
\usepackage{subcaption}

\usepackage[nameinlink, capitalise, noabbrev]{cleveref}
\usepackage[authoryear,sort,round]{natbib}  
\usepackage{subcaption}
\usepackage{xfrac}
\usepackage{nicefrac}
\usepackage{tikz}
\usepackage{pgfplots}
\usepackage[ruled,linesnumbered,noend]{algorithm2e}
\makeatletter
\newcommand{\nosemic}{\renewcommand{\@endalgocfline}{\relax}}
\newcommand{\dosemic}{\renewcommand{\@endalgocfline}{\algocf@endline}}
\let\oldnl\nl
\newcommand{\nonl}{\renewcommand{\nl}{\let\nl\oldnl}}
\makeatother
\SetKwComment{Comment}{// }{}
\SetCommentSty{mycommfont}

\usetikzlibrary{3d,calc}

\newtheorem{theorem}{Theorem}[section]

\theoremstyle{remark}
\newtheorem{remark}[theorem]{Remark}

\theoremstyle{definition}

\numberwithin{equation}{section}

\newcommand{\N}{\mathbb{N}}

\newcommand{\R}{\mathbb{R}}

\newcommand{\abs}[1]{\left\vert #1 \right\vert}

\DeclareMathOperator{\argmin}{arg\,min}
\DeclareMathOperator{\argmax}{arg\,max}

\DeclareMathOperator{\supp}{supp}

\renewcommand{\d}{\,\mathrm{d}}

\newcommand{\eps}{\varepsilon}

\newcommand{\M}{\mathcal M}

\newcommand{\wmean}{\mathsf{m}_\beta}

\newcommand{\rv}[1]{#1}
\newcommand{\rvv}[1]{#1}

\Crefname{algocf}{Algorithm}{Algorithms}

\title{Ensemble-based gradient inference for particle methods in optimization and sampling}
\makeatletter
\author{Claudia Schillings\thanks{Freie Universität Berlin, Germany}, Claudia Totzeck\thanks{University of Wuppertal, Germany}, and Philipp Wacker\thanks{School of Mathematics and Statistics, University of Canterbury, Christchurch, NZ, \texttt{phkwacker@gmail.com}}}
\makeatother
\usepackage{cleveref}

\begin{document}

\maketitle
\begin{abstract}    
    We propose an approach based on function evaluations and Bayesian inference to extract higher-order differential information of objective functions {from a given ensemble of particles}. \rv{Pointwise evaluation of some potential V in an ensemble} contains implicit information about first or higher order derivatives, which can be made explicit with little computational effort (ensemble-based gradient inference -- EGI). We suggest to use this information for the improvement of established ensemble-based numerical methods for optimization and sampling such as Consensus-based optimization and Langevin-based samplers. Numerical studies indicate that the augmented algorithms are often superior to their gradient-free variants, in particular the augmented methods help the ensembles to escape their initial domain, to explore multimodal, non-Gaussian settings and to speed up the collapse at the end of optimization dynamics.
    The code for the numerical examples in this manuscript can be found in the paper's Github repository\footnote{ \href{https://github.com/MercuryBench/ensemble-based-gradient.git}{https://github.com/MercuryBench/ensemble-based-gradient.git}.}

    MSC subject class: 62F15, 65N75, 90C56, 90C26, 35Q83, 37N40, 60H10  

\end{abstract}


\section{Introduction}

Global optimization of nonconvex objective functions and sampling from nonstandard distributions are widespread applications in industry, finance and other disciplines. Although the problems read very simple, and there is a variety of algorithms available, it is still very challenging to design well-performing algorithms and to prove their convergence. 
We motivate both the optimzation and sampling problem via the inverse setting, where the goal is to identify unknown parameters $x\in\mathbb R^d$ from noisy observations
\begin{equation}\label{eq:invprsetting}
    y=G(x)+\eta\,,
\end{equation}
with $G:\mathbb R^d \to \mathbb R^l$ denoting the parameter-to-observation map and $\eta$ being the noise. Under suitable assumptions on $G$ and $\eta$, \rv{and after specifying a prior measure $\pi_0$ on the unknown parameter $x$, an} application of Bayes' theorem gives as solution of the (Bayesian) inverse problem the following characterization of the posterior distribution
\begin{equation}
   \frac{ \mathrm d \mu(x)}{\mathrm d \lambda(x)}\propto \exp(-V(x)) 
\end{equation}
\rv{with respect to} the Lebesgue measure $\lambda$, where $V$ denotes a regularized potential. \rv{For example, if the prior $\pi_0$ is Gaussian with mean $\hat m$ and covariance $\hat \Sigma$, and if the noise term $\eta$ is Gaussian with mean $0$ and covariance $\Gamma$, then $V(x) = \frac{1}{2}\|y-G(x)\|_\Gamma^2 + \frac{1}{2}\|x-\hat m\|_{\hat \Sigma}^2$.} The aim of sampling methods is then to generate samples (approximately) distributed according to the \rv{measure $\mu$}. The connection to the optimization setting is via the maximum-a-posteriori (MAP) estimate (assuming the unique existence)  given by
\begin{equation}
    x^\star = \argmax\limits_x \exp(-V(x))=\argmin\limits_x V(x)\,.
\end{equation}

\rv{The optimization and sampling algorithms proposed in the following will work with a general potential $V$, but we will compare their performance to some well-known sampling algorithms which can be applied only in settings where the measure of interest arises as a potential from the inverse problem \eqref{eq:invprsetting}. This means that we will (sometimes) \rvv{restrict} the setting so that $V$ arises as the potential from a Bayesian inverse problem of the type above to make the performance of the algorithms comparable.}

The Bayesian approach to inverse problems has become very popular over the last decades and there has been a lot of research effort towards efficient methods, in particular gradient-free methods in order to allow the use of black-box solvers for the underlying forward problem. Kalman-Wasserstein flows have been the basis to design various efficient, gradient-free samplers, e.g. Ensemble Kalman Sampler \citep{garbuno2020interacting,nusken2019note}, Affine Invariant Interacting Langevin Dynamics \citep{garbuno2020affine}. {Also in the optimization setting, gradient free methods such as Consensus-based optimization method (CBO) and Ensemble Kalman Inversion (EKI) have become very popular, see e.g. \cite{carrillo2018analytical, carrillo2021consensus,totzeck2022trends, schillings2017analysis,ClS-4} and the references therein.} These methods have in common that they rely on an ensemble of particles, which is transformed into posterior samples or concentrates around the minimizer in the limit 'pseudo-time to infinity'. We will show in the following that the ensemble itself can be used to efficiently approximate derivative information (without any further evaluations of the forward problem). \rv{The difference to methods like EKI or EnRML \citep{chen2012ensemble}, which perform statistical linearization \citep{chada2020iterative}, is that we interpret evaluations as data points in linear inverse problem (this is similar to the concept of Simplex Gradients, see discussion below), taking into account locality via Taylor's theorem. }  To demonstrate the potential of the simultaneous estimation of the derivative information and parameter estimation, we consider the ensemble based gradient augmentation of the following state-of-the art optimization and sampling methods:
\begin{itemize}
    \item Consensus-based optimization (CBO)
    \item Ensemble Langevin sampler (LS)
    \item Metropolis-adjusted Langevin algorithm (MALA)
    \item Affine invariant interacting Langevin dynamics for Bayesian inference (ALDI) and Ensemble Kalman sampler (EKS).
\end{itemize}

In order to keep the presentation self-contained and for later reference we review the methods in the following.

\subsection{Review of existing computational methods}\label{sub:existingmethods}
We consider a potential $V$ as described above which may be the negative logdensity of some Bayesian posterior measure, or just a function we want to minimize. We will call the task of finding the minimal value of $V$ (if it exists) the \textit{optimization task}, and the task of constructing samples from the measure with unnormalized Lebesgue-density $\exp(-V)$ will be called the \textit{sampling task}.

\paragraph{CBO (for optimization)}
The consensus-based optimization (CBO) method \rvv{\citep{carrillo2018analytical,pinnau2017consensus}} describes the collective dynamical behavior of an ensemble of particles exploring the state space experiencing some diffusive behaviour and finally collapsing onto a joint weighted ensemble mean, which is by construction an approximation of the minimizer of $V.$  
Given a measure $\rho\in\M(\R^d)$ we define the weighted mean of $\rho$ as
\begin{align}
    \wmean(\rho) := 
    \frac{\int x \exp(-\beta V(x))\d\rho(x)}{\int \exp(-\beta V(x))\d\rho(x)}.
\end{align}
This definition of weighted mean is motivated by the Laplace principle \rvv{\citep{pinnau2017consensus}}. Indeed, assuming that $V$ admits a unique global minimizer, it can be shown that for $\beta\rightarrow\infty,$ $\wmean(\rho)$ tends to the $\argmin\limits_{x\in \supp{\rho}} V(x)$.

The CBO dynamics combines exploration of the landscape given by $V$ with aggregation of the ensemble around $\wmean(\rho).$ More succinctly, it follows the following system of stochastic differential equations (SDEs), where $\rho := \sum_{i=1}^J\delta_{x^i}.$
\begin{align}\label{eq:CBO}
    \d x^i &= -\lambda(x^i - \wmean(\rho)) \d t + \sigma \abs{x^i - \wmean(\rho)} \d W_t^i, \qquad i=1\ldots,J
\end{align}
with initial conditions $x_0^i \sim \mathcal P_2(\R^d)$ drawn independently. It is important to note that both the drift and the diffusion part of the dynamics scale with the distance $|x^i - \wmean(\rho)|.$ On the one hand this allows for collapse at $\wmean(\rho)$ which is important for optimization tasks. On the other hand, the dynamics slows down as the ensemble variance becomes small. \rv{Let us note at this point that this algorithm only indirectly uses information about the potential $V$, namely by pushing all particles towards the weighted mean $\wmean$ with the hope of improving their position. Our contribution will be to add a third term which nudges all particles in a direction informed by approximated gradient information, acquired from the existing pointwise evaluations $V(x^i)$.}

We will call the algorithm corresponding to \eqref{eq:CBO} ``vanilla CBO'' and in case the componentwise noise modification of \cite{carrillo2018analytical} is used, we will specify this as ``component-wise noise vanilla CBO''.
In \cite{pinnau2017consensus} the Consensus-based optimization method (CBO) was proposed as alternative to heuristic gradient-free particle optimization methods such as evolutionary or genetic algorithms and simulated annealing \citep{simon2013evolutionary,back1996evolutionary,laarhoven1987simulated} with the intention to prove convergence for the corresponding mean-field dynamics \citep{carrillo2021consensus}. Based on these publications a variety of researchers from stochastic analysis \citep{huang2022meanfield,kalise2022consensus}, PDE analysis \citep{fornaiser2021globally,fornasier2021anisotropic} found ways to relax assumptions, give further insight in the internal process of the method \citep{fornaiser2021globally}, open the class of applications like constraint problems \citep{fornasier2021sphereML,fornasier2020hypersurface,ha2022stiefel}, machine learning \citep{carrillo2021consensus}, multiobjective problems \citep{borghi2022multi,klamroth2022multi} and sampling \citep{carrillo2022sampling}. Others aim to make the algorithm more reliable \citep{totzeck2020personal}. As the original method with minimal dynamics is fairly understood, there are mainly two paths to go, either improve the method itself or use the established ideas to prove convergence results for other well-known optimization dynamics, see for example \citep{grassi2021pso,huang2021pso} for Particle Swarm Optimization.

\paragraph{Langevin dynamics and MALA (for sampling)}
{A well-known sampling dynamics is given by the (overdamped) Langevin equation, cp. \cite{Yang2019OptimalSO} for a recent overview for the connection of Langevin dynamics and Markov chain Monte Carlo (MCMC) methods. Given a symmetrical and positive definite matrix $M$, the} (overdamped) Langevin equation preconditioned by $M$ models a particle subject to the influence of the potential $V$:
\begin{equation}
\label{eq:langevin}\d x_t = - M\nabla V(x_t)\d t + \sqrt{2}M^{1/2} \d W_t.
\end{equation}
{It can be shown that the invariant measure of this stochastic differential equation is equal to $\mu$, i.e. the target measure $\mu$ with unnormalized Lebesgue-density $\exp(-V )$ \citep{garbuno2020affine, doi:10.1137/S0036141096303359}}. In applications the gradient of $V$ may be expensive to compute or simply not available. It is therefore of interest to construct a surrogate model with a gradient approximation. 

\paragraph{ALDI and EKS}
Another recent ensemble-based method with pointwise function evaluations is the gradient-free ALDI sampler \citep{garbuno2020affine}. It performs an implicit approximation to the gradient of the log-likelihood. There is a close relationship between ALDI and the Ensemble Kalman sampler (EKS)\footnote{\rv{In fact, they only differ by a correction term vanishing in the limit $J\to\infty$, see the discussion in \citet{nusken2019note}}} \citep{garbuno2020interacting,garbuno2020affine} which is an efficient gradient-free interacting particle sampler originating from the Ensemble Kalman filter. EKS can be applied (only) if the negative logdensity of $\rho$ can be written in the form $V(x) = \frac{1}{2}\|y-G(x)\|_\Gamma^2 + \frac{1}{2}\|x-\mu_0\|_{\Sigma_0}^2$, which is typical for a Bayesian statistical problem of type \eqref{eq:invprsetting}. In this special case, the gradient-free ALDI is the coupled system of SDEs given by 
\begin{equation}\label{eq:ALDI}
\begin{split}
    \d x^i_t &= -\{D(X_t)\Gamma^{-1}(G(x^i_t)-y) +C(X_t)\Sigma_0^{-1}(x_t^i-\mu_0)\}\d t\\ & \qquad +\frac{d+1}{J}(x_t^i-\bar x_t)\d t + \sqrt{2}C(X_t)^\frac{1}{2}\d W_t^i
\end{split}
\end{equation}
where $d$ is the dimension of the space and
\begin{equation}
\begin{split}
     C(X) &= \frac1N\sum_{j=1}^J (x^j-\bar x)\otimes (x^j-\bar x)\\
     D(X) &= \frac1N\sum_{j=1}^J (x^j-\bar x)\otimes (G(x^j)-\bar G),\quad \bar G =\frac1J\sum_{j=1}^J G(x^j) 
\end{split}
\end{equation}
with initial conditions analogous to the methods above.
If $G$ is a linear map, samples from $\rho$ are provable accurately generated. One of its main strengths stems from the fact that the Ensemble Kalman method constructs a covariance-preconditioned approximation to the gradient: Note that if $V(x) = \frac{1}{2}\|G(x) - y\|_\Gamma^2$ and $G$ linear, then
\begin{align*}
    C(X)\cdot \nabla V(x^i) &= \frac1J\sum_{j=1}^J (x^j-\bar x)\langle x^j-\bar x, \nabla V(x^i)\rangle\\
    &=\frac1J\sum_{j=1}^J (x^j-\bar x)\langle x^j-\bar x, G^T\Gamma^{-1}(Gx^i - y)\rangle\\
    &=\frac1J\sum_{j=1}^J (x^j-\bar x)\langle G(x^j)-G(\bar x), \Gamma^{-1}(Gx^i
    - y)\rangle\\
    &= D(X)\cdot \Gamma^{-1}(G(x^i)-y).
\end{align*}   
For $G$ linear, the approximation is exact. In fact, in the linear regime \eqref{eq:ALDI} corresponds to 
\[ \d x_t^i = -C(X)\cdot \nabla V(x_t^i) \d t+ \frac{d+1}{J}(x_t^i-\bar x_t) \d t+ \sqrt{2}C(X_t)^\frac{1}{2}\d W_t^i,\]
{which is an overdamped Langevin equation with a correction term ensuring the affine invariance of the resulting scheme \rvv{\citep{garbuno2020affine}.}}

\subsection{Main idea and contributions of the paper}

We argue that tracking an ensemble of particles $\{x^i\}_{i=1}^J$ with pointwise evaluations $\{V(x^i)\}_{i=1}^J$ of some potential function $V$ carries implicit gradient (and higher-order differential) information which can be converted into explicit information basically ``for free'', via an often negligible cost of solving one linear equation system.  \rv{This concept is being actively used in some areas of derivative-free optimization, but seems less known in the communities of Uncertainty Quantification (UQ) and Statistical Computing, as evidenced by the shortage of computational methods in UQ utilizing this concept. Our manuscript strives to close this gap.} 

The CBO algorithm for optimization is completely gradient-agnostic and can benefit from incorporation of gradient information via EGI in order to speed up local convergence (``better convergence to minima'') and improve overall performance (``convergence to better minima'').

Langevin dynamics and MALA in its original form need explicit gradient information. By managing a full ensemble instead of just one particle and thereby providing approximated gradient information for each member of the ensemble via EGI, we can run approximated Langevin dynamics and MALA without the need for an explicit gradient functional.

EKS and ALDI utilize a powerful property of the empirical covariance of an ensemble in order to compute a preconditioned gradient without the need for any additional computation. This allows us to sample very efficiently from a Bayesian posterior. Unfortunately, this is an approximation that holds only in the linear, Gaussian setting. With EGI, we can obtain a better approximation of the gradient, allowing for construction of non-Gaussian posterior samples.

Summarized, we present, motivate, and test the following novel algorithms. 
    \begin{itemize}
        \item \textbf{EGI}: Gradient approximation from ensemble point evaluation (as a general method)
        \item  \textbf{EGI-CBO}: CBO augmented by gradient information
        \item \textbf{EGI-LS}: Coupled Langevin dynamics with approximated gradient information
        \item \textbf{EGI-MALA}: Coupled MALA dynamics with approximated gradient information
        \item \textbf{EGI-ALDI}: Gradient-free ALDI with approximated gradients, with some variants
    \end{itemize}

{The rest of the article is structured as follows: \Cref{sec:gradinference} describes how pointwise evaluation of some function $V$ evaluated in an ensemble $\{x^i\}_{i=1}^J$ can be converted into explicit inexact gradient information via solving a linear inverse problem from straightforward Taylor approximation. We call this method EGI -- Ensemble-based gradient inference. \Cref{sec:CBO} develops EGI-CBO, a consensus-based optimization method augmented by inexact gradient information. Finally, \Cref{sec:sampling} generalizes this idea to ensemble-based sampling methods which normally use exact gradient evaluation, but can be shown to work with inexact gradients supplied by EGI as well.}

\section{Ensemble-based gradient inference (EGI) from pointwise ensemble evaluation} \label{sec:gradinference}
\rv{What follows below is strongly related to the concept of Simplex Gradients \citep{bortz1998simplex,custodio2008using,conn2009introduction,regis2015calculus,coope2019efficient,coope2021gradient}, but our method generalizes this idea (approximation to arbitrary derivative order, a specific ensemble-related ansatz for all derivatives, error estimation via Taylor's formula, potential for tradeoff between localization and globalization, and the possibility of generating samples instead of point estimators for the derivatives). Nevertheless there is considerable overlap.  We will see that there is some connection to the concept of ordinary least squares regression, or the Scatterplot fitting technique LOESS \citep{cleveland1979robust,cleveland1988locally,fox2011r}. Since our method is neither a subset of Simplex Gradients nor of the idea of LOESS, but rather a generalization or variant of both, we will give it a new name (EGI). For the sake of completeness we will give a full derivation from the ground up instead of referring to other methods.}
Consider an unknown function $V: \R^d\to \R$ evaluated pointwise in an ensemble $\{x^k\}_{k=1}^J\subset \R^d$. Our goal is to infer gradient information (as well as possibly higher-order derivatives) at one or each of the ensemble members' positions $x^j$, i.e. $\nabla V(x^j)$ (and possibly higher-order derivatives). In short, we want to solve the ill-posed problem
\[ \{V(x^k)\}_{k=1}^J \mapsto \nabla V(x^\star), \]
where $x^\star$ is an element of the ensemble $\{x^k\}_{k=1}^J$\footnote{It is conceivable to allow $x^\star$ to be an arbitrary point ``near'' the ensemble by appropriately modifying the ansatz below.}. We now describe our approach where we set $x^\star = x^j$ for some fixed $j\in\{1,\ldots, J\}$.
 By Taylor's formula, for any $i\in\{1,\ldots,J\}$,
 \begin{equation}\label{eq:zerothform}
\begin{split}
    V(x^i) - V(x^j) &\approx \nabla V(x^j)^T(x^i - x^j)\\
    &+\frac{1}{2}(x^i-x^j)^THV(x^j)(x^i-x^j) +  \eps_i^j\cdot  \frac{\|x^i-x^j\|^3}{6},
\end{split}
 \end{equation}
where we model the action of the unknown tensors $D^3 V(x^j)$ on $[x^i-x^j,x^i-x^j,x^i-x^j]$ by independent random variables $\{\eps_i^j\}_{i=1}^J \sim \mathcal N(0,\gamma^2)$. 
 We make the ansatz \rv{
 \begin{equation}\label{eq:ansatz}
 \begin{split}
     \nabla V(x^j) &\approx \sum_{\substack{k=1 \\ k\neq j}}^J u^{j,1}_k \frac{x^k-x^j}{\|x^k-x^j\|}\\
     HV(x^j) &\approx \sum_{\substack{k=1 \\ k\neq j}}^J u_k^{j,2}\frac{(x^k-x^j)}{\|x^k-x^j\|}\otimes \frac{(x^k-x^j)}{\|x^k-x^j\|}
 \end{split}
 \end{equation}}
 for $u^{j,1}_k, u^{j,2}_k\in \R$ yet to be determined. Clearly, this means that our approximation for $\nabla $ will be an element of the affine space spanned by the ensemble, \rvv{and} $HV$ is now approximated by \rvv{a matrix with rank bounded by $\max\{J,d\}$}. We introduce an additional error slack term $\xi > 0$, which will be used for localization purposes, see \Cref{rem:xi}. 
 
 \rv{\begin{remark}
     We want to comment on the choice of our ansatz \eqref{eq:ansatz}; why do we parametrize gradient as a linear combination of the deviations $x^k-x^j$ and the Hessian via similar rank-1 terms instead of inferring the gradient and Hessian as a vector or matrix directly? There are several reasons for that:
     \begin{itemize}
         \item Our ansatz allows us to straightforwardly handle all orders of differentiability in the same way: It does not matter whether we just want to infer the gradient, or also the Hessian (and this generalizes straightforwardly to higher order); the structure of the linear equation \eqref{eq:invpr} only changes minimally. In contrast, if we were to infer the Hessian as a full matrix, we would have to be very careful with the indexing, and even more so for higher-order differential terms (which would become high-dimensional tensor objects that would need to be unpacked and correctly indexed).
         \item By considering a decomposition of the Hessian into rank-one terms, we only get $J$ additional terms instead of $\mathcal O(J^2)$, which would be the actual size of the Hessian matrix. Again, this generalizes to higher-order derivatives: If we wanted to include third order derivatives, we would need to include a mere $J$ further terms instead of $\mathcal O(J^3)$ which would quickly get out of hand. Of course, this simplification comes with a price, but one that has not been \rvv{too} high to pay in the experimental settings that we have tried so far.
         \item We are also interested in allowing the consideration of the under-determined case, i.e. $J \ll d$ (although the numerical experiments in this manuscript are mainly in the over-determined case). Our ansatz makes sure that we are not trying to infer differential information for which we don't have any data. For example, if $d=2$ and $J=2$ (this is still under-determined since we only ``see'' the one-dimensional affine subspace spanned by the ensemble), our ansatz explicitly approximates the gradient only on the one-dimensional affine subspace and does not postulate any components on the orthogonal component. This becomes even more pronounced for the Hessian, where we would need $J>d^2$ ensemble members to have a chance at recovering the full Hessian.
     \end{itemize}
 \end{remark}}
 
 Then \rv{\eqref{eq:zerothform} becomes}
 \begin{equation}\label{eq:firstform}
\begin{split}
       V(x^i) - V(x^j) &=\sum_{\rv{\substack{k=1 \\ k\neq j}}}^Ju_k^{j,1} \cdot \frac{(x^k-x^j)^T(x^i-x^j)}{\|x^k-x^j\|}\\ &+\frac{1}{2}\sum_{\rv{\substack{k=1 \\ k\neq j}}}^Ju_k^{j,2}\cdot \left[\frac{(x^k-x^j)^T(x^i-x^j)}{\|x^k-x^j\|}\right]^2 +  \eps_i^j \cdot  \left(\frac{\|x^i-x^j\|^3}{6} + \xi\right).
\end{split}
 \end{equation}

\rv{The role of $\xi$ is clarified in more detail in \Cref{rem:xi} below.}


\begin{remark}
Alternatively we could include mixed terms in the formulation of $HV(x^j)$, i.e. $HV(x^i) \approx \sum_{k,l}u_{k,l}^{j,2}\frac{(x^k-x^j)}{\|x^k-x^j\|}\otimes\frac{(x^l-x^j)}{\|x^l-x^j\|}$ for $u_{k,l}^{j,2}\in\R.$ Note that this comes at the cost that the number of coefficients scales with $J^2$ instead of $J$. For \rv{everyone's sanity}, we stick with the rank-$J$ ansatz for $HV(x^j)$. \rv{This extension is left as an exercise for the reader}  
\end{remark}

\rv{\begin{remark}
    There is some connection between Ensemble-based gradient inference and the idea of using Lagrange polynomials to perform polynomial regression (and thus the concept of $\Lambda$-poised sample sets). The latter has also been analysed in the context of derivative-free optimization, see \citep{powell2001lagrange,conn2008geometry,cartis2019derivative,ehrhardt2021inexact} and is again related to simplex gradients \citep{custodio2008using}. 
\end{remark}}

We now describe how to set up gradient inference as a linear inverse problem.

\subsection{Ensemble-based gradient inference as an inverse problem}

Let $\{x^i\}_{i=1}^J$ be an ensemble of points with pointwise evaluations $\{V(x^i)\}_{i=1}^J$ of an unknown function $V$. We fix $j\in\{1,\ldots, J\}$, and our goal is to infer $\nabla V(x^j)$ (and $HV(x^j)$) from the data. We set the following notational shorthands:
\begin{alignat*}{4}
    X^j&= (x^1-x^j, \ldots, x^i-x^j,\ldots, x^J-x^j)\in\R^{d \times \rv{(J-1)}},\qquad && X^j_{:,i} = x^i - x^j,\\
    Z^j&= \left(\frac{x^1-x^j}{\|x^1-x^j\|}, \ldots, \frac{x^J-x^j}{\|x^J-x^j\|}\right)\in\R^{d \times \rv{(J-1)}},\qquad && Z_{:,i}^j = \frac{x^i-x^j}{\|x^i-x^j\|}, \\
    y^j &= (V(x^1)-V(x^j),\ldots, V(x^J)-V(x^j))^T \in \R^{J-1}, \qquad&& y_i^j = V(x^i)-V(x^j).
\end{alignat*}
Note that the occurence of the term $J-1$ is due to the fact that we forego the entry where we would have $x^j - x^j$ in the enumeration. Then with $u^{j,i} = (u_k^{j,i})_{k=1}^J\in \R^{\rv{J-1}}$ \rv{(where $u_k^{j,1}$ and $u_k^{j,2}$ are the coefficients in the gradient and Hessian approximation, respectively)} and $u^j = (u^{j,1},u^{j,2})\in \R^{2\rv{(J-1)}}$, \rv{\eqref{eq:firstform} becomes}

\begin{align}\label{eq:invprlong}
     y^j&= \begin{bmatrix} {X^j}^TZ^j & \frac{({X^j}^TZ^j)^{\odot 2}}{2}\end{bmatrix}\cdot \begin{pmatrix}
   u^{j,1}\\ u^{j,2}
    \end{pmatrix} + \left(\frac{1}{3!}\mathrm{diag}(\{\|x^i-x^j\|^3\}_{i=1}^J)+\xi \cdot \rv{\mathrm{I}_{\rv{(J-1)}\times \rv{(J-1)}}}\right)\cdot \eps^j,
\end{align}
where the Hadamard square $({X^j}^TZ^j)^{\odot 2}$ is taken component-wise, i.e. $[({X^j}^TZ^j)^{\odot 2}]_{i,j} = [({X^j}^TZ^j)_{i,j}]^2$. We can further compact this form by setting $A^j = ({X^j}^TZ^j,  \frac{({X^j}^TZ^j)^{\odot 2}}{2})\in \R^{d\times (2\rv{(J-1)})}$ and $\Gamma_{\xi,\gamma}^j = \left(\frac{\gamma^2}{3!}\mathrm{diag}(\{\|x^i-x^j\|^3\}_{i=1}^J)+\xi\gamma^2\right)$. Then
\begin{equation}\label{eq:invpr}
    y^j = A^j u^j + \rv{\Gamma_{\xi,\gamma}^j}\eps^j
\end{equation}
with \rv{$\eps^j\sim \mathcal N(0,\mathrm{I})$}, is a linear inverse problem for the coefficient vector $u^j$. If ${\bar u}^j$ 
is the least squares solution of this inverse problem, then we can recover gradient information via
\begin{align}
    G^j_\xi(\{x^i\}_{i=1}^J)&:=Z^j\cdot {{\bar u}^{j,1}} = \sum_{\rv{\substack{k=1 \\ k\neq j}}}^J {\bar u}_k^{j,1} \frac{x^k-x^j}{\|x^k-x^j\|}\approx \nabla V(x^j)  \label{eq:lsqgradient}\\
    H^j_\xi(\{x^i\}_{i=1}^J) &:=  \sum_{\rv{\substack{k=1 \\ k\neq j}}}^J{\bar u}_k^{j,2}\frac{x^k-x^j}{\|x^k-x^j\|}\otimes \frac{x^k-x^j}{\|x^k-x^j\|} \approx HV(x^j) \label{eq:lsqhessian}
\end{align}
This notation hides the implicit dependence on $V$, and the specific way $u^j$ is obtained (although we assume this to be the least square solution to \eqref{eq:invpr} unless stated otherwise). 
\rv{\begin{remark}\label{rem:xi}
The motivation for the introduction of the error slack term $\xi$ is two-fold: First, if there are two ensemble members $x^i,x^j$ very close to each other, then this error term acts as a safeguard against unwanted overfitting. \rv{The general idea is this: If the potential $V$ is suspected to be of form $V_0 + V_\eps$, where $V_0$ is a general trend and $V_\eps$ exhibits undesirable (and less relevant) high-frequency local oscillations (e.g. from noisy data ), then $y_i-y^j$ is dominated by $V_\eps(x^i)-V_\eps(x^j)$. Since $x^i$ and $x^j$ are assumed to be close together, this would lead to a strong contribution of this unwanted oscillation when inverting \eqref{eq:invpr}, which is partly avoided by ``blowing up'' the right hand side by a minimum distance $\xi$}.  Second, if all ensemble members are very close to each other (in the magnitude of machine precision), then it buffers unwanted numerical instability. This error term can also be removed from the model by setting $\xi = 0$. On the other hand, letting $\xi\to\infty$ recovers the least-squares quadratic regression function through the data points, under the additional condition that the regression function passes through $(x^j,V(x^j))$. In a sense, $\xi=0$ corresponds to ``local'' gradient information, because it takes into account mostly nearby points, and $\xi\to\infty$ yields ``global'' gradient information. The effect of $\xi$ is illustrated in \Cref{fig:localapprox}. 
\end{remark}}

For brevity we write $G^j(\{x^i\}_{i=1}^J) := G^j_0(\{x^i\}_{i=1}^J)$ in the case $\xi=0$ and similar for the Hessian $H^j$.  Details of the inference are given in Algorithm~\ref{alg:grad}.
In the following we will refer to the case $\xi=0$ as \textit{local approximation} and $\xi>0$ \textit{global gradient approximation}. This characterization is motivated by the observation that $\xi$ enters the error term in \eqref{eq:invprlong}. Therefore, for $\xi=0$, the Taylor approximation in the location of a particle $x^i$ close to $x^j$ needs to be much better than in any particle far away from $x^j$. In the other extreme case $\xi \rightarrow \infty$ all particles enter with equal weight in the gradient approximation, hence the second-order polynomial approximation will be a ``global'' approximation instead of a local Taylor approximation. See \Cref{fig:localapprox} for a visualization of the effect of $\xi$.
\begin{remark}
The Hessian approximation can be written compactly in \texttt{Python} Einstein sum notation via $ H^j(\{x^i\}_{i=1}^J)= $~\texttt{np.einsum('k,lk,mk->lm', ${\bar u}^{j,2}$, $Z^j$, $Z^j$)}.  
\end{remark}

\begin{algorithm}[H]
\DontPrintSemicolon
\caption{Ensemble-based Gradient Inference \textbf{(EGI)}} \label{alg:grad}
\KwData{$\{x^i, V^i\}_{i=1}^J, \gamma>0, \xi\geq 0$, $j\in\N$}
\KwResult{$G= G^j_\xi(\{x^i\}_{i=1}^J), H=H^j_\xi(\{x^i\}_{i=1}^J)$ (approx. to $\nabla V(x^j)$ and $HV(x^j)$)}
    $X\gets (x^1-x^j,\ldots, x^J-x^j)\in \R^{d\times \rv{(J-1)}}$\;
    $Z\gets \begin{pmatrix} \frac{x^1-x^j}{\|x^1-x^j\|},&\cdots &,\frac{x^J-x^j}{\|x^J-x^j\|} \end{pmatrix}\in \R^{d\times \rv{(J-1)}}$\;
    $y\gets (V^1-V^j,\ldots, V^J-V^j)\in \R^\rv{(J-1)}$\;
    $A\gets (X^TZ, \tfrac{1}{2}(X^TZ)^{\odot 2})\in \R^{d\times (2\rv{(J-1)})}$\;
    $\Gamma \gets\gamma^2 \cdot \left[ \tfrac{1}{3!}\operatorname{diag}(\|x^i-x^j\|^3)_{i=1}^J + \xi\right]\in \R^{d\times d}$\;
    $(u^1,u^2)^T \gets \mathtt{lsq}(\Gamma^{-1}A, \Gamma^{-1}y)$\Comment{Solve $\Gamma^{-1}A (u^1,u^2)^T = \Gamma^{-1}y$ via Least-Squares}
    $G \gets  \sum_{k=1}^J {u}_k^{1} \frac{x^k-x^j}{\|x^k-x^j\|}$\;
    $H \gets  \sum_{k=1}^J { u}_k^{2} \frac{x^k-x^j}{\|x^k-x^j\|}\otimes  \frac{x^k-x^j}{\|x^k-x^j\|}$\;
\end{algorithm}

\rv{\begin{remark}\label{rem:extrap}Assume that $\gamma, \xi$ are fixed: If we need to compute approximations of $\nabla V$ and $HV$ not just in $x^j$, but in all members of the ensemble (as will be necessary for some of the applications to follow), there are always two options to consider:
\begin{enumerate}
    \item Full EGI on each ensemble member: Perform \Cref{alg:grad} on $\{x^i, V^i\}_{i=1}^J$ for all values $j\in\{1,\ldots,J\}$, so we obtain $\{G_\xi^j,H_\xi^j\}$ for $j\in\{1,\ldots,J\}$. 
    \item Extrapolate one single EGI on the rest of the ensemble: Perform \Cref{alg:grad} on $\{x^i, V^i\}_{i=1}^J$ for just one (suitably chosen) $j\in\{1,\ldots,J\}$, so we obtain $\{G_\xi^j,H_\xi^j\}$. Then another tangent approximation gives us a way of approximating $\nabla V(x^i)$ even for $i\neq j$:
    \[ \nabla V(x^i) \approx \nabla V(x^j) + HV(x^j)\cdot (x^i-x^j) \approx G^j_\xi + H_\xi^j \cdot (x^i - x^j).\]
    The Hessian in $x^i$ can then be approximated by $H_\xi^j$ (implicitly assuming quadraticity of $V$), or inferred similarly to $\nabla V$ from even higher differential information in $x^j$. This approach is computationally favorable (since we do not have to solve more than one linear equation) but comes at the expense of further approximation error.
\end{enumerate}
Note that the second option allows us to see how EGI is a full generalization of the concept of polynomial regression: By choosing an arbitrary reference point $x^j$ and setting $\xi = \infty$ (or, practically, sufficiently large), and performing the extrapolation variant outlined above, we get exactly the ``global'' quadratic polynomial which is the least-squares solution to the linear regression problem of fitting $\{x^i, V^i\}$. This is demonstrated in \Cref{fig:localapprox}(d) (note that the quadratic regression is visually anchored to the points $(x^i, V^i)$ for compatibility with the other experiments, but the shape of the quadratic function is exactly the quadratic linear regression function). \end{remark}}

\subsection{Bayesian Ensemble-based Gradient Inference}
The least-squares approach can without major changes be turned into a Bayesian sampling algorithm for gradient and Hessian information by solving the linear inverse problem \eqref{eq:invpr} via the Gaussian update formula \rv{ instead of using a least squares approach: We choose a covariance matrix $\Sigma \in \R^{2n\times 2n}$, and assume a Gaussian prior measure on the coefficients $(u^{1,n},u^{2,n})\sim \mathcal N(0,\Sigma).$\footnote{Of course we can also choose a more complicated prior than Gaussian, but then we might not have an explicit formula for the posterior.} Now we interpret \eqref{eq:invpr} (recorded again for convenience) 
\begin{equation*}
    y^j = A^j u^j + \rv{\Gamma_{\xi,\gamma}^j}\eps^j
\end{equation*}
as a linear Bayesian inverse problem for the coefficients $u^j$. Since the map $A^j$ is linear, and the prior is Gaussian, we know that the posterior measure on $u^j$ is Gaussian, as well, and its form $\mathcal N(\hat \mu,\hat \Sigma)$ can be explicitly computed. By sampling from the posterior we obtain coefficients $u^{j,n}$ which are to be understood as coefficients in the decomposition of the approximated gradient and Hessian. This procedure is described in \Cref{alg:gradBayes}. We just remark that any subsequent algorithm using the deterministic least squares approximation \Cref{alg:grad} can be modified to use Bayesian samples of the approximated gradient (\Cref{alg:gradBayes}).
}

\begin{algorithm}[H]
\DontPrintSemicolon
\caption{Bayesian Ensemble-based Gradient Inference}\label{alg:gradBayes}
\KwData{$\{x^i, V^i\}_{i=1}^J, \gamma>0, \xi\geq 0$, index $j\in\N$, $\Sigma > 0$,  number of samples $N\in\N$}
\KwResult{samples $\{G^{n,j}\}_{n=1}^N$, $\{H^{n,j}\}_{n=1}^N$ (approximations to $\nabla V(x^j)$ and $HV(x^j)$) and Bayesian maximum-a-posteriori estimators $G_\text{MAP},H_\text{MAP}$.}
    \rv{\texttt{lines 1--5 of \Cref{alg:grad}}\;}
    $K\gets \Sigma {A^j}^T(\Gamma + {A^j}\Sigma {A^j}^T)^{-1}$\;
    $\hat \mu \gets (\hat\mu^1,\hat\mu^2)\gets Ky$ \;
    $\hat \Sigma\gets \Sigma - K{A^j}\Sigma$\;
    $G_\text{MAP} \gets  \sum_{k=1}^J {\hat\mu}^1_k \frac{x^k-x^j}{\|x^k-x^j\|}$\;
    $H_\text{MAP} \gets  \sum_{k=1}^J {\hat\mu}^2_k \frac{x^k-x^j}{\|x^k-x^j\|}\otimes  \frac{x^k-x^j}{\|x^k-x^j\|}$\; 
    \For{$n\gets1$ \KwTo $N$ }{
        $(u^{1,n},u^{2,n})^T \sim \mathcal N(\hat\mu, \hat \Sigma)$ \Comment{sample from posterior}
        $G^{n,j} \gets  \sum_{k=1}^J u^{1,n}_k \frac{x^k-x^j}{\|x^k-x^j\|}$\;
        $H^{n,j} \gets  \sum_{k=1}^J u^{2,n}_k \frac{x^k-x^j}{\|x^k-x^j\|}\otimes  \frac{x^k-x^j}{\|x^k-x^j\|}$\;
    }
\end{algorithm}

\begin{remark}
    The numerical complexity of \Cref{alg:grad,alg:gradBayes} is dominated by the cost of solving a linear equation of size $J\times J$, since we assume pointwise evaluation of $V(x^i)$ to be available initially.
\end{remark}

\rv{\begin{remark}
    We will use mainly the gradient approximations -- with the exception of \Cref{alg:EGI-ALDI-extra}, which also uses Hessian approximations. This is mainly for the reason that there are more first-order algorithms that can be modified to use approximated gradients instead of exact gradients; also vanilla CBO does not use gradient information at all, so it makes sense to first look at improvements generated by leveraging approximated gradients. It is entirely conceivable to modify many more algorithms to employ second-order (or higher-order) differential approximations as well, but this would go beyond the envisioned scope of this manuscript which is mainly to bring together the idea of inferring differential information with sampling and optimization algorithms.
\end{remark}}

\subsection{Examples}
\begin{figure}
    \centering
     \begin{subfigure}[t]{0.49\textwidth}
    \includegraphics[width=\textwidth]{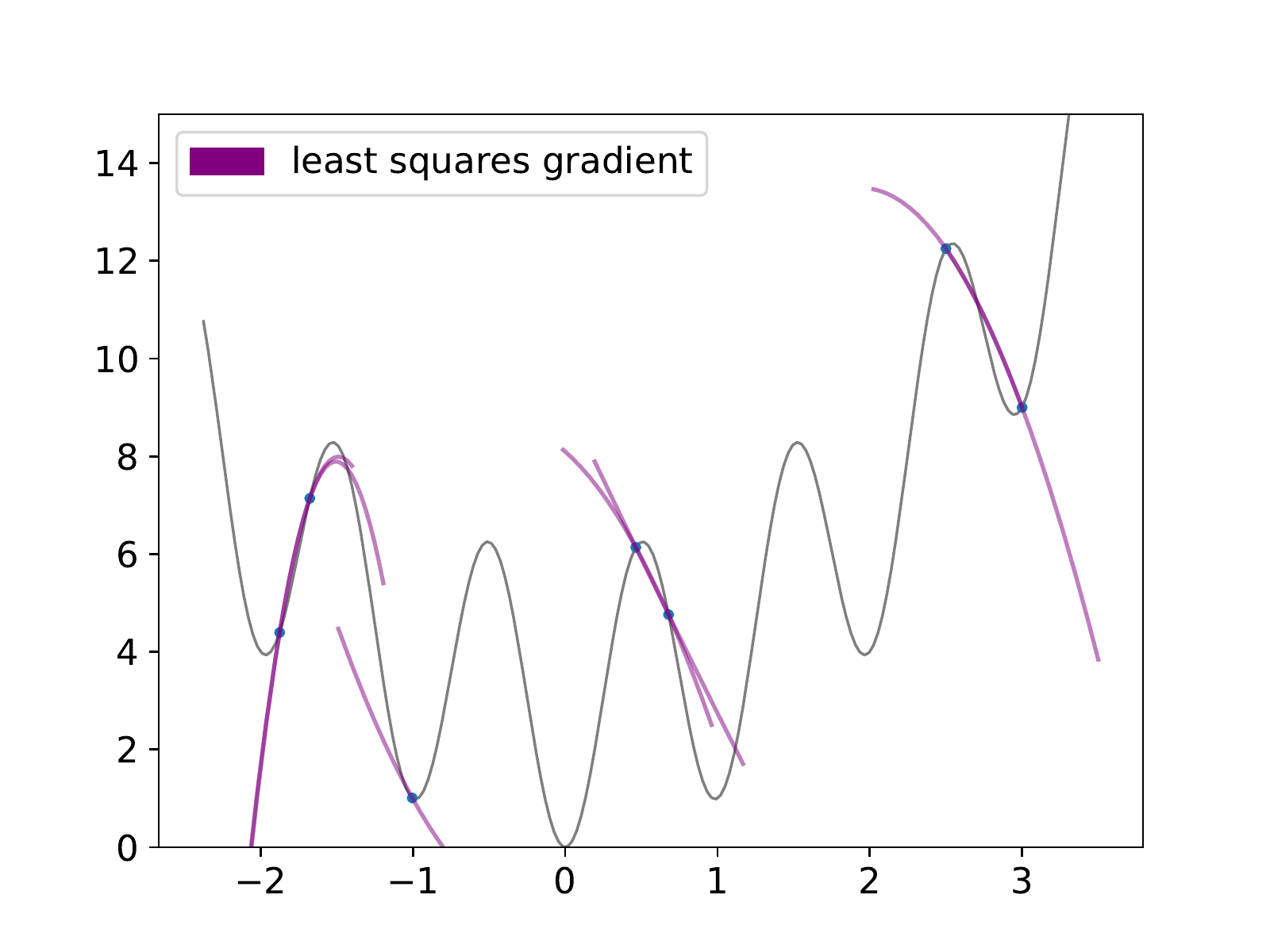}
    \caption{Local approximation ($\xi=0$) in each ensemble member}
     \end{subfigure}
     \hfill
     \begin{subfigure}[t]{0.49\textwidth}
    \includegraphics[width=\textwidth]{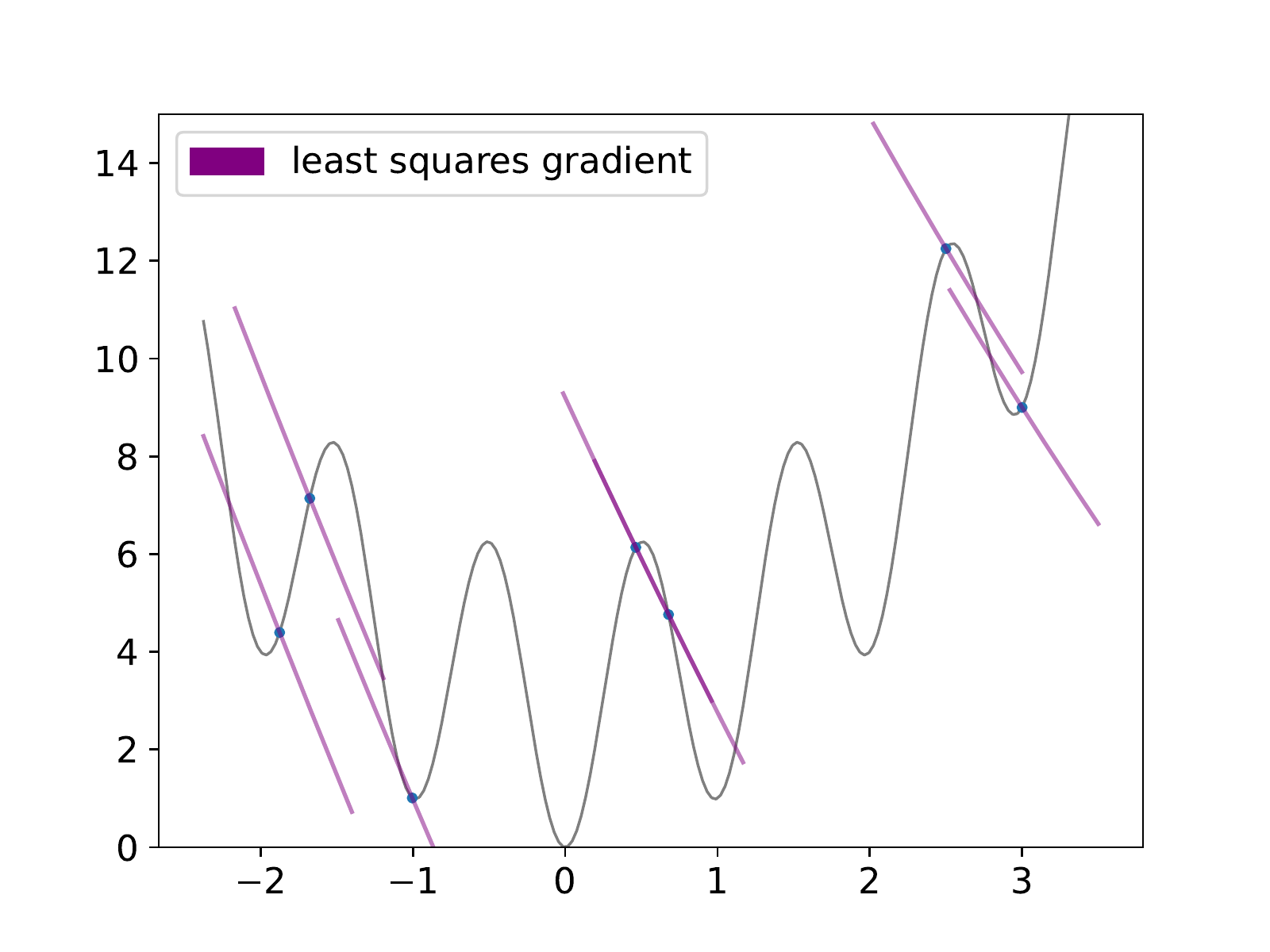}
    \caption{\rv{Local approximation ($\xi=0$), extrapolated from member at $x=0.679$}}
     \end{subfigure}
     \begin{subfigure}[t]{0.49\textwidth}
    \includegraphics[width=\textwidth]{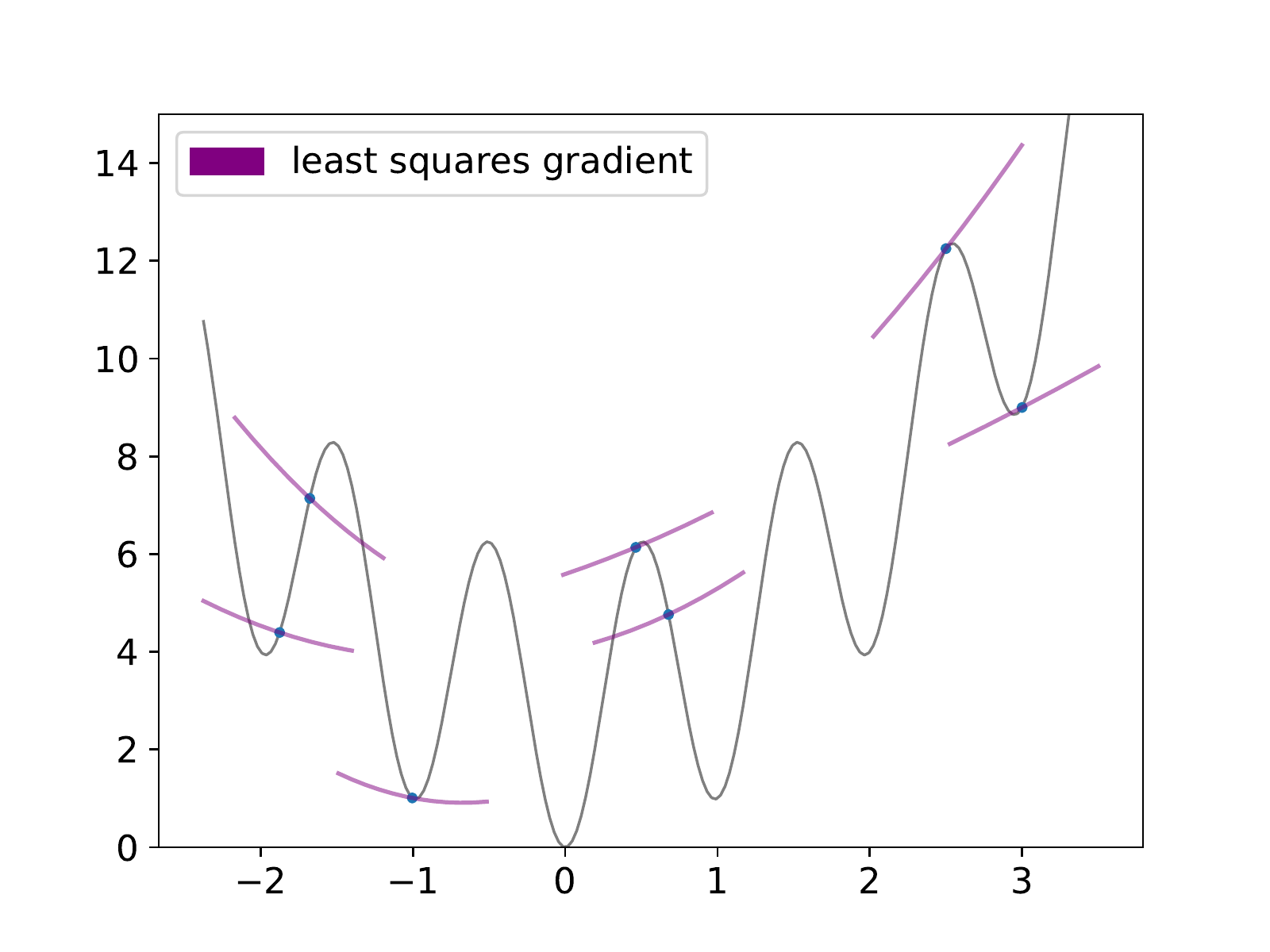}
    \caption{Global approximation ($\xi=1000$) in each ensemble member}
     \end{subfigure}
     \hfill
     \begin{subfigure}[t]{0.49\textwidth}
    \includegraphics[width=\textwidth]{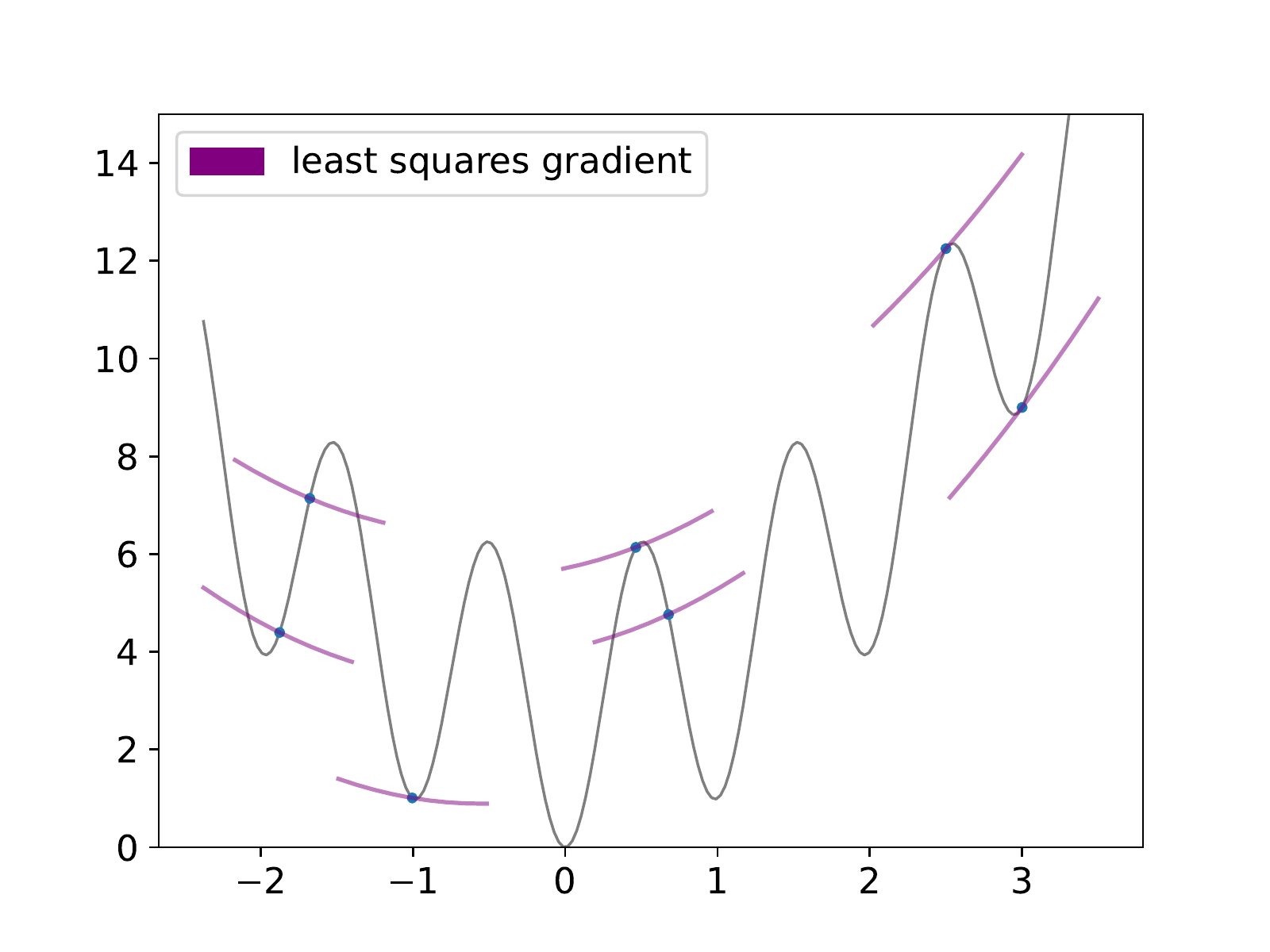}
    \caption{\rv{Global approximation ($\xi=1000$), extrapolated from member at $x=0.679$}}
     \end{subfigure}
     \begin{subfigure}[t]{0.49\textwidth}
    \includegraphics[width=\textwidth]{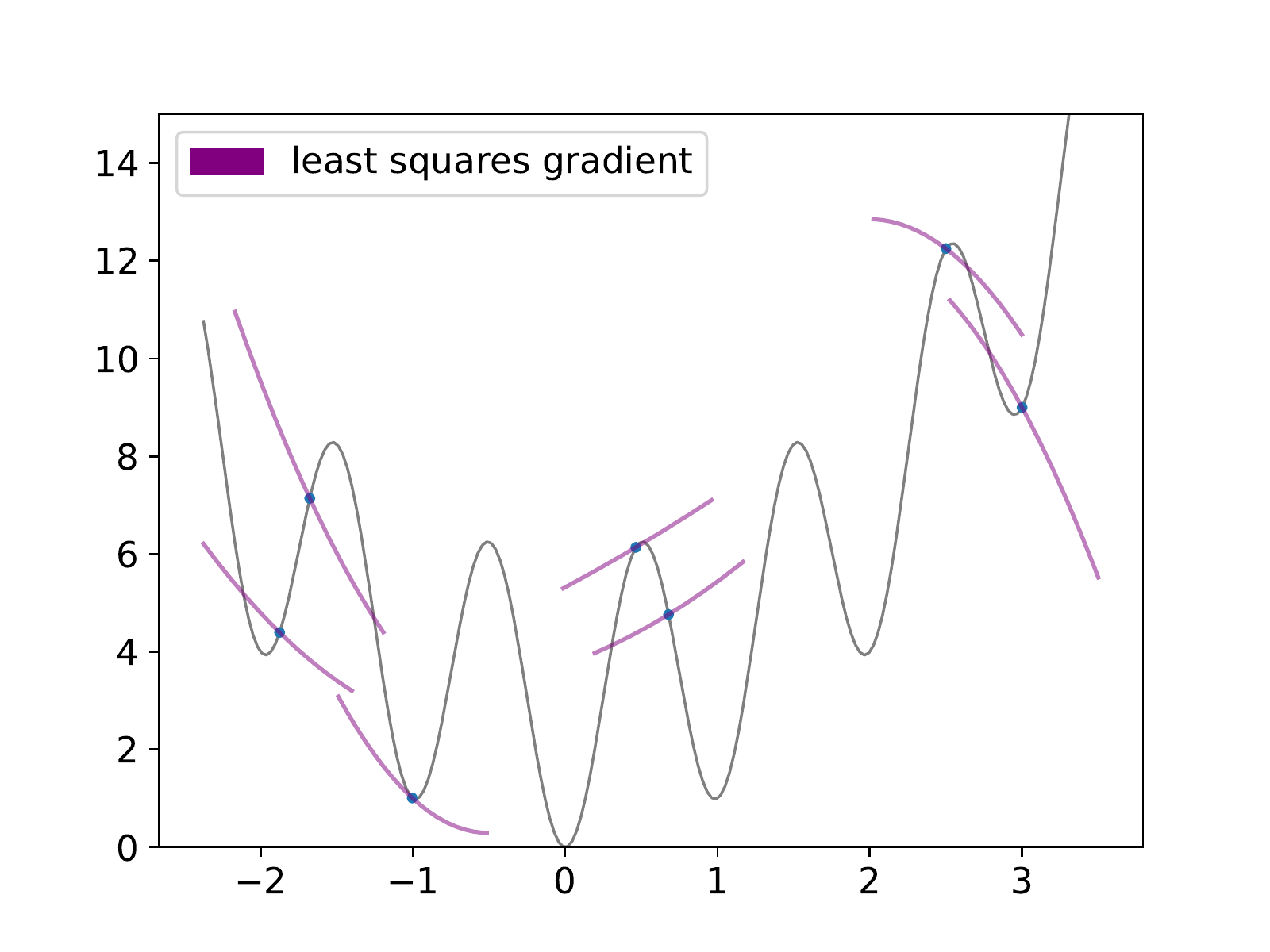}
    \caption{Intermediate range approximation ($\xi=1$) in each ensemble member}
     \end{subfigure}\hfill
     \begin{subfigure}[t]{0.49\textwidth}
    \includegraphics[width=\textwidth]{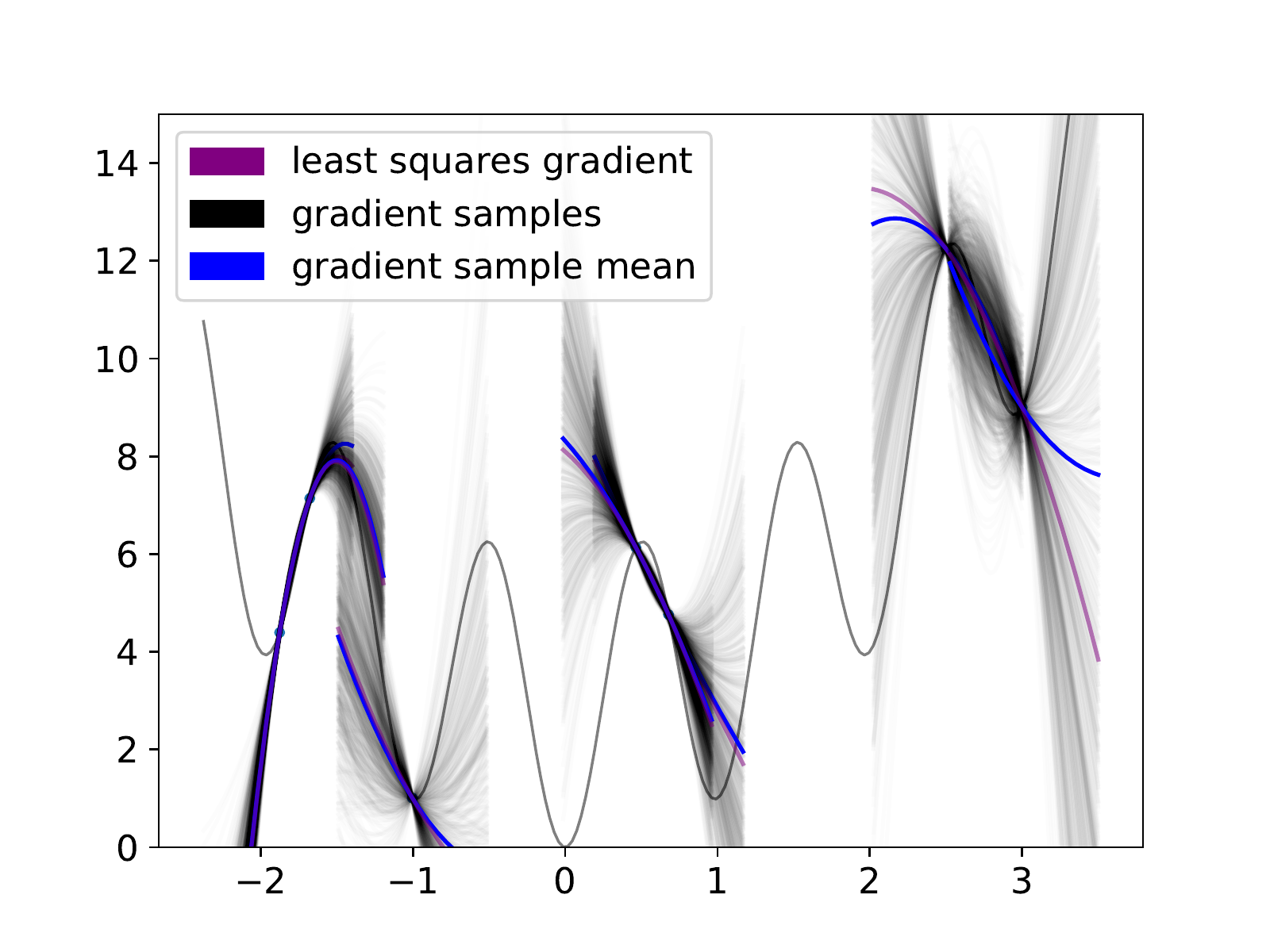}
    \caption{Local approximation ($\xi=0$) in each member, sampled quadratic approximations. }
     \end{subfigure}
    \caption{Approximation of (a variant of) the Rastrigin function $x\mapsto x^2+3(1-\cos(2\pi x))$ via finite evaluations with various choices of $\xi$. Code: \texttt{1d\_gradinf.py}}
    \label{fig:localapprox}
\end{figure}

We illustrate the proposed approximation of gradient and Hessian with the help of two well-known benchmarks for optimization.
\paragraph{Rastrigin function in 1d}
We set $V:\R\to\R$ with $V(x) = x^2 + 3\cdot(1-\cos(2\pi x))$, which is (a slightly differently scaled variant of) the Rastrigin function\rv{, and $J=7$}. Figure \ref{fig:localapprox} demonstrates how we can locally approximate a function from given pointwise evaluations\rv{, and shows the effect of the value of $\xi$ (locality) and whether we perform EGI on each ensemble member or extrapolate from a single position.} $\xi=0$ corresponds to local Taylor approximation while the case $\xi \gg 1$ approaches a least-squares quadratic fit through all data points (albeit always going through the current reference point $(x^j, V(x^j))$), \rv{which corresponds to $\xi \to \infty$ and extrapolated gradient approximation (i.e. item 2 in \Cref{rem:extrap}). We also show the (slightly nonsensical) result of choosing $\xi = 0$ and extrapolating the gradient: This means that gradient information is collected strongly locally (since $\xi = 0$), but then this gradient is extrapolated to all of $V$, which is rarely a good fit. The figure on the bottom right }also shows local quadratic approximation samples obtained by interpreting \eqref{eq:invpr} as a Bayesian inverse problem (after defining a prior on the coefficients $ u^j$), taking i.i.d. samples $\hat{ u}^j$ from the posterior.

\begin{figure}
    \centering
    \begin{subfigure}[t]{0.49\textwidth}
    \includegraphics[width=\textwidth]{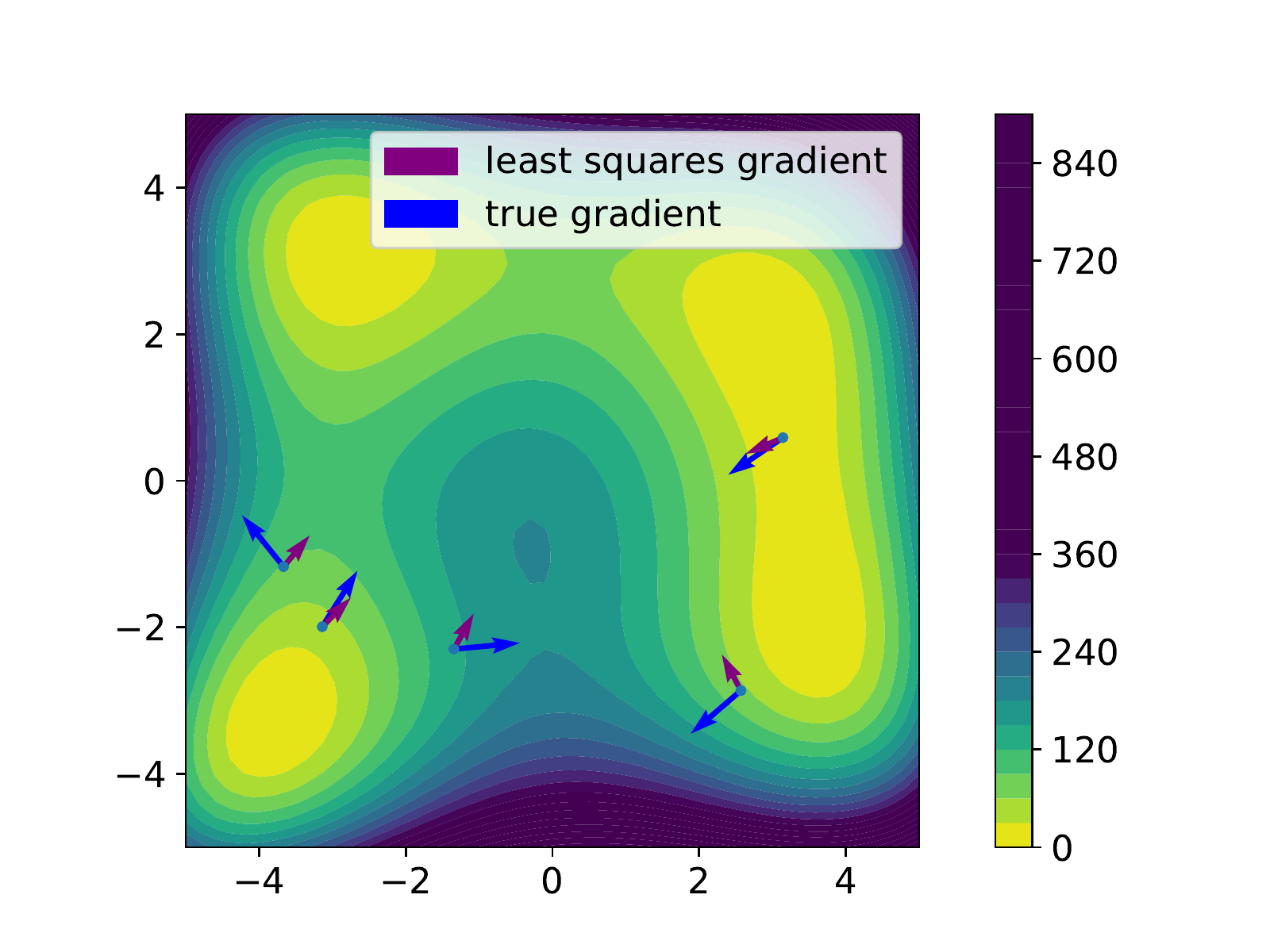}
    \caption{$J=5$, $\xi=0$}
    \end{subfigure}
    \begin{subfigure}[t]{0.49\textwidth}
    \includegraphics[width=\textwidth]{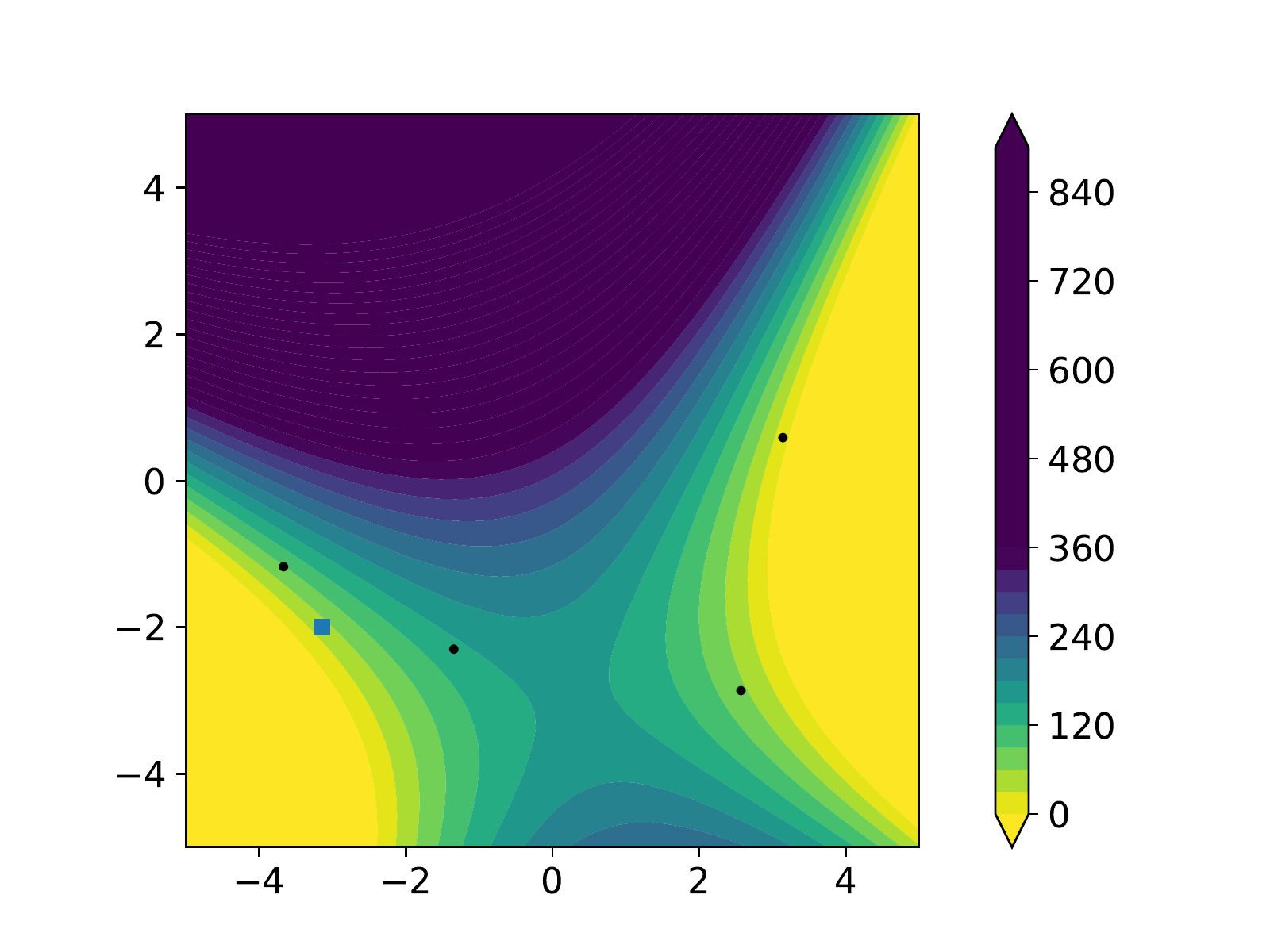}
    \caption{quadratic approximation centered at $x^j$ (marked with square).}
    \end{subfigure}
    
    \begin{subfigure}[t]{0.49\textwidth}
    \includegraphics[width=\textwidth]{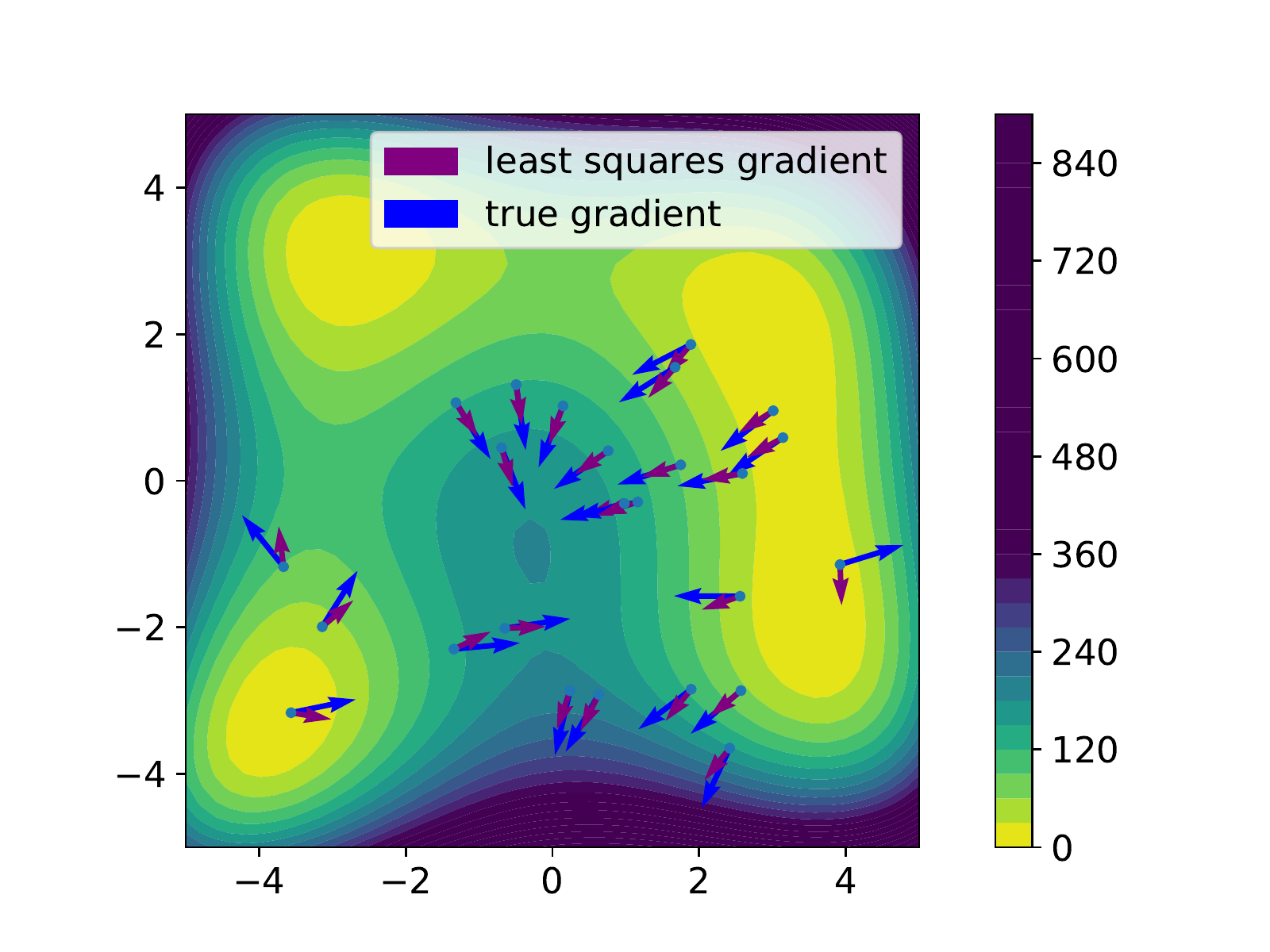}
    \caption{$J=25$, $\xi=0$}
    \end{subfigure}
    \begin{subfigure}[t]{0.49\textwidth}
    \includegraphics[width=\textwidth]{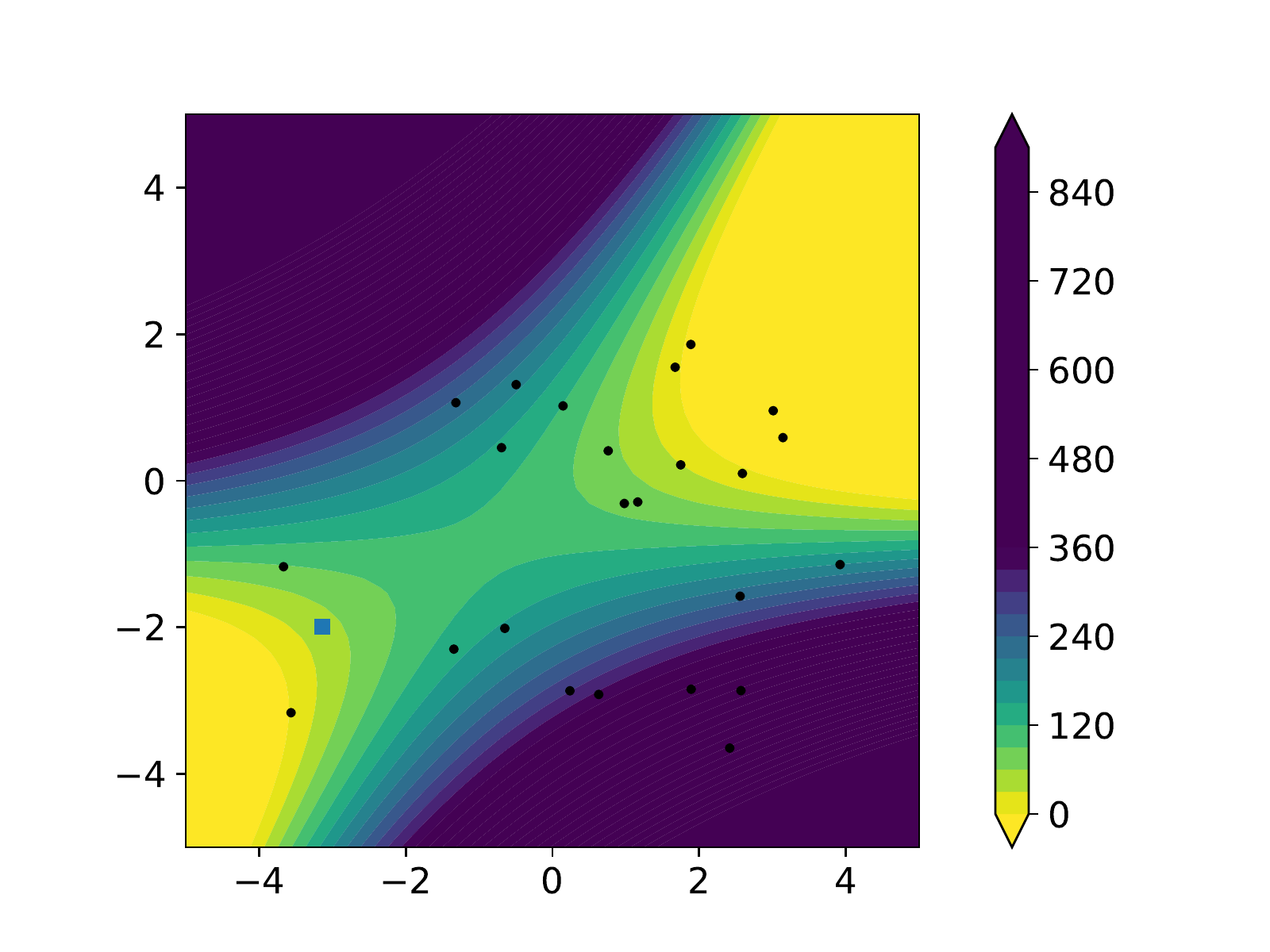}
    \caption{quadratic approximation centered at $x^j$ (marked with square).}
    \end{subfigure}
    
    \begin{subfigure}[t]{0.49\textwidth}
    \includegraphics[width=\textwidth]{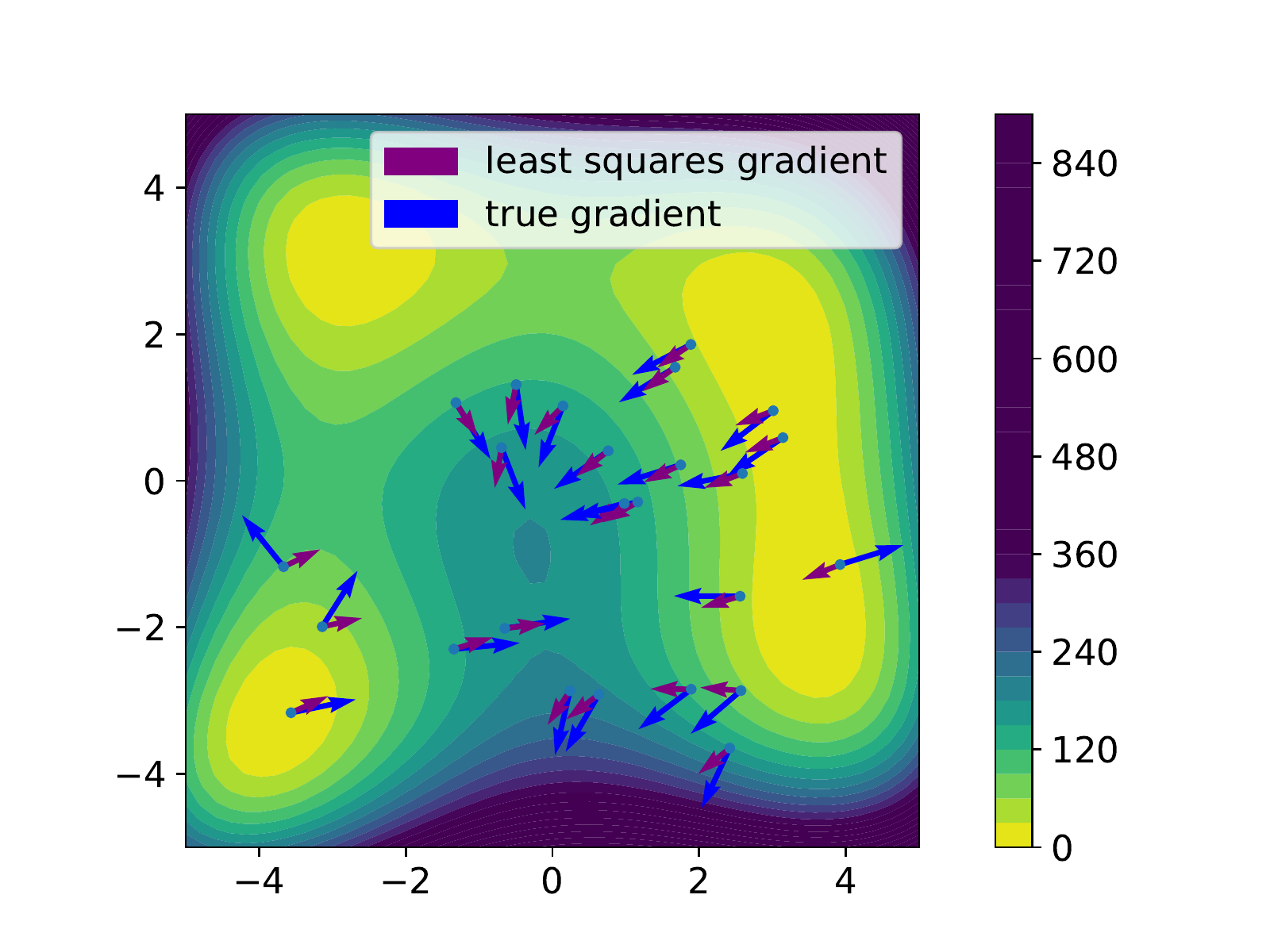}
    \caption{$J=25$, $\xi=1000$}
    \end{subfigure}
    \begin{subfigure}[t]{0.49\textwidth}
    \includegraphics[width=\textwidth]{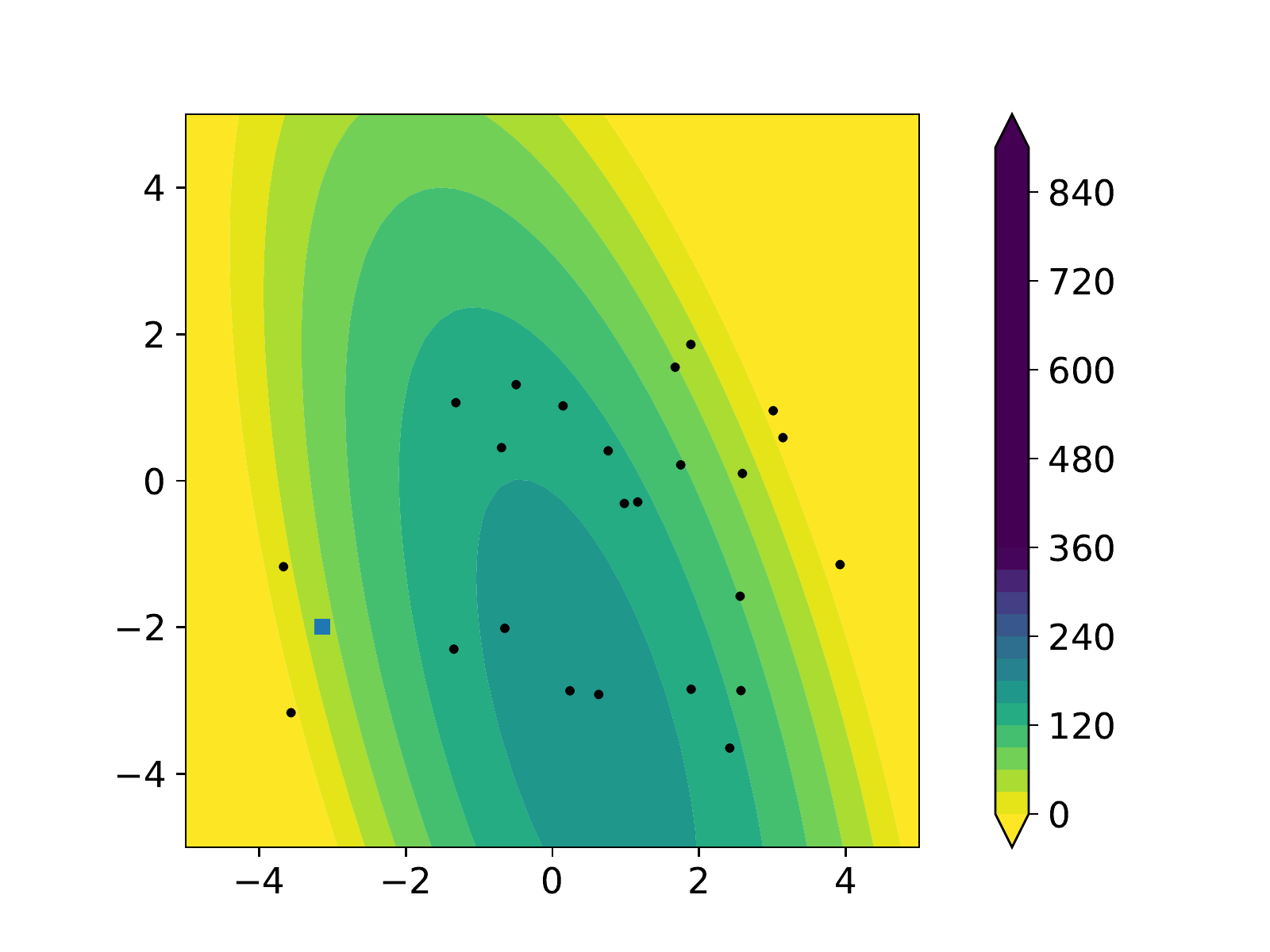}
    \caption{quadratic approximation centered at $x^j$ (marked with square).}
    \end{subfigure}
    
    \caption{Approximation of Himmelblau function by pointwise evaluation. Left column shows approximated gradients $G^j(\{x^i\},V)$ in comparison to actual gradients (vectors are rescaled to unit length). Right columns show quadratic function obtained by extending local quadratic inferred approximation to whole domain. Code: \texttt{2d\_gradinf\_J5\_xi0.py, 2d\_gradinf\_J25\_xi0.py, 2d\_gradinf\_J25\_xi1000.py}} 
    \label{fig:gradinf2d}
\end{figure}

\paragraph{Himmelblau function in 2d}
As $V$ we choose the Himmelblau function $V:(x,y)\mapsto (x^2+y-11)^2 +(x+y^2-7)^2$ on $\R^2$ and proceed along the same lines as for the Rastrigin function. Figure \ref{fig:gradinf2d} shows the influence of $J$ and $\xi$. The column on the right-hand side shows the level sets of \begin{equation}\label{eq:quadraticapprox}
 x\mapsto V(x^j) + \left(G^j(\{x^i\}_{i=1}^J)\right)^T(x-x^j) + \frac12(x-x^j)^T\cdot H^j(\{x^i\}_{i=1}^J)\cdot (x-x^j)
\end{equation}
for an arbitrarily chosen ``reference particle'' $x^j$. 

\begin{figure}
    \centering
    \begin{subfigure}[t]{0.49\textwidth}
    \includegraphics[width=\textwidth]{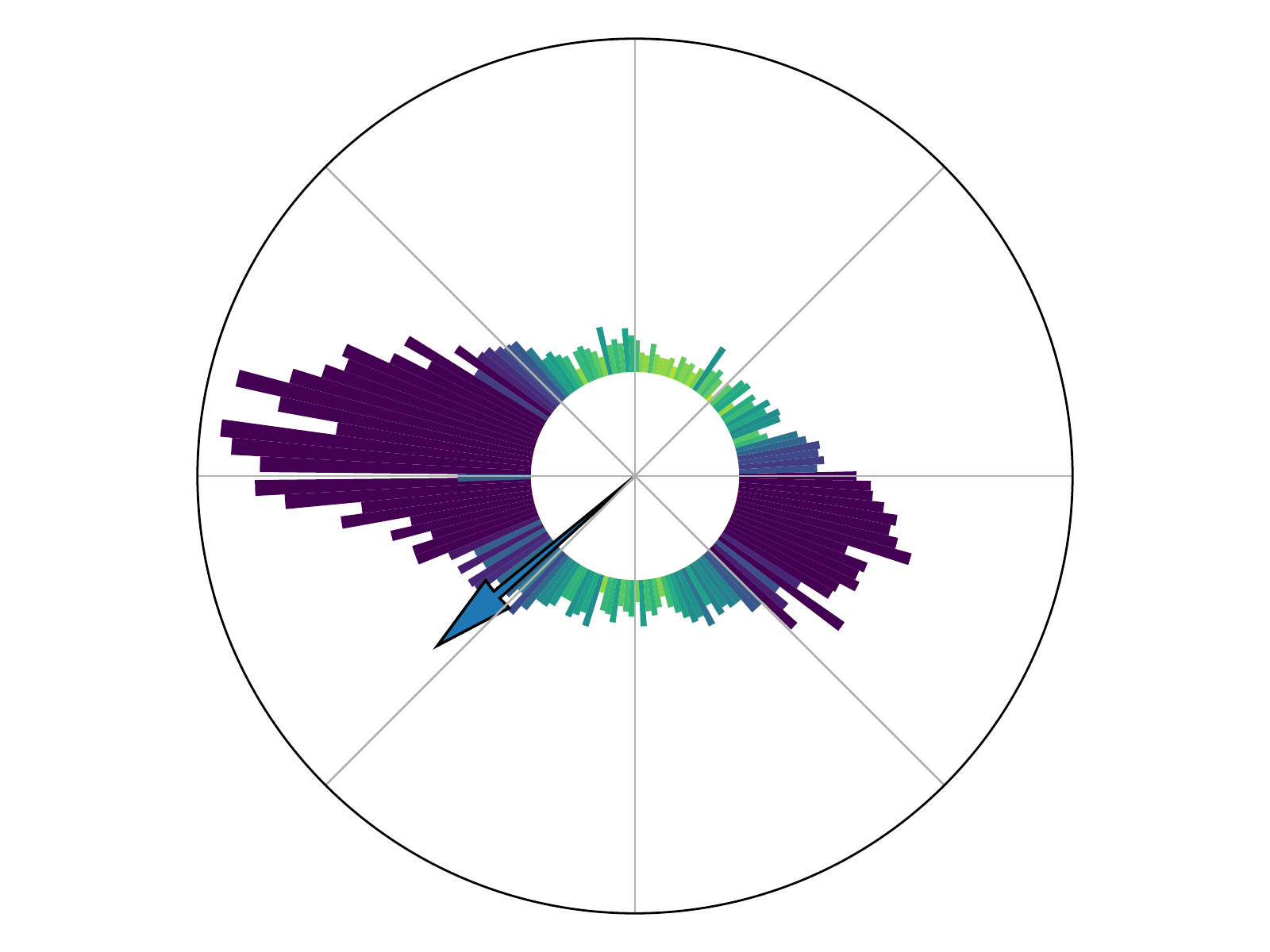}
    \caption{$J=5$}
    \end{subfigure}
    \begin{subfigure}[t]{0.49\textwidth}
    \includegraphics[width=\textwidth]{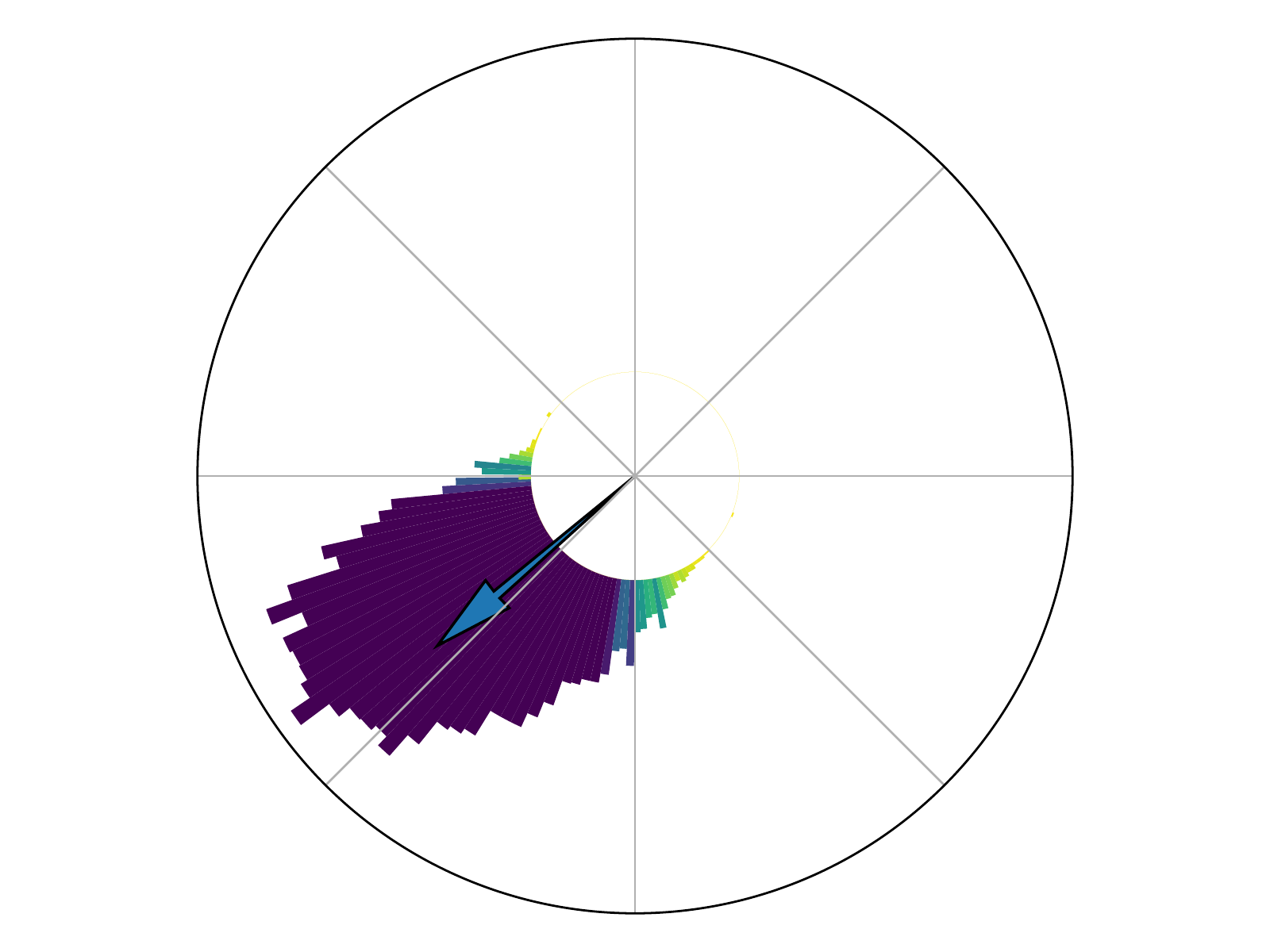}
    \caption{$J=25$}
    \end{subfigure}
    \caption{Posterior gradient samples after choosing a prior on the coefficients $ u^j$: Radial histogram of gradient (direction) samples, evaluated at a specific ensemble member, with true gradient (blue arrow) for reference. Clearly, increasing the number of ensemble members $J$ contracts the Bayesian posterior around the ground truth gradient vector. Note that magnitude of any gradient vectors is disregarded in this illustration. Code: \texttt{2d\_bayes\_gradinf.py}}
    \label{fig:bayesian_EGI}
\end{figure}

The left column of \Cref{fig:gradinf2d} shows (for an ensemble $(x^j,V(x^j))_j$) a comparison between true gradients $\nabla V(x^j)$ and the least squares approximations obtained from \Cref{alg:grad}. Note that the arrows are scaled to the same magnitude, so only the angle between ground truth and inferred gradient is a relevant measure of approximation quality. Note that taking a larger ensemble improves the local gradient approximation, as to be expected. The right column shows level sets of the quadratic function in \Cref{eq:quadraticapprox}.  For $\xi=0$, the quadratic approximation to $V$ obtained from (inferred) gradient and Hessian works well only locally, while the global approximation $\xi = 1000$ shows deficits when judging actual gradient approximation (as can be seen by comparing the arrows in the figures belonging to $J=25, \xi=0$ with $J=25, \xi=1000$). This is a similar phenomenon as shown in \Cref{fig:localapprox}. Finally, \Cref{fig:bayesian_EGI} shows an illustration of \Cref{alg:gradBayes}. It shows how the true gradient $\nabla V(x^i)$ in a specific ensemble member $x^i$ is approximated. In the smaller ensemble of size $J=5$, the posterior gradient samples predominantly point along a direction different to the true gradient, but the latter can be seen to be in the support of this measure. By increasing the ensemble size to $J=25$ and thereby providing more information, it can be seen that the support of the posterior measure aligns with and contracts on the true gradient.

\medskip

We now describe a series of examples of optimization and sampling algorithms which can be augmented with the gradient approximation presented above.

\section{Consensus-based optimization augmented by EGI}\label{sec:CBO}
\rv{In this section we consider rather general optimization problems of form \begin{equation}
    \min_x V(x).
\end{equation}
Any mention of $G_\xi^j$ will be the result of \Cref{alg:grad} as applied to this function $V$. }

{For optimization problems with unknown structure it is advantageous to have gradient-free optimization algorithms that allow for treating the objective as black-box. Therefore the original formulation of CBO does not use any gradient information. With the gradient inference idea discussed above, we are able to keep the gradient-free property of CBO and still augment the algorithm with higher order differential information. 

\subsection{EGI-CBO}}
Let $V:\R^d\to[0,\infty)$ be a (possibly non-convex) function, $\beta>0$ an inverse heat parameter. We propose the following gradient-EGI-CBO method describing the dynamics of the particles via
\begin{align}
    \d x^i = -\kappa G^0_\xi(\{\wmean(\rho), x^1,\ldots, x^J\})\d t - \lambda (x^i - \wmean)\d t + \sigma|x^i - \wmean| \d W_t^i
\end{align}
 with initial conditions $x_0^i \sim \mathcal P_2(\R^d)$ drawn independently. Note that the ensemble is augmented by the weighted mean and we take the least square gradient and Hessian approximation centered at the weighted mean and use this gradient approximation for all particles. The full algorithm for this method -- EGI-CBO -- can be found in \Cref{alg:augCBO}.

The difference to CBO is the additional (approximated) gradient term. The motivation for incorporating this is threefold: (1) Assuming that the spatial resolution of the ensemble is coarser than the distance of individual minima and the ensemble is still spread out, the gradient term can help to jump over local minima. (2) The gradient term converges to the true gradient projected onto the affine subspace spanned by the ensemble.\footnote{\rv{This is due to the fact that a collapsing ensemble renders the EGI method into a finite difference approximation on the spanned subspace. }} This may facilitate the convergence to the true global minimizer. Indeed, by the Laplace principle and the quantitative non-asymptotic Laplace principle \rvv{\citep{fornaiser2021globally}} the approximation quality of CBO strongly depends on the temperature parameter. Additional gradient information is expected to improve the approximation quality, and drive the weighted mean towards the true minimizer. (3) In the later phase of the dynamics, the additional gradient term helps with accelerated collapse around the actual position of the (possibly local) minimum.
This alleviates an issue of CBO where the ensemble does indeed collapse in a vicinity of a (possibly local) minimum $x^\star$, but with consensus not necessarily converging to the actual position of $x^\star$. Both features, ``convergence to better minima'' and ``better convergence to minima'', can be seen from the following numerical experiments.

\subsection{Numerical examples: EGI-CBO}

For simplicity, we use the least squares version (\Cref{alg:grad}) for the gradient approximation for the following results. Note that a sampling approach via \Cref{alg:gradBayes} is feasible as well and might be preferred in Machine Learning applications as the behaviour resembles stochastic gradient descent methods.

\rv{For clarity of presentation, each algorithm from now on comments on the amount of additional linear systems that need to be solved (as compared to the vanilla version of the algorithm). For example, EGI-CBO (\Cref{alg:augCBO}) needs to solve $J$ additional linear systems per iteration as compared to CBO.}

\begin{algorithm}[H]
\DontPrintSemicolon
\caption{EGI-CBO}\label{alg:augCBO}
\Comment{additional linear equations: $J$} 
\KwData{$N \in \N, \{x_0^j\}_{j=1}^J, \alpha, \lambda, \sigma, \kappa, \xi, \tau\geq 0$, \texttt{noise}$\,\in\{\text{norm-proportional},\text{component-wise}\}$, \rv{\texttt{extrapolate}$\,\in\{\texttt{True},\texttt{False}\}$}}
\KwResult{minimizer $\wmean$ of \rv{$V$}}
\For{$n\gets0$ \KwTo $N-1$ }{
    $m_n^\alpha \gets \frac{\sum_{j=1}^J \exp(-\alpha \rv{V}(x_n^j))\cdot x_n^j}{\sum_{j=1}^J \exp(-\alpha \rv{V}(x_n^j))}$ \Comment{Weighted mean -- use \texttt{logsumexp} to avoid underflow}
    $m_n\gets \frac{1}{J}\sum_{j=1}^J x_n^j$ \Comment{Unweighted mean}
    \For{$j\gets1$ \KwTo $J$ }{
    \label{line:gradientCBO}$g_n \gets G^0_{\rv{\xi}}(\{m_n\}\cup\{x_n^i\}_{i=1}^J)$\;
    \label{line:hessCBO}$H_n \gets H^0_{\rv{\xi}}(\{m_n\}\cup\{x_n^i\}_{i=1}^J)$
    \Comment{Gradient and Hessian approximation (least squares) at unweighted mean, via \Cref{alg:grad}}
    $W_n^j\sim \mathcal N(0,I^d)$\;
    \rv{\uIf{\rm\texttt{extrapolate} is \texttt{True}}{
    $g_n^j \gets g_n + H_n\cdot (x_n^j - m_n)$ \;
    }
    \Else{
    $g_n^j \gets g_n$ \;
    }}
    \Switch{\rm\texttt{noise}}{
            \Case{norm-proportional}{
                $\sigma_n^j\gets \sigma \|x_n^j-m_n^\alpha\| W_n^j$
            }
            \Case{component-wise}{
                $\sigma_n^j\gets \sigma (x_n^j-m_n^\alpha)\odot W_n^j$
            }
        }
    $x_{n+1}^j \gets x_n^j -\tau \kappa \rv{g_n^j} - \tau \lambda (x
    _n^j- m_n^\alpha) +\sqrt{\tau}\sigma^j_n$ \Comment{Euler-Maruyama step}
    }
        
    }
\end{algorithm}

\begin{remark}
    If we choose $\lambda = 0$ and $\sigma=0$ in \Cref{alg:augCBO}, then this corresponds to a kind of ensemble-based gradient descent method. Unfortunately this seems to perform quite badly in practice: When the ensemble moves to a position such that its least squares approximation has a vanishing gradient (and this happened frequently in our numerical experiments), then the iteration essentially stops, without approaching the minimizer any further. For example, imagine $V(x) = x^2$, and ensemble of size $J=2$ at positions $-1$ and $+1$. Then the estimated gradient will be $0$, i.e. the ensemble will not move any further (although this issue can be alleviated by extrapolating gradients via second-order differential information, as done in \Cref{alg:EGI-ALDI-extra}). The impact of contraction and stochastic exploration means that this cannot happen easily with EGI-CBO.
\end{remark}


\paragraph{Rastrigin function in 2d} 
We aim to minimize the two-dimensional Rastrigin function $V(x_1,x_2) = 2\cdot 10 + x_1^2 - 10 \cos(2\pi x_1) + y_1^2 - 10\cos(2\pi y_1)$. We start with an ensemble of $J=4$ particles, which is a moderately large ensemble in two dimensions, and we also consider the case of $J=2$, which yields an ensemble spanning an affine subspace of dimension 1 in each iteration. We further set $\alpha=100$, $\lambda=1.5$, $\sigma=0.7$, $\kappa = 0.5$, $\xi=0$, $\tau=0.01$ and $N=1000$ iterations. Note that we choose the initial ensemble uniform in the set $[-4,-1]^2$, which does not contain the global minimum. As discussed in \cite{kalise2022consensus}, this is a much harder case where only few studies of the CBO method exist.

\begin{figure}
     \centering
     \begin{subfigure}[t]{0.45\textwidth}
         \centering
\includegraphics[trim=10 20 50 100, clip,width=\textwidth]{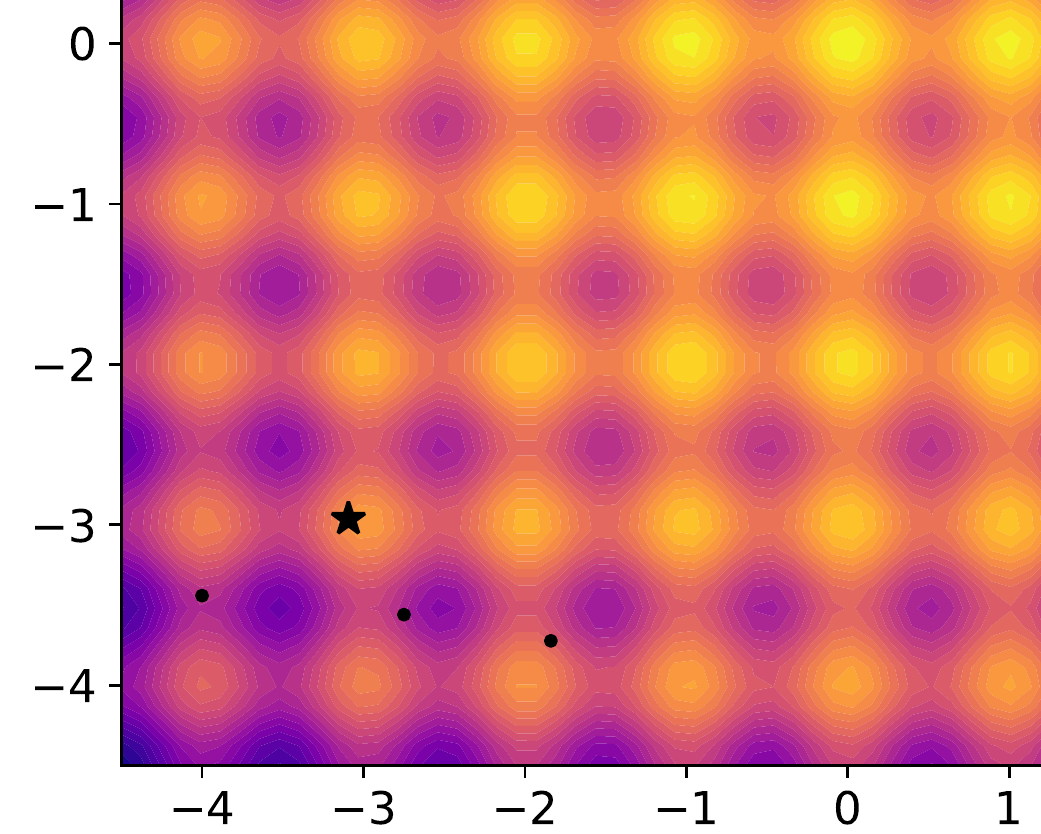}
         \caption{Initial ensemble, $J=4$}
     \end{subfigure}\hfill
      \begin{subfigure}[t]{0.45\textwidth}
         \centering
\includegraphics[trim=10 20 50 100, clip,width=\textwidth]{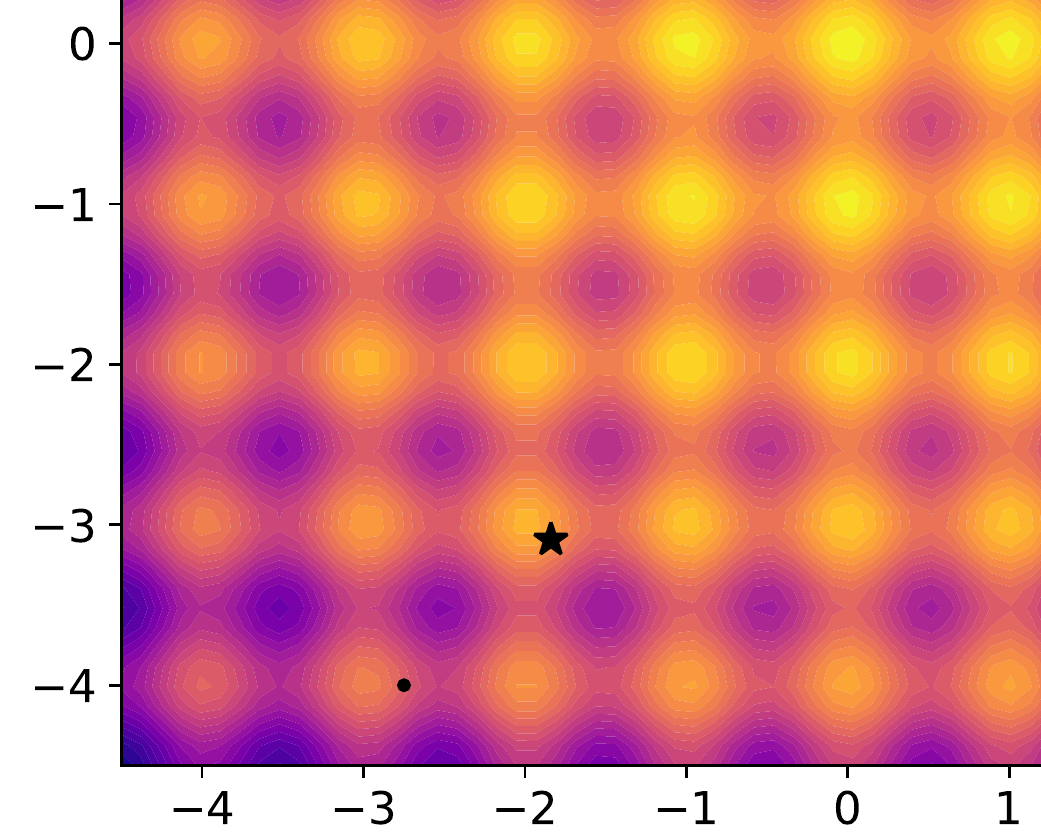}
         \caption{Initial ensemble, $J=2$}
     \end{subfigure}\\[1em]
     
          \begin{subfigure}[t]{0.45\textwidth}
         \centering
        \includegraphics[trim=10 20 50 100, clip,width=\textwidth]{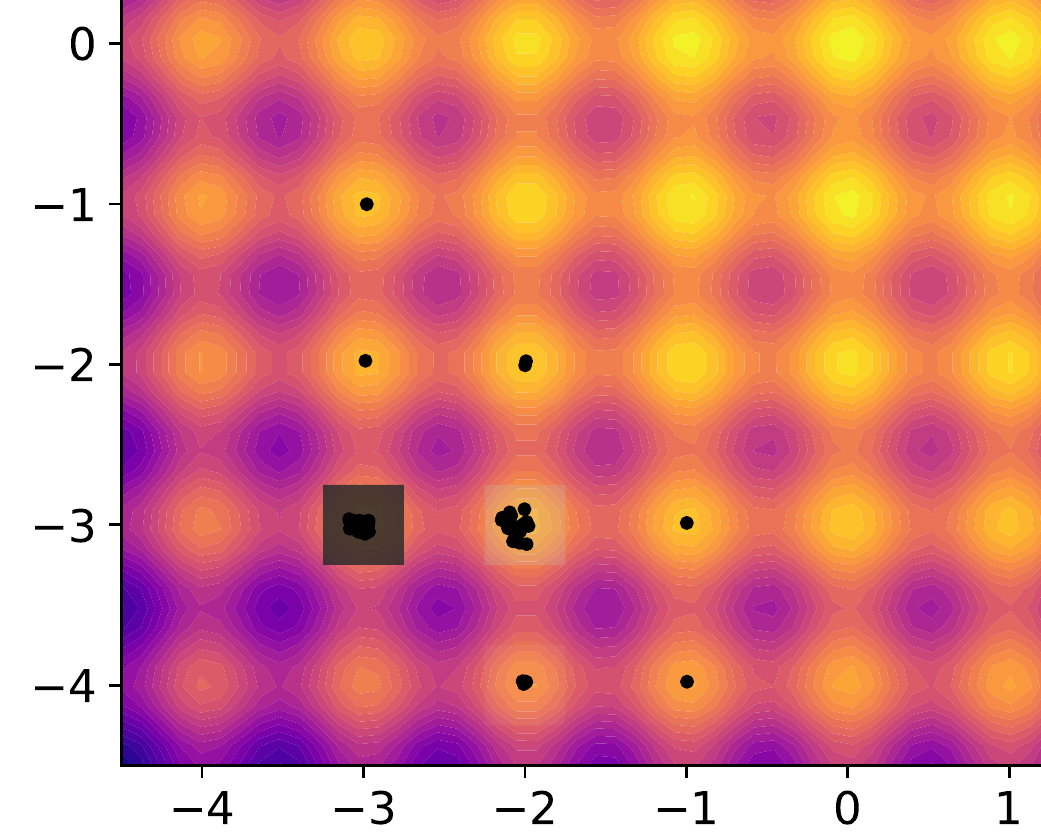}
         \caption{100 runs of CBO ($J=4$).}
     \end{subfigure}
     \hfill
      \begin{subfigure}[t]{0.45\textwidth}
         \centering
        \includegraphics[trim=10 20 50 100, clip,width=\textwidth]{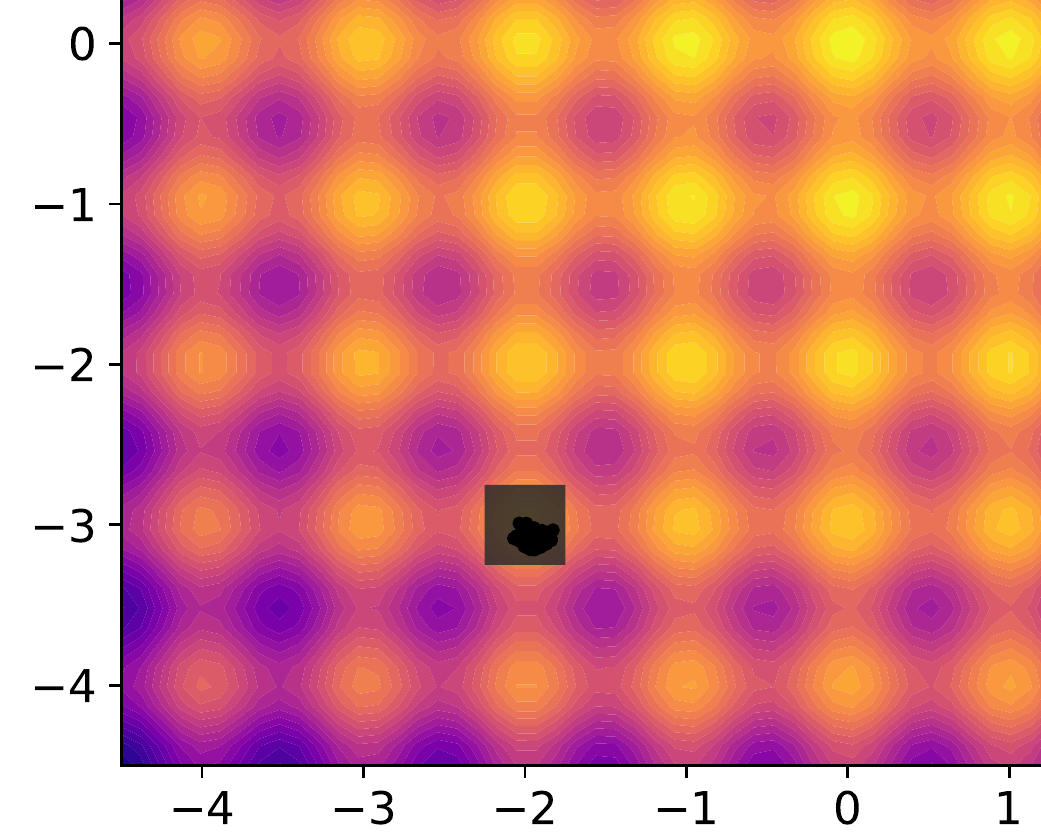}
         \caption{100 runs of CBO ($J=2$)}
     \end{subfigure}\\[1em]
     
     \begin{subfigure}[t]{0.45\textwidth}
         \centering
        \includegraphics[trim=10 20 50 100, clip,width=\textwidth]{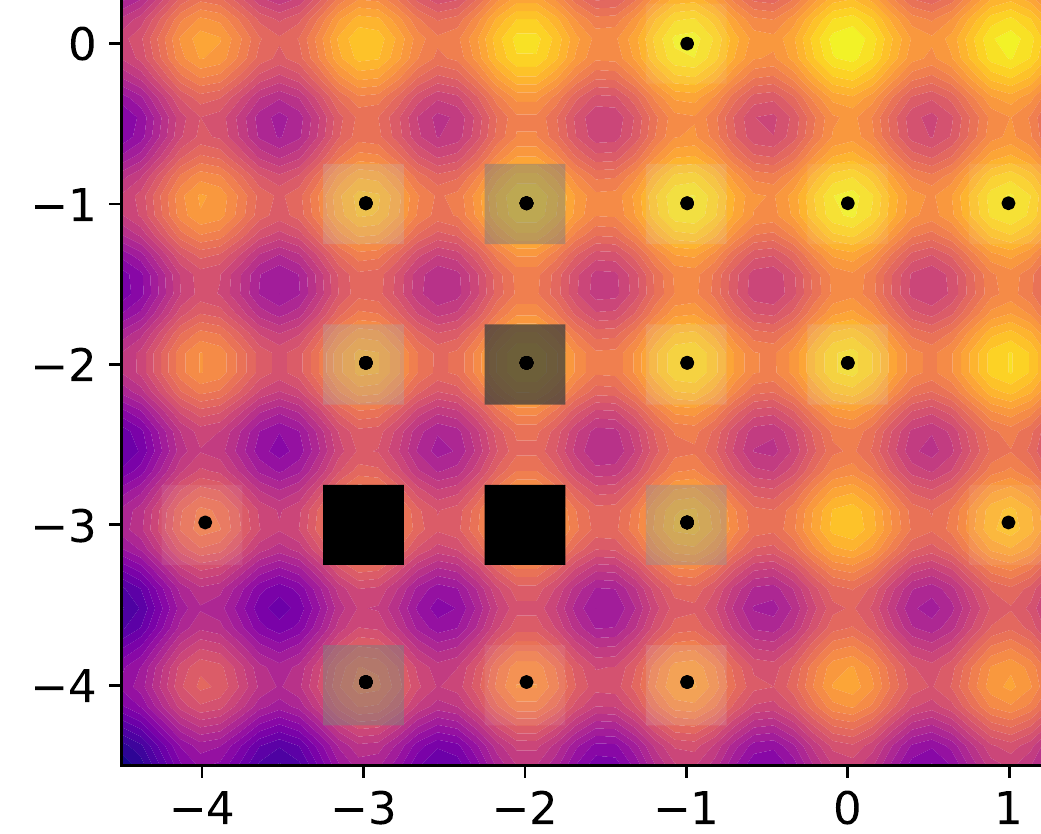}
         \caption{100 runs of EGI-CBO ($J=4$).}
     \end{subfigure}
     \hfill
     \begin{subfigure}[t]{0.45\textwidth}
         \centering
        \includegraphics[trim=10 20 50 100, clip,width=\textwidth]{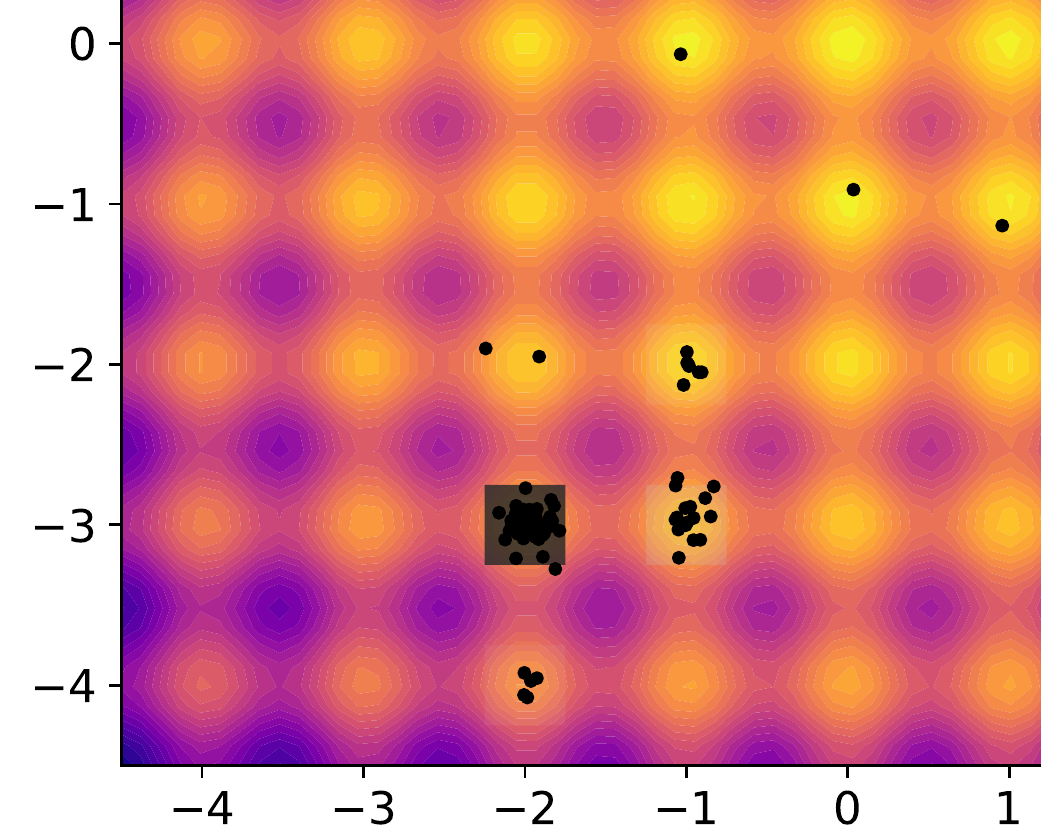}
         \caption{100 runs of EGI-CBO ($J=2$).}
     \end{subfigure}
     \caption{ Code: \texttt{testMC\_cbo\_2d\_rastrigin.py}, \texttt{testMC\_cbo\_2d\_rastrigin\_J2.py}, \texttt{testMC\_aug\_cbo\_2d\_rastrigin.py},\texttt{testMC\_aug\_cbo\_2d\_rastrigin\_J2.py}}
    \label{fig:sim_2d_weighted}
\end{figure}

The results can be observed in \Cref{fig:sim_2d_weighted}: Here we compare CBO and EGI-CBO with ensemble sizes $J=4$ and $J=2$. We run 100 Monte Carlo simulations (with the same initial ensemble but independently sampled noise in the iterations) for each of these four settings and plot the position of the final iteration's weighted mean as a black dot in the function domain. We also superimpose a 2d histogram to give a sense of the distribution of final weighted means of the MC simulations: \rv{We decompose the region into square bins and shade each bin according to the number of final iteration's weighted means ending up in that square. The darker a square, the more individual runs have the final weighted mean in there. For example, \Cref{fig:sim_2d_weighted}(e) shows that there are a lot of runs ending in the square centered around $(-3,3)$, and a lot less ending up in the square centered around $(-2,-2)$.} \rvv{We} can make out the following features:

\begin{itemize}
    \item Convergence towards local minima: It can be observed that the CBO dynamics tends to terminate in a local minimum (and never in the global minimum), but on a position with nonvanishing slope of \rv{$V$}. This is due to the fact that the weighted mean acts as a barrier to the left of the ensemble: The particle closest to the weighted mean experience little to none drift or diffusion itself, with approaching particles being attracted to it. This is resolved by the approximate gradient term in EGI-CBO.
    \item Exploration of ``better minima''. Due to the gradient term in EGI-CBO, the ensemble is able to experience a substantial shift along the essential slope of \rv{$V$}, even surpassing some intermittent local minima. CBO on the other hand can only exhibit deterministic contraction towards $\wmean$ and stochastic exploration (which ignores the shape of \rvv{$V$}). Note that this effect is stronger for ``global'' EGI-CBO with $\xi>0$, see below.
    \item This holds to some extent also in the underdetermined case $J=2$, although convergence towards local minima is only slightly accelerated: With only two live particles, gradient information will never point directly towards local minimum.
\end{itemize}

\begin{remark}
In addition to the observations discussed above, the EGI-CBO turns out to reliably find various local minima of the highly multi-modal function within $100$ Monte Carlo runs. This opens a new field of applications for CBO and might be worth to be investigated in future work.
\end{remark}

\begin{figure}
     \centering
     \begin{subfigure}[t]{0.4\textwidth}
         \centering
\includegraphics[width=\textwidth]{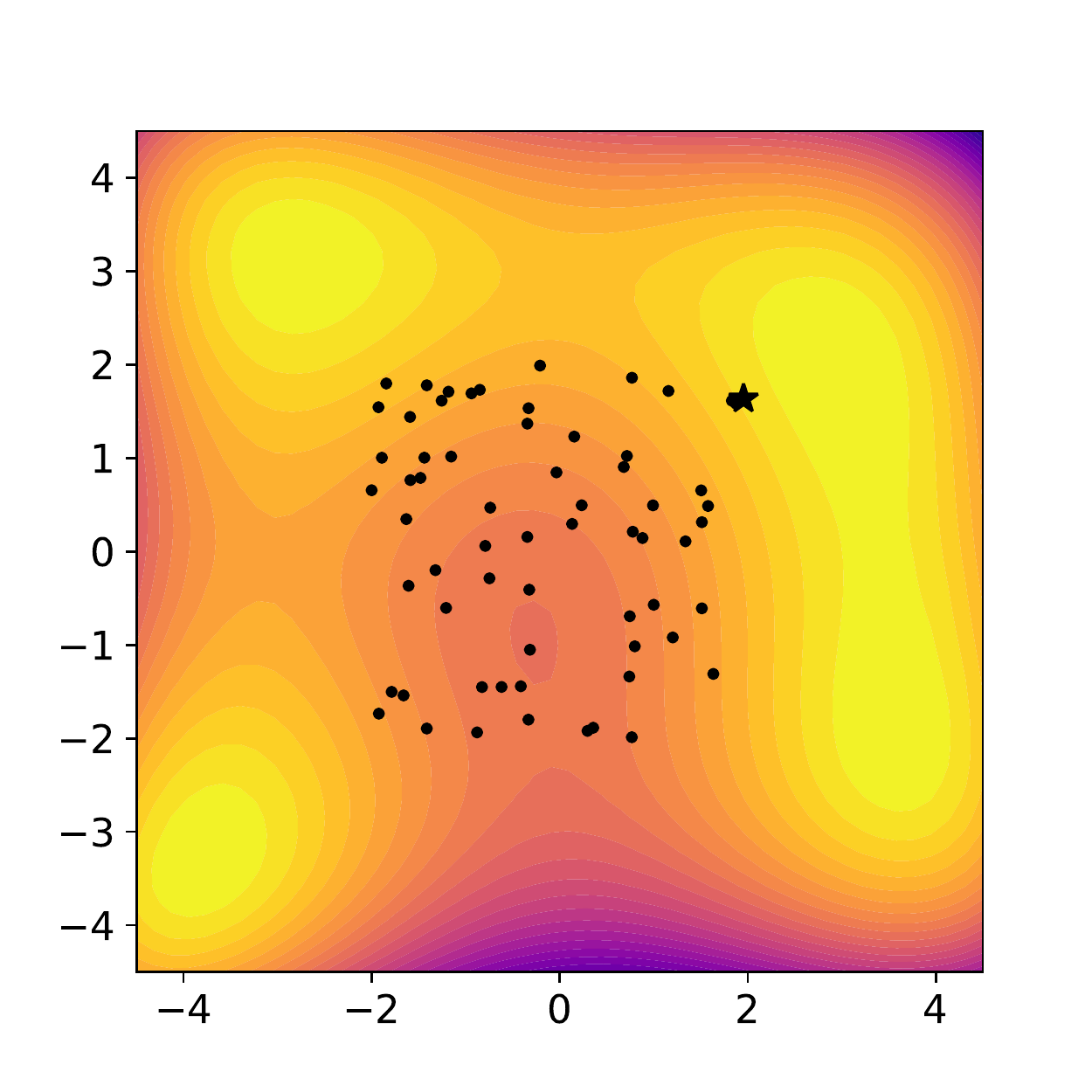}
         \caption{Initial ensemble $J=64$ (black dots) with initial weighted mean (star)}
     \end{subfigure}\hfill
          \begin{subfigure}[t]{0.4\textwidth}
         \centering
        \includegraphics[width=\textwidth]{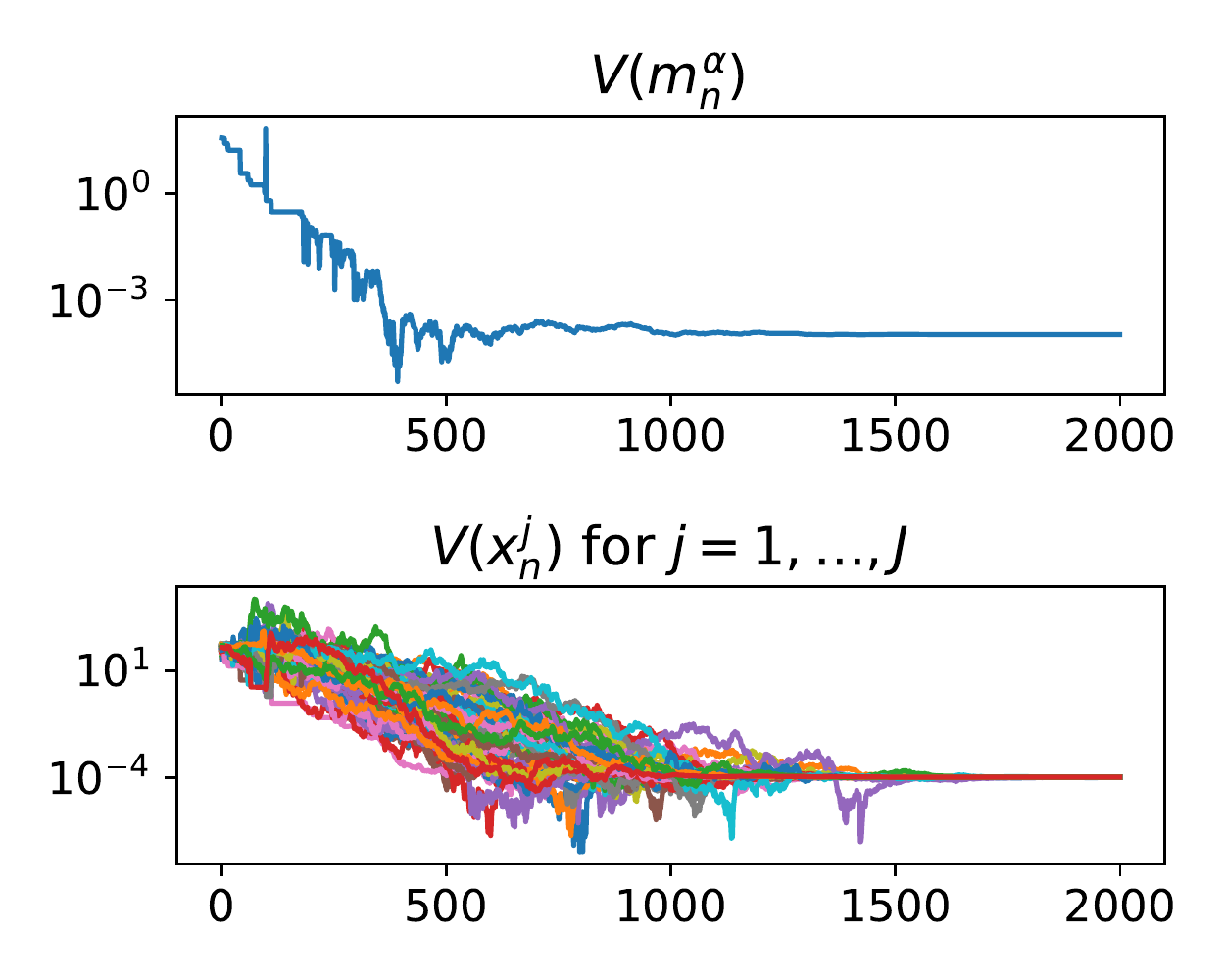}
         \caption{Representative run of CBO ($J=64$). }
     \end{subfigure}\\
     \begin{subfigure}[t]{0.4\textwidth}
         \centering
\includegraphics[width=\textwidth]{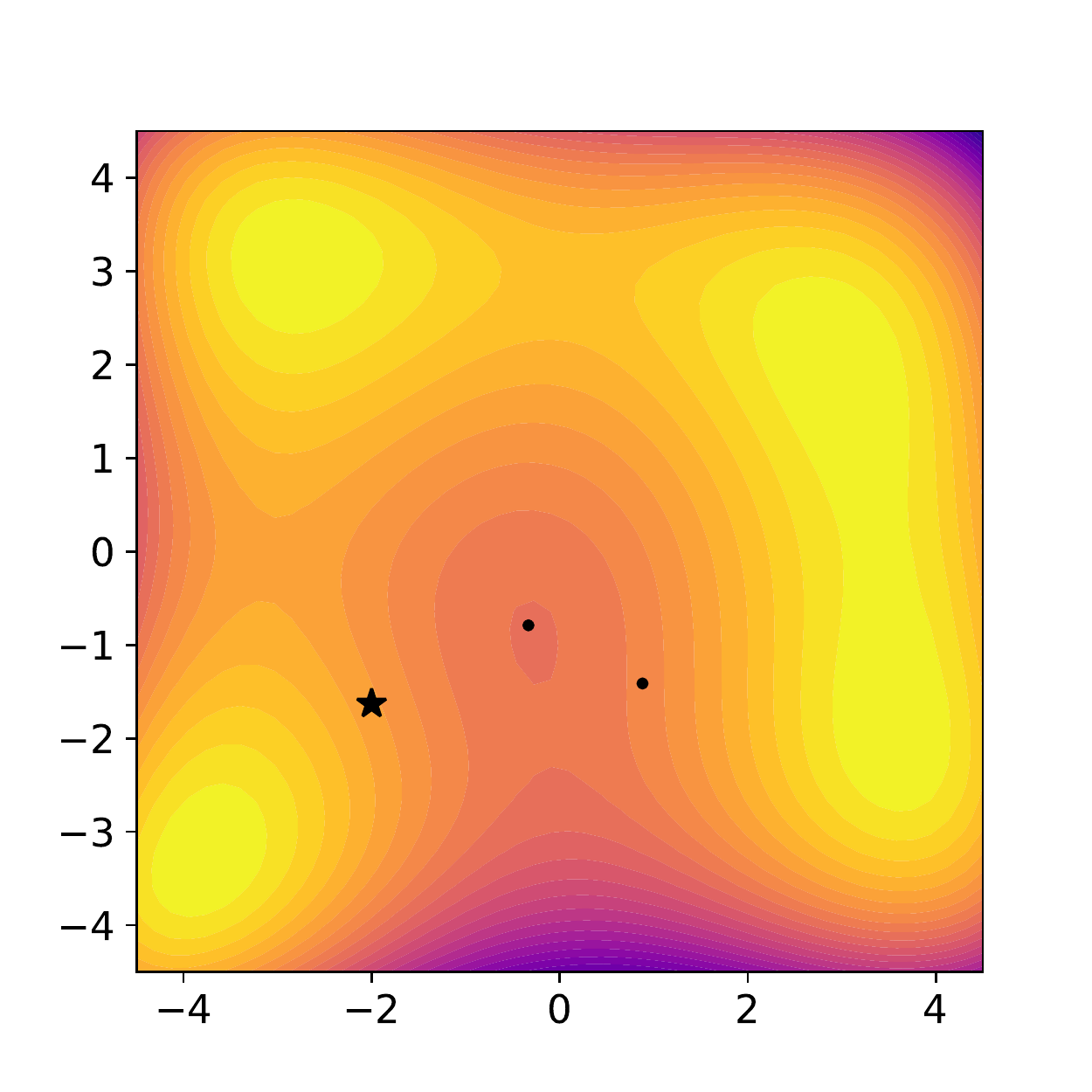}
         \caption{Initial ensemble $J=3$ (black dots) with initial weighted mean (star)}
     \end{subfigure}\hfill
          \begin{subfigure}[t]{0.4\textwidth}
         \centering
        \includegraphics[width=\textwidth]{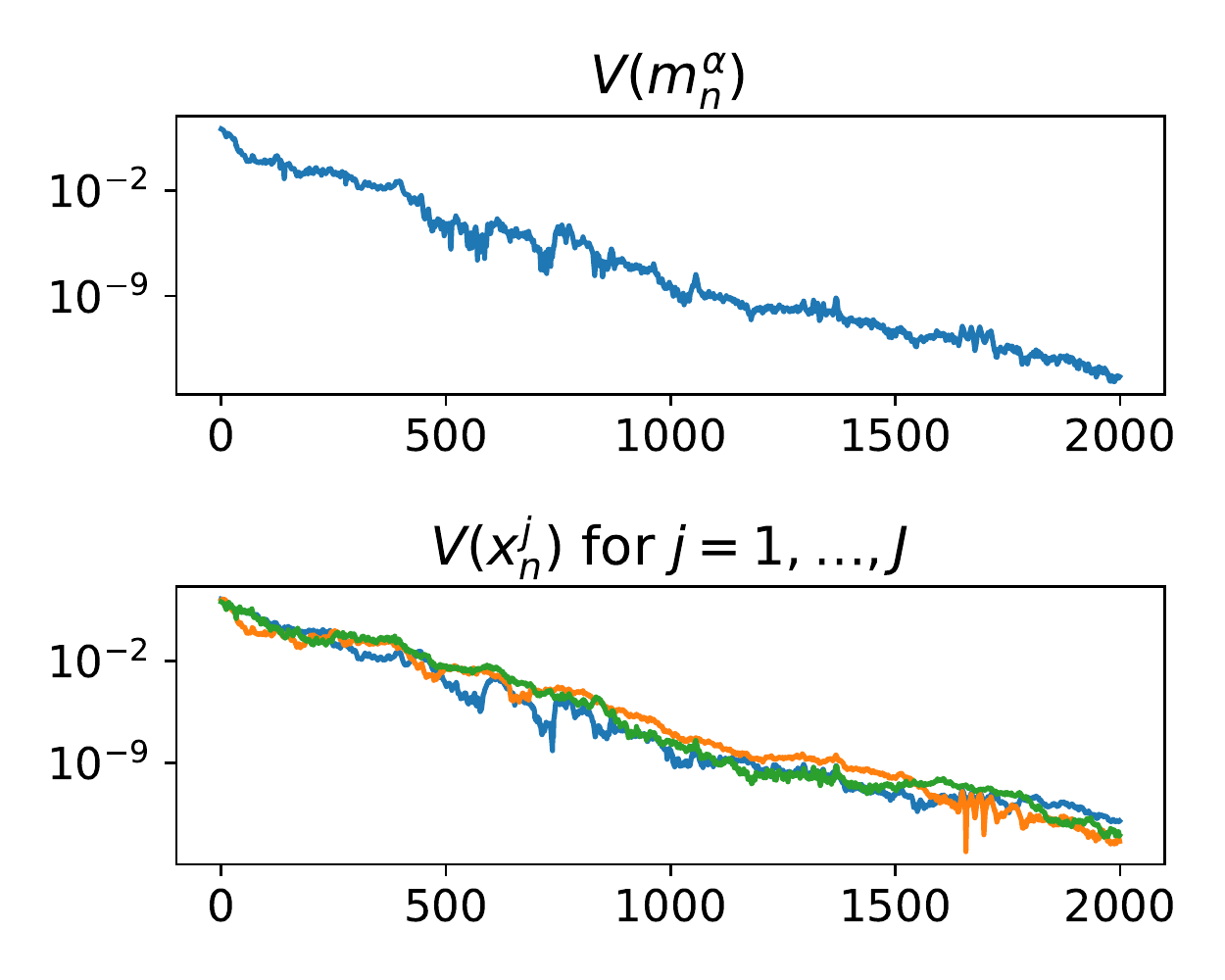}
         \caption{Representative run of EGI-CBO ($J=3$)\rv{, $\gamma=0$.}}
         \end{subfigure}
          \hfill
          \begin{subfigure}[t]{0.4\textwidth}
         \centering
        \includegraphics[width=\textwidth]{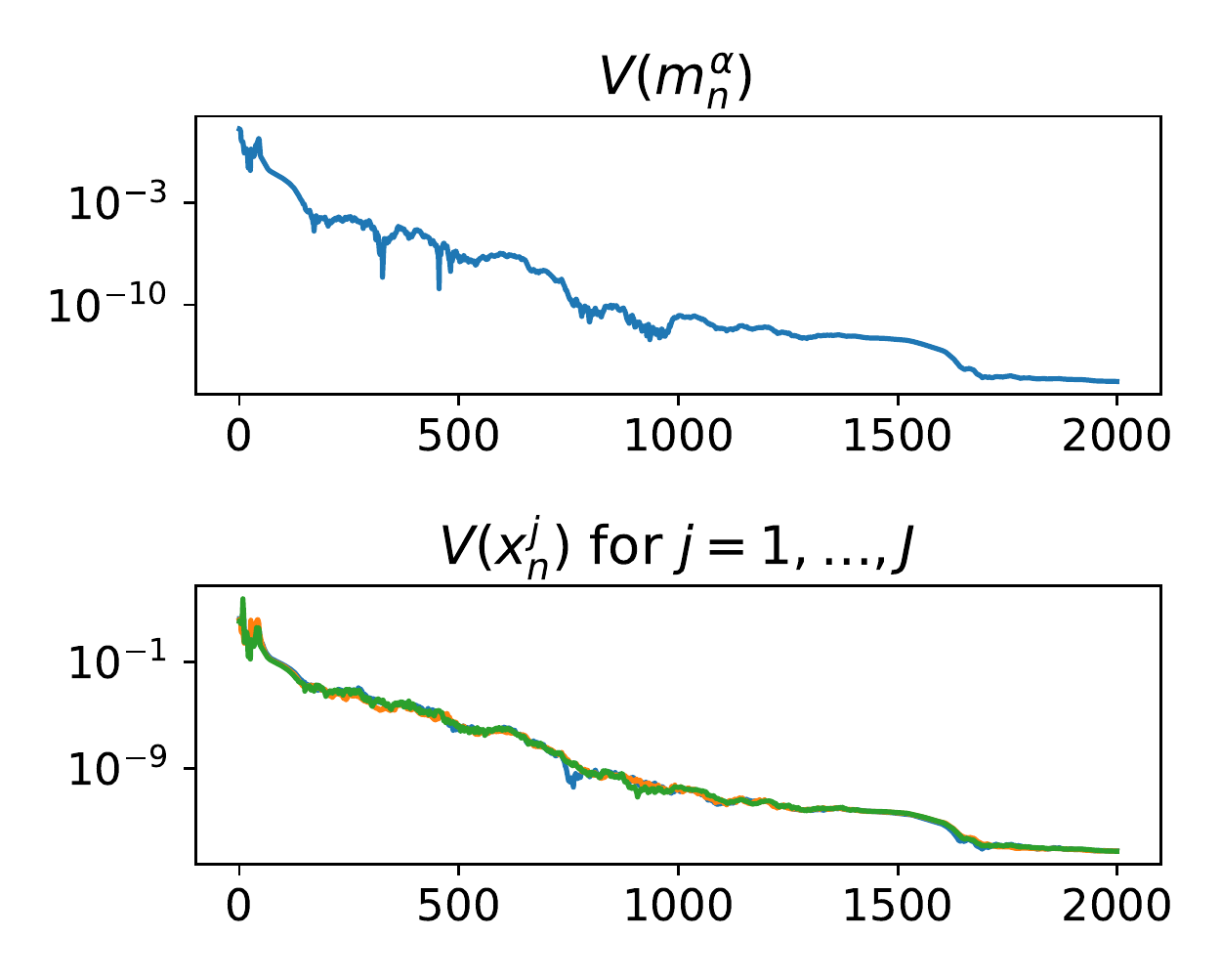}
         \caption{Representative run of EGI-CBO ($J=3$)\rv{, $\gamma=1e4$, with extrapolated gradients.}}
         \end{subfigure}
         \caption{Application of CBO and EGI-CBO to the Himmelblau test function. Code: \texttt{test\_cbo\_2d\_himmelblau.py}, \texttt{test\_aug\_cbo\_2d\_himmelblau.py}, \texttt{test\_aug\_linreg\_cbo\_2d\_himmelblau.py}.}
    \label{fig:sim_2d_himmelblau}
\end{figure}

\paragraph{Himmelblau function in 2d}
In contrast to the Rastrigin function with steep valleys the Himmelblau function is rather flat in the neighborhood of its minima. We also compare the behaviour of CBO and EGI-CBO in this setting. The Himmelblau function exhibits four global minima with value $0$. Figure \ref{fig:sim_2d_himmelblau} shows a comparison of a single run of CBO with $64$ particles and a single run of EGI-CBO with only $3$ particles. We remark that all our experiments show similar results, so we show only one specific instance. This allows us to better represent the dynamics. It can be observed that EGI-CBO with $\gamma = 0$ (i.e. employing ``local EGI'' within CBO) actually converges to one of the global minima, with an exponential rate, while the ensemble of CBO collapses to a point close to a minimum, but not towards the minimum itself. \rv{For comparison, we also show the performance of EGI-CBO with $\gamma=1e4$ (i.e. employing ``global EGI'' within CBO), with extrapolated gradients. This corresponds to augmenting CBO with gradients obtained from ordinary least squares regression. It can be observed that this is an improvement over vanilla CBO, but the exponential decay of the objective function tapers off after roughly 1000 iterations. We commit a more in-depth analysis of this effect to future research.}

\paragraph{High-dimensional smooth example}
To further highlight the advantages of the gradient augmentation, we employ a high-dimensional smooth convex example, $V:x\mapsto \frac{1}{2}\|x-(1,\ldots,1)\|^2$ on $\R^d$, where $d=10$. Clearly, $x = (1,\ldots,1)$ is the global minimum of $V$. We sample the initial ensemble uniformly on $[-4,-1]^d$ and emphasize that this domain excludes the global minimum. For the parameters, we set $\alpha=100$, $\lambda=1$, $\sigma=0.2$ and we use component-wise noise for dimension-robustness. We start with the over-determined case $J=20$. EGI-CBO ($\kappa = 4.0$) exhibits exponential convergence, which is not that surprising given that it essentially performs gradient descent, with its ensemble size having enough descriptive power to span the full space. Vanilla CBO does not converge to the minimum due to contraction of the ensemble on a point different from the minimum. The fact that the initial ensemble is far away from the minimum is an additional adversarial factor. Further experiments show that this is not resolved by increasing the noise term as the stochastic dynamics becomes unstable for a noise level larger than a certain threshold, even for component-wise noise.

\begin{remark}\label{rem:cbotheory}
We want to emphasize that this is not in contradiction with the theory for CBO \rvv{\citep{carrillo2018analytical,fornaiser2021globally}} as the proofs consider the mean-field setting where the diffusion instantaneously extends the support of the distribution to the whole domain, and many proofs assume additionally that the unique global minimizer is contained in the support of the initial distribution.  
\end{remark}

The case $J=5$ is more interesting: While CBO performs badly (as to be expected, see Remark~\ref{rem:cbotheory}), EGI-CBO is subjected to enough noise that \rv{the subspace spanned by the ensemble -- by virtue of additive noise not in alignment with this affine subspace -- moves enough such that} accumulated  gradient information \rv{(in the sense that the subspace has varied sufficiently throughout the iterations)} brings this underdetermined ensemble quite close to the global minimum, although the ensemble has size lower than the spatial dimension.

\begin{figure}

     \begin{subfigure}[t]{0.45\textwidth}
\includegraphics[width=\textwidth]{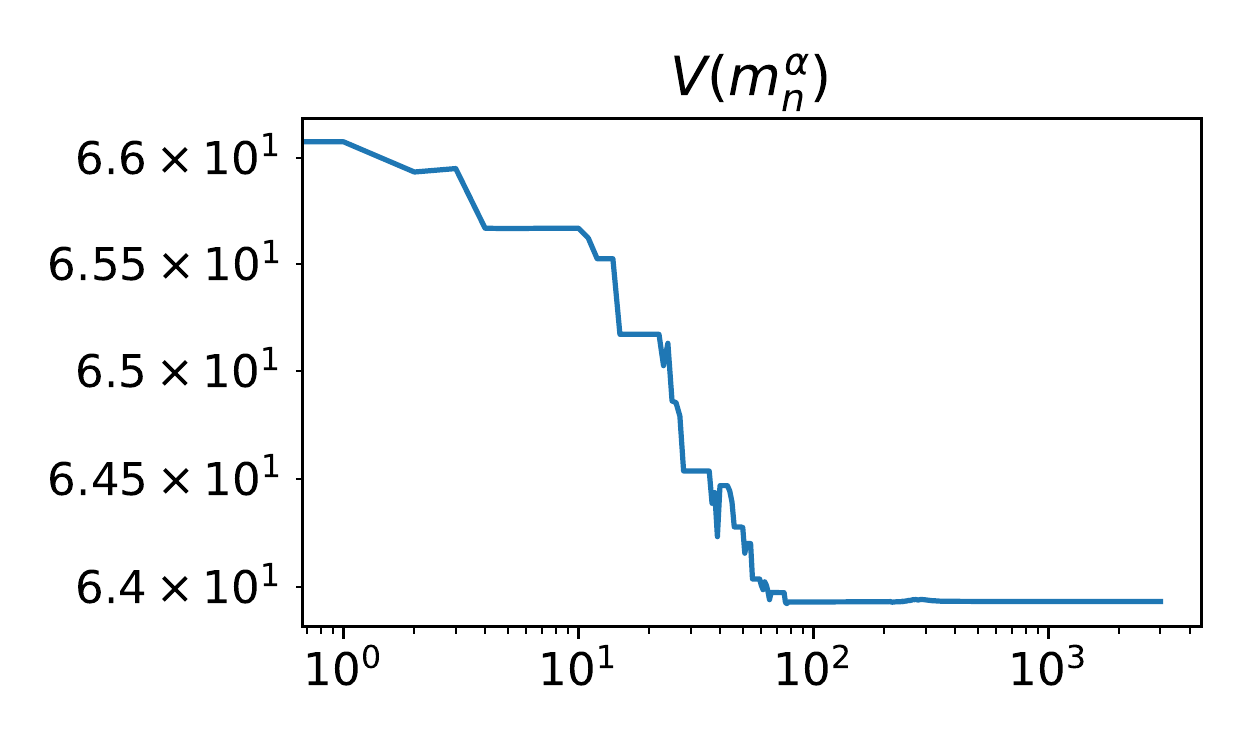}
\caption{CBO, $J=5$}
\end{subfigure}
     \begin{subfigure}[t]{0.45\textwidth}
\includegraphics[width=\textwidth]{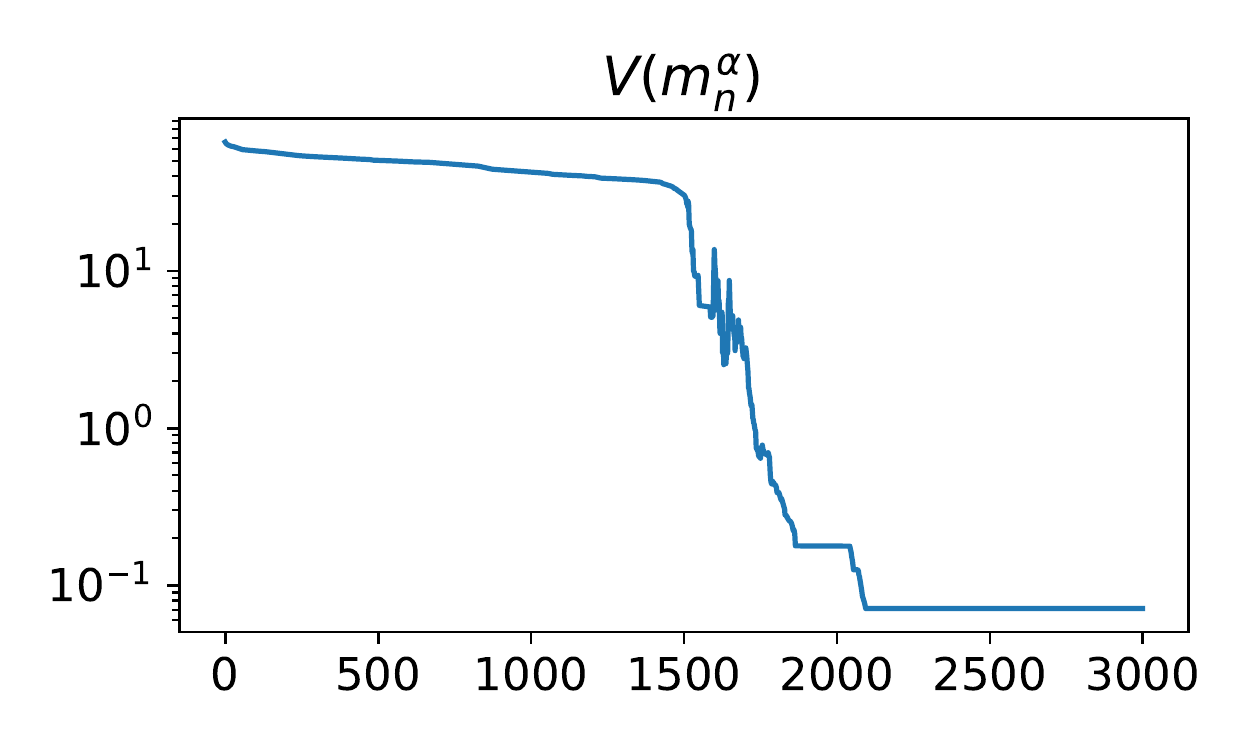}
\caption{EGI-CBO, $J=5$}     
\end{subfigure}

\begin{subfigure}[t]{0.45\textwidth}
\includegraphics[width=\textwidth]{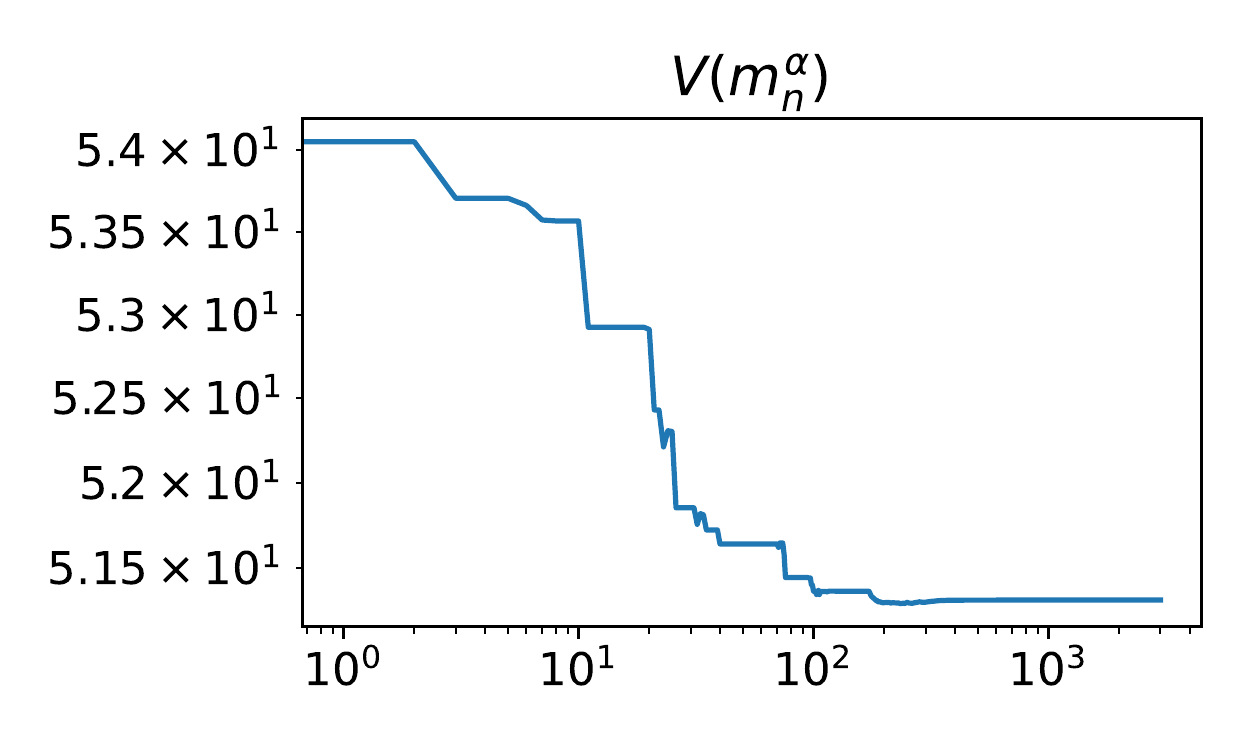}
\caption{CBO, $J=20$}
\end{subfigure}
     \begin{subfigure}[t]{0.45\textwidth}
\includegraphics[width=\textwidth]{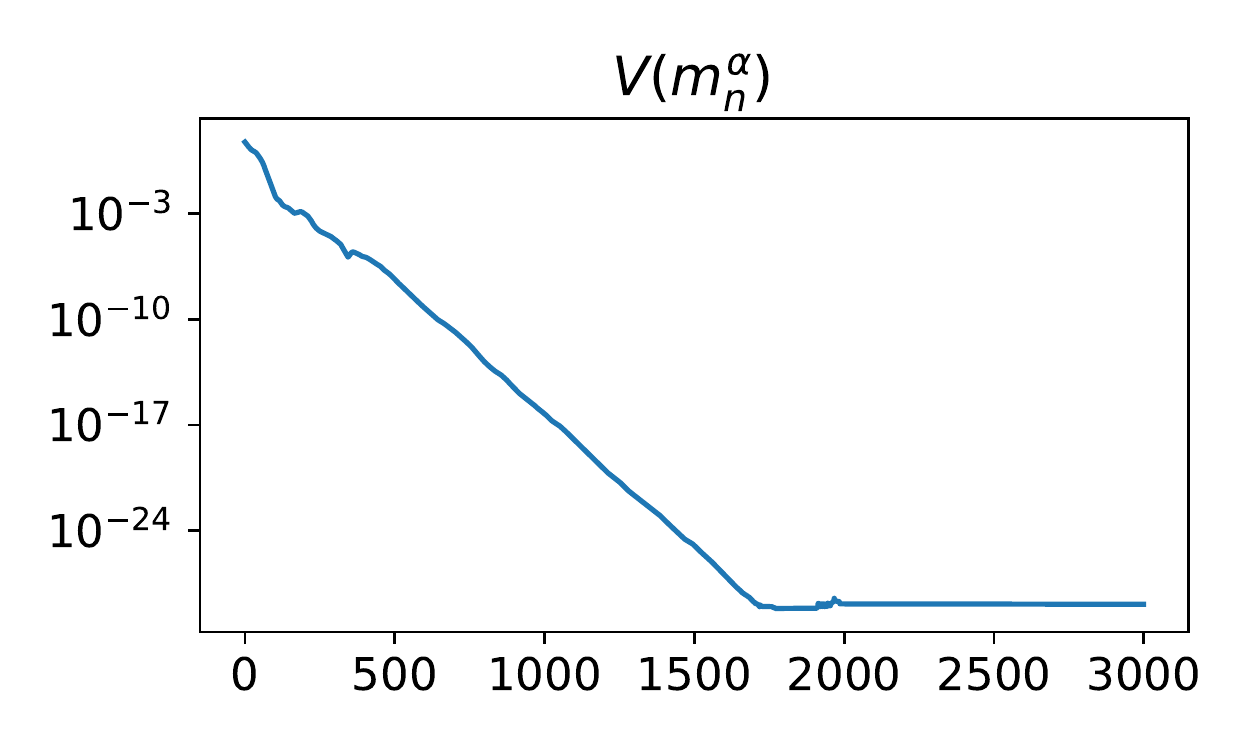}
\caption{EGI-CBO, $J=20$}
\end{subfigure}
         \caption{Performance of CBO and EGI-CBO on the function $V: x\mapsto \|x-(1,\ldots,1)\|^2$ in $\R^{10}$. Diagrams show evaluation of $V$ on the weighted mean over the course of the whole iteration. Note that CBO is plotted over log-iterations in contrast to EGI-CBO, which is plotted over linear iterations. CBO levels off close to the initialization, with exponentially growing plateau lengths, while EGI-CBO converges exponentially fast to the global minimum (more quickly for larger ensembles, with jumps for smaller ensembles). Code: \texttt{test\_cbo\_ndnorm.py}}
\end{figure}

In the previous numerical examples the additional smoothing error term in \eqref{eq:invprlong} is switched off ($\xi=0$). As discussed in Figure~\ref{fig:localapprox} this corresponds to local gradient approximations. In next section we study the influence of nonlocal gradient approximations.

\subsection{Global versus Local EGI-CBO}
\begin{figure}
  \begin{subfigure}[t]{0.45\textwidth}
\includegraphics[width=\textwidth]{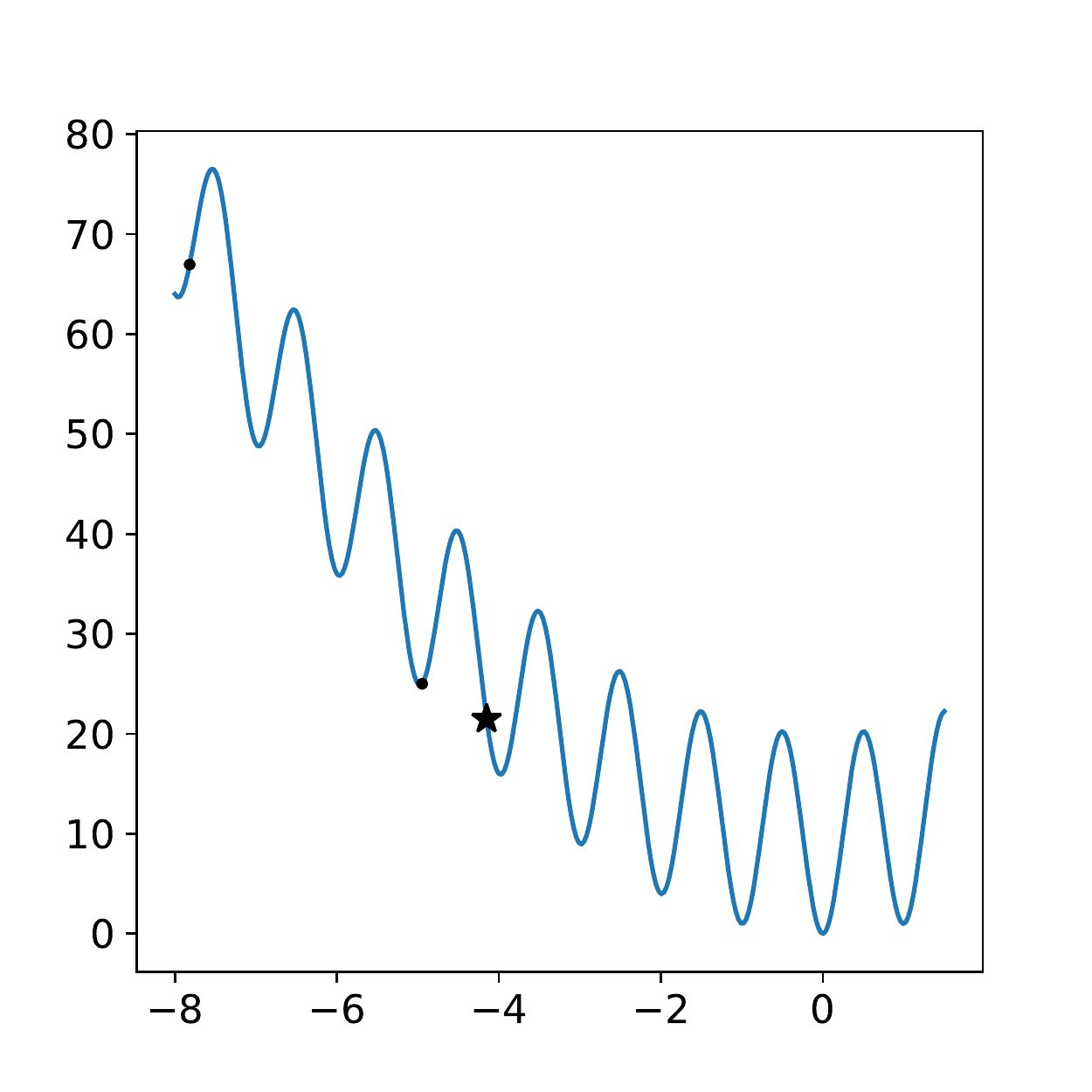}
\caption{Initial ensemble ($J=3$) chosen for all simulations. Star marks weighted mean.}
\end{subfigure} \hfill \begin{subfigure}[t]{0.45\textwidth}
\includegraphics[width=\textwidth]{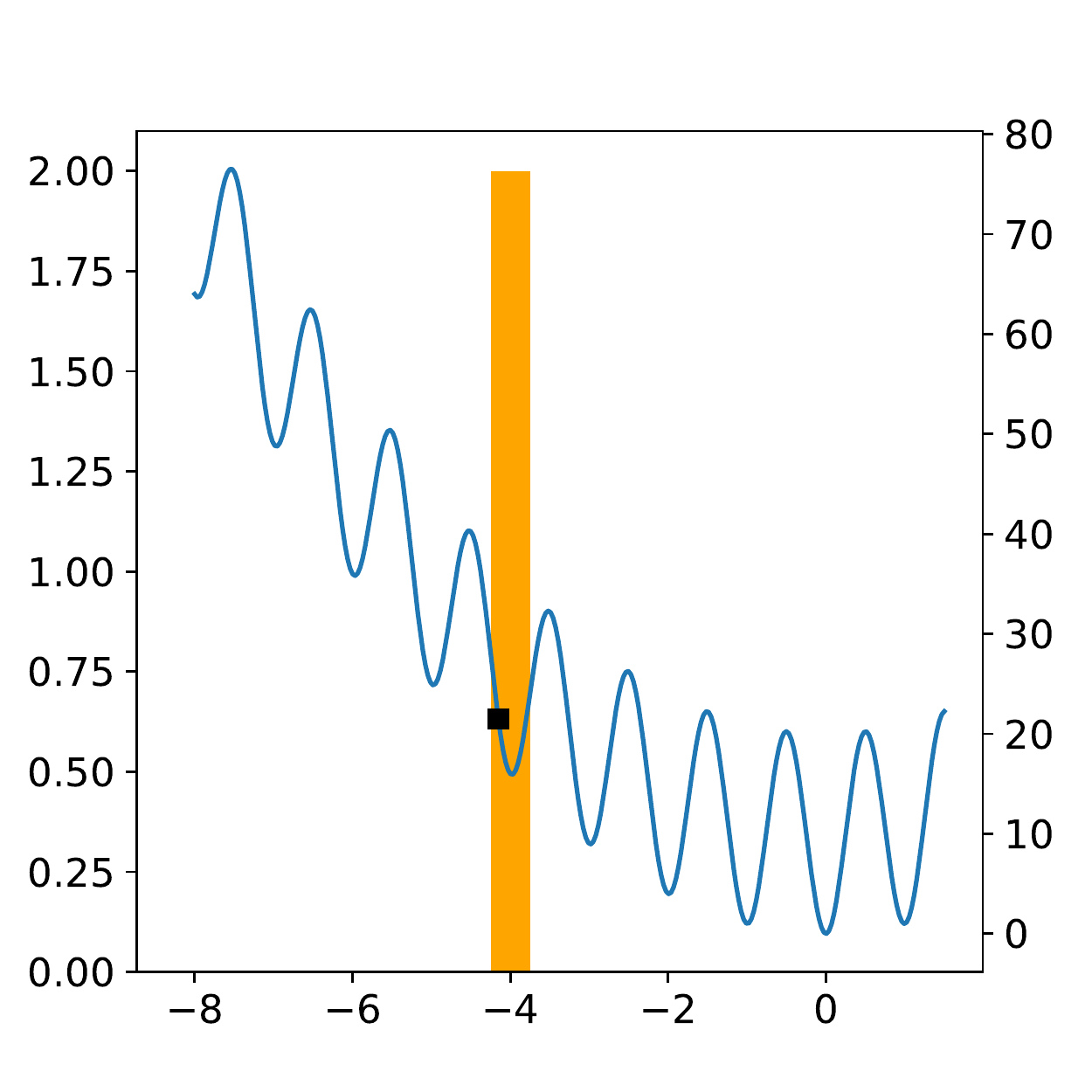}
\caption{Histogram of final weighted mean of 100 runs of CBO. All results are equal to the initial weighted mean of the ensemble.}
\end{subfigure}

  \begin{subfigure}[t]{0.45\textwidth}
\includegraphics[width=\textwidth]{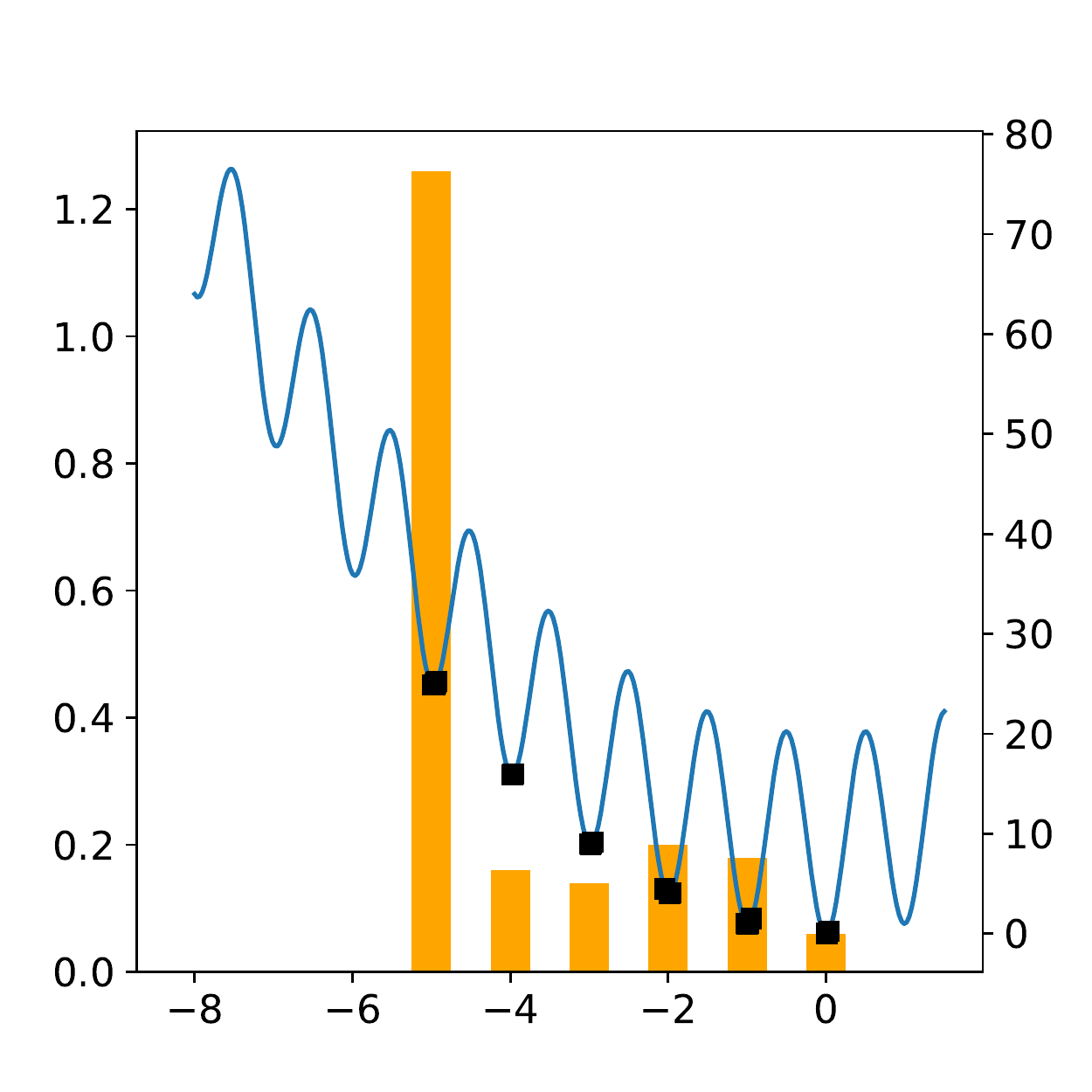}
\caption{Histogram of final weighted mean of 100 runs of EGI-CBO ($\xi=0$).}
\end{subfigure}\hfill
  \begin{subfigure}[t]{0.45\textwidth}
\includegraphics[width=\textwidth]{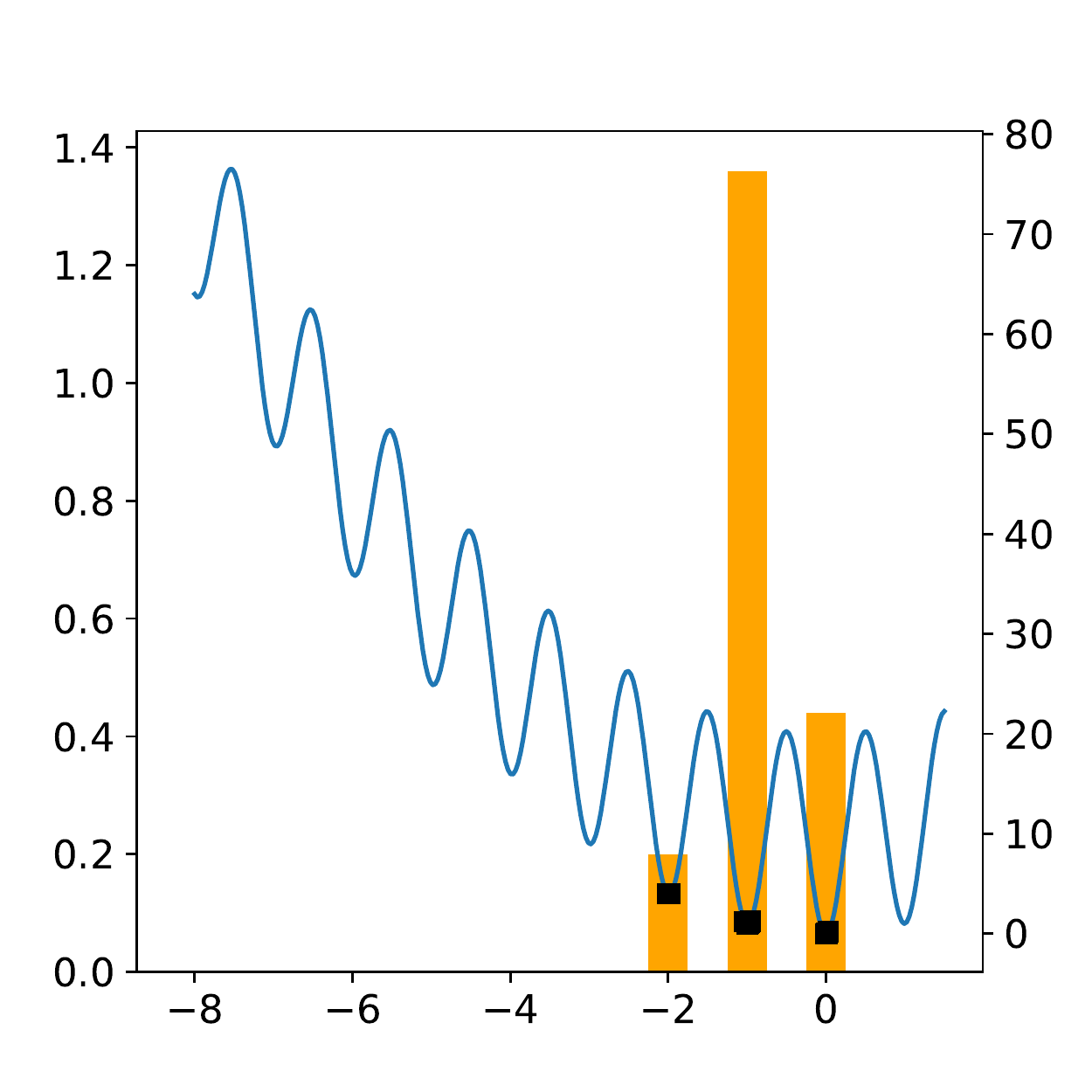}\caption{Histogram of final weighted mean of 100 runs of global EGI-CBO ($\xi > 0$).}
\end{subfigure}
\caption{Demonstration of benefits of global EGI-CBO for multimodal optimization. Code: \texttt{testMC\_cbo\_globalapprox\_1d\_rastrigin.py}}
\label{fig:globally_augmented}
\end{figure}

Figure~\ref{fig:localapprox} showing global EGI for the one-dimensional Rastrigin function suggests that ``global'' EGI-CBO (i.e. $\xi > 0$) can lead to improved exploration behaviour. In the following we investigate the influence of $\xi$ in more detail. We therefore employ EGI-CBO as described in algorithm \ref{alg:augCBO} with $\xi>0$. This means that line \ref{line:gradientCBO} is modified to $g_n \gets G_\xi^0(\{m_n\}\cup\{x_n^i\}_{i=1}^J)$ (i.e. with explicit dependence on $\xi$), where we now compare the settings $\xi  = 0$ and $\xi >0$ for varying values. We set $\lambda = 1.0$, $\sigma = 0.5$ and in the augmented case we have additionally $\kappa = 2.0$ as well as $\xi = 100$. Parameters of the time discretization are $T=10$, $\tau = 0.01$. In Figure~\ref{fig:globally_augmented} we note that augmentation drives the ensemble towards the true minimizer. In contrast, the reference solution of CBO gets stuck in higher level set regions.
 
\subsection{Discussion of EGI-CBO}
To conclude the section on EGI-CBO we discuss some limitations of the method.

\begin{figure}
    \centering
    \includegraphics[width=0.75\textwidth]{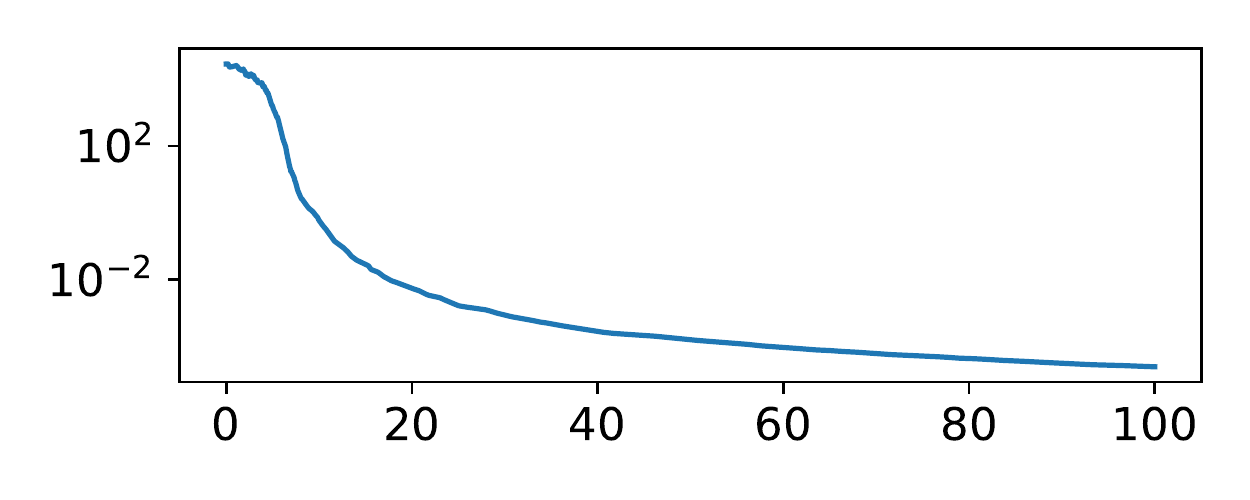}
    \caption{\rv{Subexponential } Convergence toward a higher-order minimum.}
    \label{fig:x4}
\end{figure}

    \paragraph{Subexponential convergence to higher-order minima.} For further investigation of the speed of convergence, we choose $V(x) = \|x\|^4$ in $d=50$ and set $J=10$, $\kappa=2.5$, $\sigma=0.2$, $\lambda=2.5$. In contrast to $\|x-(1,\ldots,1)\|^2$, where EGI-CBO performed very nicely, we note that the gradient of $V$ vanishes rapidly in the neighborhood of the global minimum vanishes. Figure~\ref{fig:x4} indicates that flat regions around the minimizer slightly diminish the advantage of EGI-CBO over vanilla CBO. Indeed, EGI-CBO exhibits only subexponential convergence. Nevertheless, it still outperforms CBO which does not converge to the minimum at all. (The latter fact is not demonstrated here, but experiments show the same behaviour as for the test function $\|x-(1,\ldots,1)\|^2$ above).
    
    \paragraph{Very high-dimensional multimodal functions.} In the literature \rvv{\citep{pinnau2017consensus,carrillo2021consensus}} very good performance of CBO for the high-dimensional ($d=20$) Rastrigin function is reported. In order to achieve this performance, it is necessary to either set $J\gg d$ \rvv{\citep{pinnau2017consensus} or $\sigma \gg \lambda$ \citep{carrillo2021consensus}}. In the latter case the diffusion dominates the drift towards the weighted mean, thus leaving the contractive domain of CBO which was established \rvv{theoretically \citep[Theorem 3.2]{carrillo2021consensus}}. In particular, CBO exhibits a random exploration with a slight bias towards the weighted mean, and contraction of the ensemble does not happen. 
    Employing additional gradient information in this non-contractive domain of CBO, leads to numerical instabilities which do not improve the performance of vanilla CBO.
    
    \paragraph{Smooth unimodal objective functions. } If $V$ is a smooth unimodal function with convex (or ``nearly convex'') structure -- for example the typical inverse elliptical problem for strongly informative data -- CBO should not be used, as other methods typically have much better performance. Similarly, if true gradient is cheap to obtain, CBO is probably not the method of choice as it requires many function evaluations. Nethertheless, for real-world applications often no structure is known a-priori and only functions evaluations are available. Then (augmented) CBO is a competitive alternative to other population based methods, which is furthermore backed-up by theoretical convergence proofs.

\section{Gradient-based sampling using EGI}\label{sec:sampling}
\rv{In this section we consider the problem of generating samples from a measure $\d\mu/\d x \propto \exp(-V(x))$.
Any mention of $G_\xi^j$ will be the result of \Cref{alg:grad} as applied to the function $V$. }

In contrast to our approach with CBO (a completely gradient-agnostic method which we augmented via inexact gradients), we now describe how some gradient-based sampling methods can be adapted to use inexact ensemble-based gradients instead.
\subsection{EGI-LS and EGI-MALA}
We begin with the unconditioned case, i.e. $M=\mathrm{Id}$ and propose the following gradient-free sampling method as substitute dynamics: 
\begin{align}
    \d x^j_t = -G^j(\{x^i\}_{i=1}^J) \d t + \sqrt{2}\d W^j_t
\end{align}
with initial condition $x_0^i \sim \mathcal P_2(\R^d)$ independently drawn for $i=1,\dots,J.$ A pseudocode for the implementation of the dynamics is given in Algorithm~\ref{alg:EL}.

In the same spirit, we provide results for the other ensemble-based sampling algorithms presented in the introduction (see Section~\ref{sub:existingmethods}). Note that we set $\xi=0$ by default. We expect that global gradient approximations lead to similar behaviour as in the optimization case. For some sampling applications it might make sense to allow for ``global gradient approximations'' (i.e. $\xi > 0$). For the sake of legibility we leave a detailed study to future investigations.

An Metropolis adjustment, the so-called MALA sampler, was proposed to improve the sampling accuracy of discretized Langevin dynamics in \cite{roberts1996exponential}. In our setting this yields a straight-foward generalization of MALA, the EGI-augmented Metropolis-adjusted Langevin algorithm (EGI-MALA), see Algorithm \ref{alg:EGI-MALA}.

\begin{algorithm}[H]
\caption{EGI-augmented Langevin Sampler \textbf{(EGI-LS)}}\label{alg:EL}
\Comment{additional linear equations: $J$} 
\KwData{$N \in \N, \{x_0^j\}_{j=1}^J$, step size $h$}
\KwResult{samples $\{x_n^j\}_{j=1, n=1}^{J,N}$ from $\d\mu/\d x \propto \exp(-V(x))$}
\For{$n\gets0$ \KwTo $N-1$ }{
    \For{$j\gets1$ \KwTo $J$\label{algline:for1} }{
    $g_n^j = G^j(\{x_n^i\}_{i=1}^J)$\Comment{approximate gradient in proposals, via \Cref{alg:grad}}
    $W_n^j\sim \mathcal N(0,1)$\;
    $x_{n+1}^j \gets x_n^j - h g_n^j + \sqrt{2h} W^j_n$\;
    }\label{algline:for2}
    }
\end{algorithm}
\begin{algorithm}[H]
\DontPrintSemicolon
\caption{\textbf{(EGI-MALA)}}\label{alg:EGI-MALA}
\Comment{additional linear equations: $2J$} 
\KwData{$N \in \N, \{x_0^j\}_{j=1}^J$, $\tau>0$}
\KwResult{samples $\{x_n^j\}_{j=1, n=1}^{J,N}$ from $\d\mu/\d x \propto \exp(-V(x))$}                                                          
\For{$n\gets0$ \KwTo $N-1$ }{
    \For{$j\gets1$ \KwTo $J$ }{
    \eIf{$n\geq 1$} 
    {
        $g_n^j = G^j(\{x_n^i\}_{i=1}^J\cup \{\mu_{n-1}^i\}_{i=1}^J)$ \Comment{approximate gradient, via \Cref{alg:grad}}
    }
    {
        $g_n^j = G^j(\{x_n^i\}_{i=1}^J)$ \Comment{approximate gradient, via \Cref{alg:grad}}    }
        $W_n^j\sim  \mathcal N(0,I^d)$\;
    $\mathrm{prop}_n^j \gets x_n^j - \tau g_n^j + \sqrt{2\tau} W^j_n$\Comment{Langevin-type proposal}
    $\gamma_n^j = G^j(\{\mathrm{prop}_n^i\}_{i=1}^J)$\Comment{approximate gradient in proposals, via \Cref{alg:grad}}
    }
    \For{$j\gets1$ \KwTo $J$ }{
    $q_\mathrm{fwd}^j\gets \exp\left(-\frac{1}{4 \tau}\|\mathrm{prop}_n^j - (x_n^j - \tau g_n^j) \|^2 \right)$\Comment{Metropolis adjustment}
    
    $q_\mathrm{bwd}^j\gets \exp\left(-\frac{1}{4 \tau}\|x_n^j - (\mathrm{prop}_n^j - \tau \gamma_n^j) \|^2 \right)$
    
    $\alpha_j \gets \frac{\exp(-V(\mathrm{prop}_n^j))\cdot  q_\mathrm{bwd}^j}{\exp(-V(x_n^j))\cdot q_\mathrm{fwd}^j} $
    
    $\xi \sim \mathrm{Unif}[0,1]$\;
    \eIf{$\xi \leq \alpha_j$}
    {
        $x_{n+1}^j \gets \mathrm{prop}_n^j$
        
        $\mu_n^j \gets x_n^j$ \label{alg:mem1}\Comment{keep previous iterate as memory}
    }{
        $x_{n+1}^j \gets x_n^j$
        
        $\mu_n^j \gets \mathrm{prop}_n^j$ \Comment{Keep (rejected) proposal as memory} \label{alg:mem2}
    }
    }
        
    }
\end{algorithm}

\subsection{EGI-ALDI and EGI-EKS}
 The extraordinary feature of gradient-free ALDI is that it implicitly constructs and uses gradient information of $V$  without the need of solving a linear system as demonstrated in \ref{sub:existingmethods}. Unfortunately, this approximation fails if $A$ is nonlinear. \rv{In addition, this error does not vanish for increasing ensemble size $J\to \infty$, and it also does not improve in regions of the state space where ensemble members are clustered more densely.}

For this reason, we also consider a new version of this methodology by replacing gradient-free ALDI's implicit gradient (biased for nonlinear forward maps) by our inexact gradient approximation via EGI \rv{(with the property that the exact gradient is recovered for $J\to\infty$, and the gradient approximation is better in regions of densely clustered ensemble members)}, leading to the following sampling algorithm, EGI-ALDI (gradient-free ALDI with estimated gradients):

\begin{algorithm}[H]
\DontPrintSemicolon
\caption{gradient-free ALDI with estimated gradients \textbf{(EGI-ALDI)}}\label{alg:EGI-ALDI}
\Comment{additional linear equations: $J$} 
\KwData{$N \in \N, \{x_0^j\}_{j=1}^J$, $\tau>0$}
\KwResult{samples $\{x_n^j\}_{j=1, n=1}^{J,N}$ from $\d\mu/\d x \propto \exp(-V(x))$}
\For{$n\gets0$ \KwTo $N-1$ }{
    $\bar x_n\gets \frac{1}{J}\sum_{i=1}^Jx_n^i$\;
    $C_n^{1/2}\gets \frac{1}{\sqrt J}\left(x_n^1-\bar x_n,\ldots, x_n^J-\bar x_n\right)\in \R^{d\times J}$\;
    $C_n \gets \frac{1}{J}\sum_{i=1}^J(x_n^i-\bar x)\otimes (x_n^i-\bar x)$\;
    \For{$j\gets1$ \KwTo $J$ }{
    $g_n^j \gets G^j(\{x_n^i\}_{i=1}^J\cup\{\bar x_n\})$ \Comment{approximate gradient in each particle (not in mean), via \Cref{alg:grad}}    
    $W^j_n\sim \mathcal N(0,I^J)$\;
    $x_{n+1}^j \gets x_n^j - \tau C_n\cdot g_n^j + \tau\frac{d+1}{J}(x_n^j-\bar x_n) + \sqrt{2\tau}C^{1/2}_n W^j_n$ \Comment{ALDI  step}
    
    }

    }
\end{algorithm}

\subsection{Gradient extrapolation}
All methods presented so far employ gradient approximations computed in each ensemble point, see e.g. the \texttt{for}-loop in algorithm \ref{algline:for1}--\ref{algline:for2}, which means that each iteration requires the solution of $J$ linear equations (by evaluating $G^j(\{x_n^i\}_i)$ for $j=\{1,\ldots, J\}$). It is possible to strongly cut down on computational complexity at the cost of introducing an additional source of error: Instead of approximating each gradient, we can approximate the gradient and Hessian only in a suitably chosen reference point $x^\star$ (e.g. the ensemble mean) and extrapolate:
\begin{equation}
    \nabla V(x_n^j) \approx G^0(\{x^\star\}\cup\{x_n^i\}_{i=1}^J) + H^0(\{x^\star\}\cup\{x_n^i\}_{i=1}^J)\cdot (x_n^j-x^\star),
\end{equation}
which is reminiscent of $\nabla V(y) = \nabla V(x) + HV(x)\cdot (y-x) + o(\|y-x\|)$. Of course, this approximation is only valid if either the ensemble is sufficiently strongly concentrated or the measure is sufficiently Gaussian-like. To be more precise, we describe this variant for the EGI-ALDI algorithm:

\begin{algorithm}[H]
\DontPrintSemicolon
\caption{gradient-free ALDI with estimated and \textbf{extrapolated} gradients (EGI-ALDI-extra)}\label{alg:EGI-ALDI-extra}
\Comment{additional linear equations: 1} 
\KwData{$N \in \N, \{x_0^j\}_{j=1}^J$, $\tau>0$}
\KwResult{samples $\{x_n^j\}_{j=1, n=1}^{J,N}$ from $\d\mu/\d x \propto \exp(-V(x))$}
\For{$n\gets0$ \KwTo $N-1$ }{
    $\bar x_n\gets \frac{1}{J}\sum_{i=1}^Jx_n^i$\;
    $C_n^{1/2}\gets \frac{1}{\sqrt J}\left(x_n^1-\bar x_n,\ldots, x_n^J-\bar x_n\right)\in \R^{d\times J}$\;
    $C_n \gets \frac{1}{J}\sum_{i=1}^J(x_n^i-\bar x)\otimes (x_n^i-\bar x)$\;
    $g_n\gets G^0(\{\bar x_n\}\cup\{x_n^i\}_{i=1}^J)$ \Comment{approximate gradient only in mean, via \Cref{alg:grad}}  
    $H_n\gets H^0(\{\bar x_n\}\cup\{x_n^i\}_{i=1}^J)$ \Comment{Hessian approximation in mean, via \Cref{alg:grad}} 
    \For{$j\gets1$ \KwTo $J$ }{
    $g_n^j \gets g_n + H_n(x_n^j-\bar x_n)$\;  
    $W^j_n\sim \mathcal N(0,I^J)$\Comment{extrapolate estimated gradient}
    $x_{n+1}^j \gets x_n^j - \tau C_n\cdot g_n^j + \tau\frac{d+1}{J}(x_n^j-\bar x_n) + \sqrt{2\tau}C^{1/2}_n W^j_n$\Comment{ALDI  step}
    }
    }
\end{algorithm}

By taking EGI-ALDI-extra and using a sampled gradient approximation (\Cref{alg:gradBayes}) instead of the least square solution (\Cref{alg:grad}) we obtain a \rv{randomized version that may be more stable; a question that we leave for future research.}



\subsection{Numerical example: Sampling from a two-dimensional non-Gaussian measure}
\begin{figure}
    \centering
    \begin{subfigure}[t]{0.49\textwidth}
    \includegraphics[width=\textwidth]{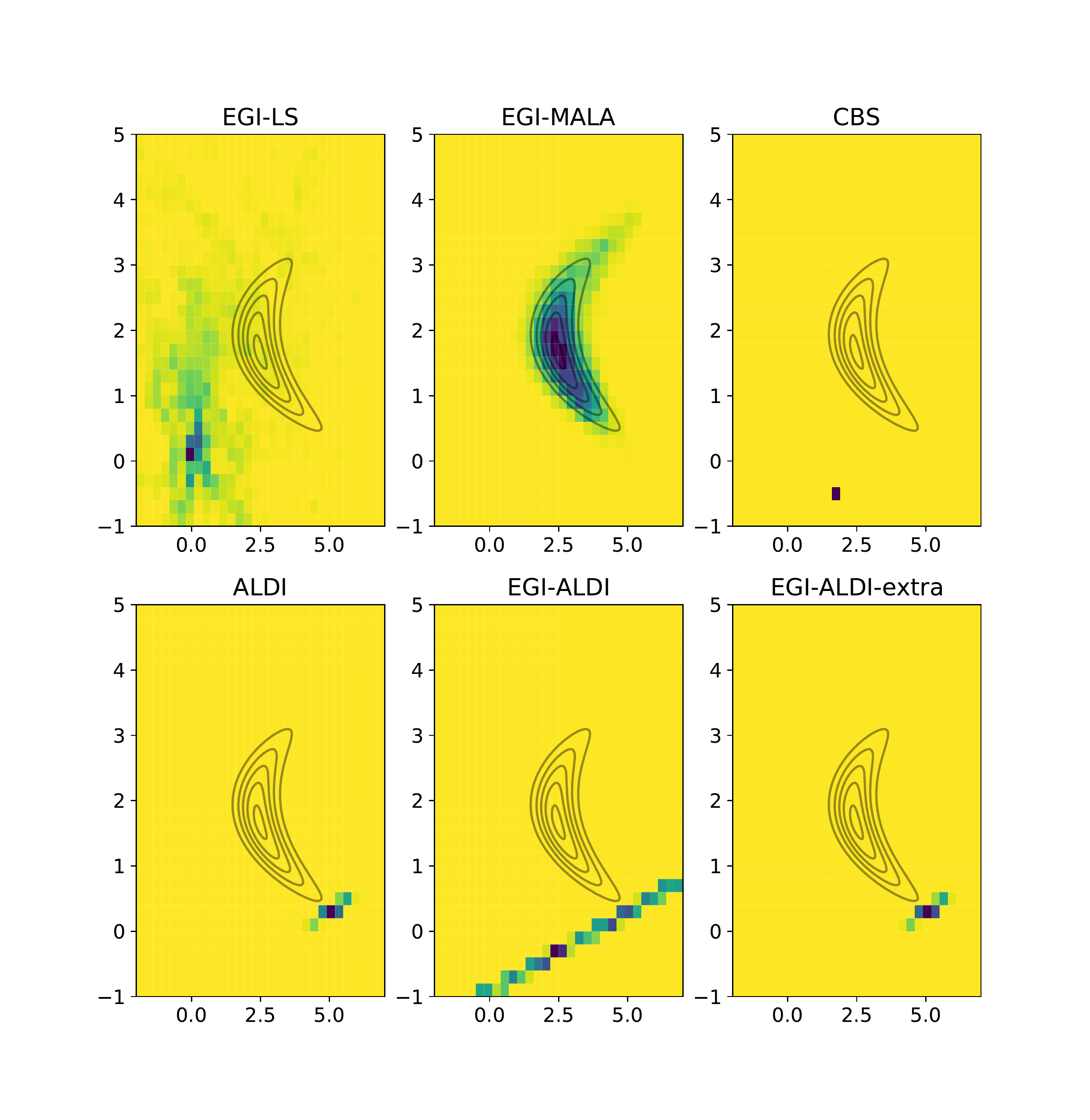}
    \caption{2d histograms of samplers. $J=2$, $N=10000$}
    \end{subfigure}\hfill\begin{subfigure}[t]{0.49\textwidth}
    \includegraphics[width=\textwidth]{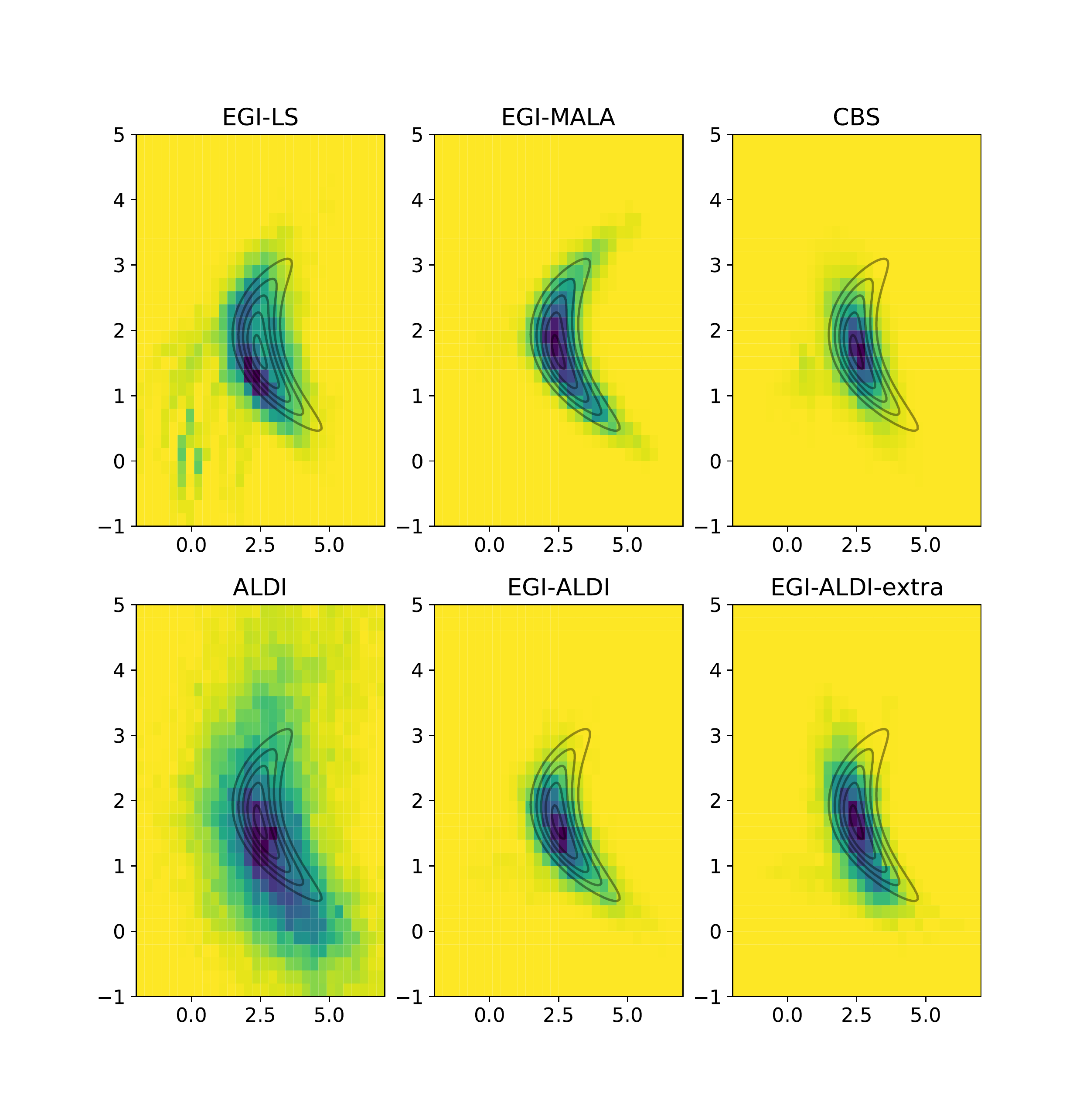}
    \caption{2d histograms of samplers. $J=20, N=2000$}
    \end{subfigure}\\
    
    \begin{subfigure}[b]{0.49\textwidth}
    \includegraphics[width=\textwidth]{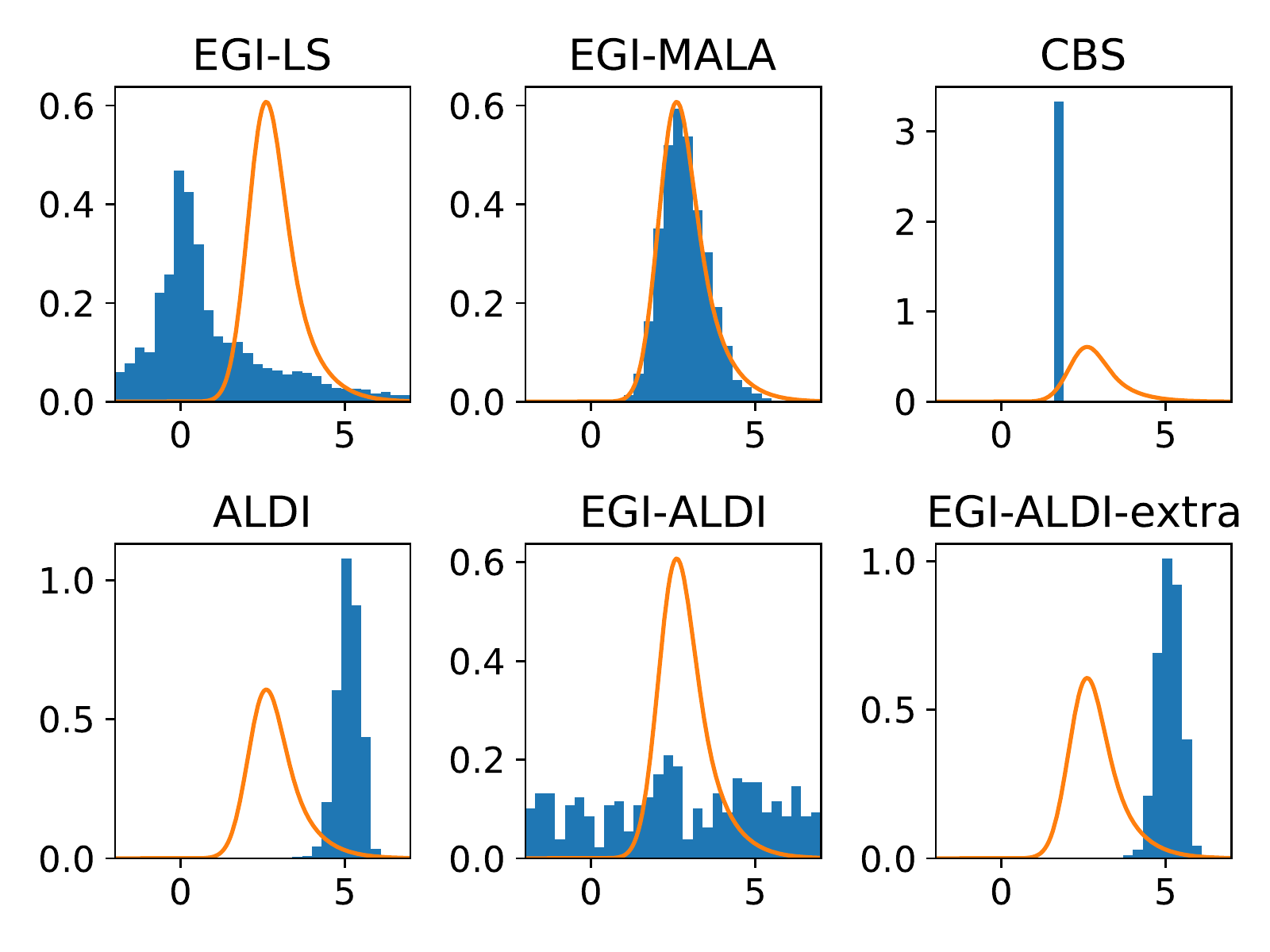}
    \caption{Marginal histogram of all six methods in comparison, where $J=2$, $N=10000$}
    \label{fig:c}
    \end{subfigure}\hfill
    \begin{subfigure}[b]{0.49\textwidth}
    \includegraphics[width=\textwidth]{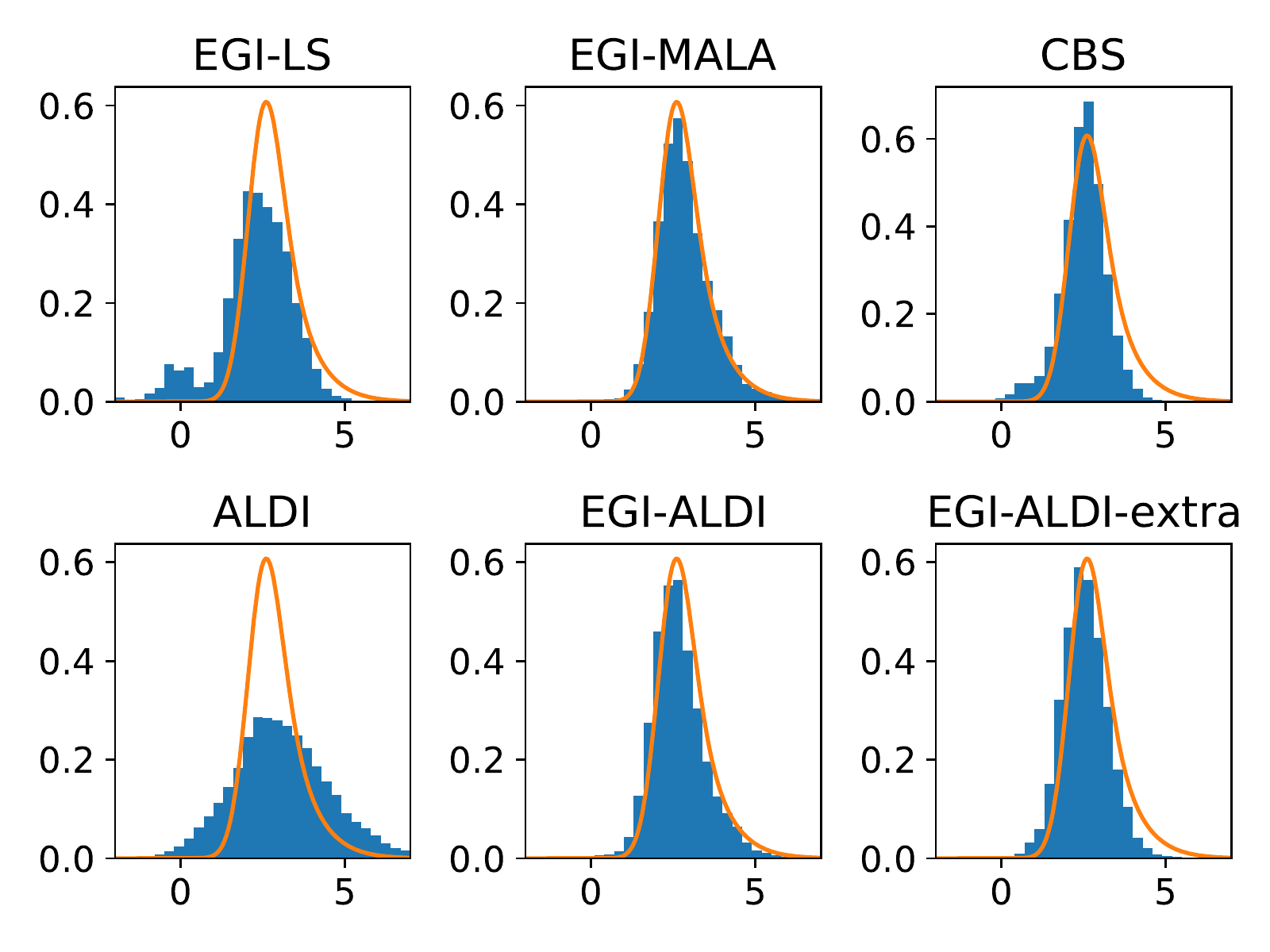}
    \caption{Marginal histogram of all six methods in comparison, where $J=20, N=2000$}
    \label{fig:d}
    \end{subfigure}
    \caption{Comparison of EGI-LS, EGI-MALA, gradient-free ALDI and EGI-ALDI. $J=2$, $N=10000$ (left column) and $J=20$, $N=4000$ (right column). Bottom row (figures \ref{fig:c}, \ref{fig:d}) show histograms of $x$-component of samples. The orange line marks the true marginal distribution. Code: \texttt{sampling\_2d\_J2.py, sampling\_2d\_J20.py}
    }
    \label{fig:sampling2d}
\end{figure}

We consider the inverse problem \rv{of inferring x, where}
\[y = A(x) + \eps\]
with $A:\R^2\to \R$, $A(x) = (x_2-2)^2-(x_1-3.5)-1$, $y=0$, and $\eps\sim \mathcal N(0,\tfrac{1}{2^2})$. We set a Gaussian prior $\mu_0 = \mathcal N(0_2, \tau^2\cdot I)$ with $\tau=2$ on the unknown parameter $x\in \R^2$. We write $\Phi(x) = \frac{1}{2\sigma^2}|y-A(x)|^2$, $\rv{V(x)} = \Phi(x) + \tfrac{1}{2\tau^2}\|x\|^2$. This leads to a Bayesian \rv{posterior $\mu^y$ on the parameter, which can be written as
\[ \frac{\d\mu^y}{\d\mu_0}(x) \propto \exp\left(-\Phi(x)\right) \text{ or  }\frac{\d\mu^y}{\d x}(x) \propto \exp\left(-\rv{V(x)}\right).\]}
Figure \ref{fig:sampling2d} shows the result of applying various samplers. We compare Ensemble Langevin (EGI-LS), Ensemble MALA (EGI-MALA), gradient-free ALDI (\cite{garbuno2020affine}, as an efficient gradient-free ensemble-based sampler), gradient-free ALDI augmented by approximated gradient information (EGI-ALDI), gradient-free ALDI with estimated and extrapolated gradients (EGI-ALDI-extra), and the CBS sampler of \cite{carrillo2022consensus}.
We start with a very small ensemble of size $J=2$ and $N=10000$. It can be observed that EGI-LS needs a larger number of ensemble members to perform well. This is due to the fact that $J=2$ does not yield a sufficiently good approximation to the gradient. Gradient-free ALDI, EGI-ALDI and EGI-ALDI-extra are restricted to the one-dimensional affine subspace spanned by the initial ensemble and also do not sample correctly from the posterior. CBS cannot extract sufficient information from its two-point ensemble: Both the subspace property as well as the fact that (in contrast to ALDI) no implicit gradient information is obtained leads to quick collapse to a point in the vicinity of the ``better'' ensemble member. EGI-MALA yields good samples: The shortcomings of EGI-LS are reigned in by the Metropolis adjustment (atrociously bad proposals due to faulty gradient approximations are rejected). For ensemble size $J=20$ (and $N=2000$) we see that all methods drastically improve: EGI-LS can improve its fit with the measure due to better gradient approximation (but still shows some bias), and ALDI is no longer constrained to a lower-dimensional affine subspace. Nevertheless, the forward mapping is sufficiently nonlinear that gradient-free ALDI incorporates substantial bias which strongly impacts its performance. CBS suffers in a similar way from its restricted applicability to non-Gaussian measures. EGI-ALDI and EGI-ALDI-extra profit from the gradient approximation but do not match EGI-MALA, which provides a near-perfect representation of the measure. Of course, this comes at the cost of needing to solve a larger number of linear equations as the following list (showing the amount of additional linear equations to solve per iteration) illustrates:
\begin{itemize}
    \item EGI-LS: $J$ linear equations (gradient approximation in each particle)
    \item EGI-MALA: $2\cdot J$ linear equations (gradient approximation in each particle and each proposal)
    \item CBS: no additional linear equations
    \item ALDI: no additional linear equations
    \item EGI-ALDI: $J$ linear equations (gradient approximation in each particle)
    \item EGI-ALDI-extra: $1$ linear equation (gradient approximation only in ensemble mean)
\end{itemize}
\begin{remark}
It is striking that EGI-ALDI-extra shows such a strong improvement over ALDI by use of just one additional linear equation solve per iteration, with EGI-ALDI not being much better (but much more expensive). EGI-MALA has the best performance but requires the highest number of linear equations to solve. 
\end{remark}

\subsection{Discussion of results}
\begin{itemize}
    \item If the system is ``almost linear'', gradient-free ALDI and CBS is much more efficient and should be preferred over EGI-LS or EGI-MALA, as there is no need to solve additional linear systems.
    \item If the ensemble size is limited (for example if evaluation of $V$ is expensive, e.g. if it involves the numerical solution of expensive computational models), then gradient approximation is a cost-effective add-on to completely gradient-agnostic methods like CBS and Random Walk Metropolis Hastings.
    \item If the measure is multimodal or has a non-negligibly curved shape and gradient information is unavailable, EGI-LS and EGI-MALA can accurately approximate measures.
    \item If the amount of ensemble members is chosen relatively small ($J\approx d$ or less), then EGI-MALA performs much better than EGI-LS due to the Metropolis-adjustment involved.
    \item The extrapolation idea can be used to cut down on the amount of linear equations that needs to be solved, from $\mathcal O(J)$ to $\mathcal O(1)$.
\end{itemize}

\section{Conclusion and outlook}
{We have described a way to turn implicit differential information from pointwise ensemble evaluation into an {estimator for first (and higher order) derivatives via Ensemble-based Gradient Inference (EGI)}. We have presented a novel way to augment Consensus-based optimization by adding an additional drift term proportional to this approximated gradient term which can be used to find better minima and accelerates local convergence. We have also demonstrated that sampling algorithms which usually work with exact gradients can be used with inexact gradients via EGI with similar performance, but without the need for explicit gradient evaluation.

There is a lot of perspective for future work: More theoretical analysis could shed some light on the right balancing between the three terms in EGI-CBO (gradient, consensus-building, and exploratory diffusion). At this point it is still unclear under which conditions EGI-CBO can be proven to be stable and/or what its limit points are. Similarly, the heuristical trade-off between local ($\xi=0$) and global ($\xi > 0$) EGI-CBO would need to be studied in more depth and this pertains to the sampling methods presented as well. In optimization settings where there are several similar optima one might be interested in finding all of them. A localization of CBO could help solve this issue and is subject to future work. {Also for sampling methods, the analysis of convergence properties will help to guide the development of EGI variants, in particular the analysis of computational costs of EGI variants vs. accuracy improvement.}
}

\section*{Acknowledgements}
PW acknowledges support from MATH+ project EF1-19: Machine Learning Enhanced Filtering Methods for Inverse Problems, funded by the Deutsche Forschungsgemeinschaft (DFG, German Research
Foundation) under Germany's Excellence Strategy – The Berlin Mathematics
Research Center MATH+ (EXC-2046/1, project ID: 390685689). 

\bibliographystyle{abbrvnat} 
\bibliography{lit} 
\end{document}